\documentclass[10pt,twocolumn,letterpaper]{article}

\usepackage{cvpr}              % To produce the CAMERA-READY version
\usepackage{booktabs}
\usepackage{multirow}
\definecolor{cvprblue}{rgb}{0.21,0.49,0.74}
\usepackage[pagebackref,breaklinks,colorlinks,allcolors=cvprblue]{hyperref}
\newcommand{\detPooledAUROC}{0.71}
\newcommand{\detPooledAUROCci}{[0.69, 0.73]}

\newcommand{\detNbeat}{4}
\newcommand{\detNmodels}{11}
\newcommand{\detNtest}{3569}

\newcommand{\detAUROCmin}{0.57}
\newcommand{\detAUROCmax}{0.77}
\newcommand{\detAUROCminAdded}{0.55}
\newcommand{\detAUROCmaxAdded}{0.76}
\newcommand{\detNbeatAdded}{1}
\newcommand{\detNmodelsAdded}{3}

\newcommand{\camAxisCMin}{37}
\newcommand{\camAxisCMax}{85}
\newcommand{\camAxisCMinF}{0.37}
\newcommand{\camAxisCMaxF}{0.85}
\newcommand{\agreeAxisCMin}{0.37}
\newcommand{\agreeAxisCMax}{0.88}
\newcommand{\camHandsMinF}{0.98}
\newcommand{\camHandsMaxF}{1.00}

\newcommand{\cellSentence}{in an exploratory comparison restricted to cells with at least ten runs, the cell in which both read the environment frame has the lowest median foreground-trajectory error on both sets (supplementary Section~R).}
\newcommand{\leverCMaxDev}{0.024}
\newcommand{\leverEscHalf}{0.318}
\newcommand{\leverEscSeven}{0.320}
\newcommand{\leverSensMask}{0.131}
\newcommand{\leverSensCam}{0.175}
\newcommand{\leverBMaxDev}{0.037}
\newcommand{\leverBMMaxDev}{0.013}

\newcommand{\leverSideChanges}{5}

\newcommand{\shareCutLow}{0.85}
\newcommand{\shareCutHigh}{0.06}
\newcommand{\shareMonLow}{0.77}
\newcommand{\shareMonHigh}{0.29}
\newcommand{\shareVdpmMin}{0.72}
\newcommand{\shareVdpmMax}{0.78}

\newcommand{\bench}{WhoMoved-Bench}
\def\confName{CVPR}
\def\confYear{2027}

\title{Who Moved? Camera-Object Motion Ambiguity in Feed-Forward 4D Reconstruction}

\author{Xuanchen Wang \enskip Heng Wang  \enskip Weidong Cai \\
University of Sydney \\
{\tt\small xwan0579@uni.sydney.edu.au}\\
{\tt\small \{heng.wang, tom.cai\}@sydney.edu.au}
}

\makeatletter
\let\wm@maketitle\@maketitle
\def\@maketitle{\wm@maketitle

  \begin{center}\includegraphics[width=\textwidth]{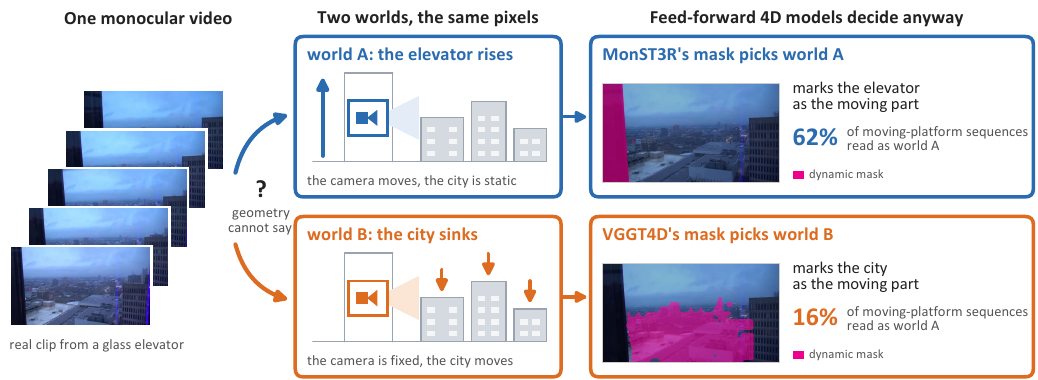}\end{center}

  {\def\@captype{figure}\caption{The same pixels, two worlds. A real clip from a glass elevator (left) is compatible with world~A, in which the elevator and its camera rise past a static city, and world~B, in which the city sinks past a fixed camera (middle); geometry cannot decide. Feed-forward 4D models decide anyway, by a learned prior (right, the dynamic mask in pink on one frame): MonST3R marks the elevator's frame as the moving part, VGGT4D marks the city, and each percentage is the share of the benchmark's moving-platform sequences on which that model's mask reads the physical frame, world~A. This paper measures that prior.}\label{fig:teaser}}
  \vspace{8pt}
  }
  
\makeatother

\begin{document}
\maketitle

\begin{abstract}
We introduce \bench{}, a benchmark for a decision that every feed-forward 4D reconstruction model makes on every monocular video and that geometry cannot make for it: when the pixels move, who moved? Four synthetic axes with exact ground truth (rigid world, dominant mover, camera-attached foreground on a moving platform, rotation-dominant flight), a matched control in which a sphere covers the same pixels spinning and frozen on a fixed camera path, ambiguity partitions of six existing benchmarks, and real driving and egocentric sets with ground truth. Across fifteen released models, a learned prior resolves the ambiguity, separately for each output. Under a dominant mover the camera absorbs the object's motion: the spinning sphere raises the camera error over its frozen twin at every coverage, the camera's rotation error carries a component along the spin axis that reverses with the spin on fourteen of fourteen models, and an oracle-assisted classical estimator on the same static pixels recovers the camera throughout. On a moving platform the mask reads the physical frame on 16 to 78\% of sequences and the camera on \camAxisCMin{} to \camAxisCMax{}\%. We propose a five-sequence data recipe with physical-frame labels and scope it against the simplest alternative: recalibrating MoRe's released threshold reaches the physical frame on up to 0.90 of real egocentric clips and the recipe's mask margin (0.35 against 0.34), training lifts the mover mask beyond the best threshold (IoU 0.43 against 0.34), three of four draws meet the pre-registered rate rule with no detected regression on six datasets, and the camera follows neither. The study was pre-registered; every deviation is a dated amendment.\footnote{The benchmark, the generator and evaluation code, the held-out split and the leaderboard are released upon publication; the leaderboard table and the test-split sequence list are in the supplementary material.}
\end{abstract}

\section{Introduction}
\label{sec:intro}

Feed-forward 4D reconstruction has moved fast. Given a monocular video, models such as MonST3R~\cite{zhang2025monst3r}, CUT3R~\cite{wang2025cut3r}, VGGT4D~\cite{vggt4d2025} and MegaSaM~\cite{li2025megasam} return a camera trajectory, per-frame depth and a decomposition of the scene into a static world and moving content. They are evaluated on datasets whose ground truth came from a camera moving through a mostly static world, a setting that hides a decision the models make on every video and that no geometry can make for them: when the pixels move, who moved?

\Cref{fig:teaser} states the problem. A rigid world seen from a moving camera and a rigid world moving past a still camera produce the same image sequence; a world in which one object drifts produces the same sequence as a world in which everything but that object drifts. Geometry fixes structure up to this choice; priors about ground planes, horizons, people, vehicles and gravity make it. The kinetic depth literature made the point for human observers seventy years ago~\cite{wallach1953kinetic}; dynamic SLAM hard-codes the prior that the largest rigid component is the world~\cite{bescos2018dynaslam,zhang2022casualsam}. For learned feed-forward models the prior is implicit, untested and, as we show, different from model to model and from output to output.

This paper measures it. We introduce \bench{}, a benchmark built around the ambiguity (\Cref{fig:axes}): a rigidly moving world with eight cues switched on and off (Axis~A), a dominant mover covering a controlled fraction $f$ of the pixels (Axis~B), a camera-attached foreground over a world seen from a moving platform (Axis~C), rotation-dominant flight (Axis~D), and a matched control in which a sphere covers the same pixels spinning and frozen on a fixed camera path (Axis~B$'$). Where ground truth exists it is exact; where several decompositions generate the same pixels the recorded generating decomposition is the reference and an attribution ratio reads the model's answer. Ambiguous partitions of six existing benchmarks and real driving and egocentric sets with ground-truth poses and masks connect the axes to practice.

The central finding is that the ambiguity is resolved by a prior, and resolved separately by each output. Under a dominant mover (Axis~B) the camera absorbs the object's motion: on the matched control a spinning sphere raises the camera error over its frozen twin at every coverage, the camera's rotation error acquires a signed component along the spin axis that reverses with the spin, and an oracle-assisted classical estimator on the same static pixels recovers the camera throughout, so the evidence is there and the networks misattribute it; the masks degrade without reversing. On a moving platform (Axis~C) the mask and the camera choose their worlds separately, and the true static region recovers 10 to 13\% of the camera error. The mask's prior on MoRe sits in its threshold: recalibration reaches the physical frame on real egocentric video with the margin that five rendered sequences give, the sequences add the mover mask, and the camera follows neither. The study was pre-registered and the test split was opened for pre-registered evaluations alone (\Cref{sec:prereg}). Our contributions are:
\begin{itemize}
  \item We introduce \bench{}: 1248 synthetic sequences on four axes with exact ground truth (624 held-out test), the 100-sequence matched control and its opposite-spin twin, the generator, an ambiguity score with top-20\% partitions for six existing benchmarks, and real driving (83 KITTI) and egocentric (216 Aria Digital Twin) clips with ground-truth poses and masks.
  \item We show through six interventions that the failure is one of attribution: the true static region as input, the matched control, its spin reversed, an oracle-assisted classical estimator on the remaining static pixels, fine-tunes of four architectures, and real video with ground truth.
  \item We propose per-output frame labels, validated on constructed reconstructions, and a five-sequence data recipe: the mask's prior on MoRe is a calibration, the recipe adds the mover mask, and the camera follows neither.
  \item We evaluate fifteen models (\Cref{tab:leaderboard,tab:agreement}) and provide practical guidance by output, a comparison with existing datasets (\Cref{tab:benchmarks}), three ceilings on inference-time repairs and the pre-registration record.
\end{itemize}

\section{Related Work}
\label{sec:related}

\subsection{Feed-Forward 4D Reconstruction}
DUSt3R~\cite{wang2024dust3r} and MASt3R~\cite{leroy2024mast3r} replaced matching and bundle adjustment with a network that regresses point maps in a common frame; VGGT~\cite{wang2025vggt} and $\pi^3$~\cite{wang2025pi3} scaled the idea to many views. For dynamic scenes, MonST3R~\cite{zhang2025monst3r} fine-tunes the pairwise model on dynamic data and reads a motion mask from optical flow, CUT3R~\cite{wang2025cut3r} keeps a persistent state across frames, Easi3R~\cite{chen2025easi3r} and VGGT4D~\cite{vggt4d2025} read motion from attention without training, TraceAnything~\cite{traceanything2025} represents a video as a field of trajectories, Any4D~\cite{any4d2025}, 4RC~\cite{fourrc2025}, V-DPM~\cite{vdpm2025}, MoRe~\cite{more2025}, UniQuery4R~\cite{uniquery4r2026}, Point4D~\cite{point4d2026} and PAGE-4D~\cite{page4d2026} add 4D heads and training recipes of their own, and Depth Anything~3~\cite{da3_2025} provides cameras and depth from any number of views. MoRe already trains its motion head with motion labels, so the recipe of \Cref{sec:recipe} contributes the bias it reveals and the data it takes, not the supervision itself. All are evaluated as released (training data in supplementary Table~4); MegaSaM~\cite{li2025megasam}, a learned depth prior inside differentiable bundle adjustment, is the optimisation-based reference, and D4RT~\cite{d4rt2025} is not evaluated because its authors have released no implementation or checkpoint at the time of submission.

\begin{figure*}[t]
  \centering
  \includegraphics[width=\textwidth]{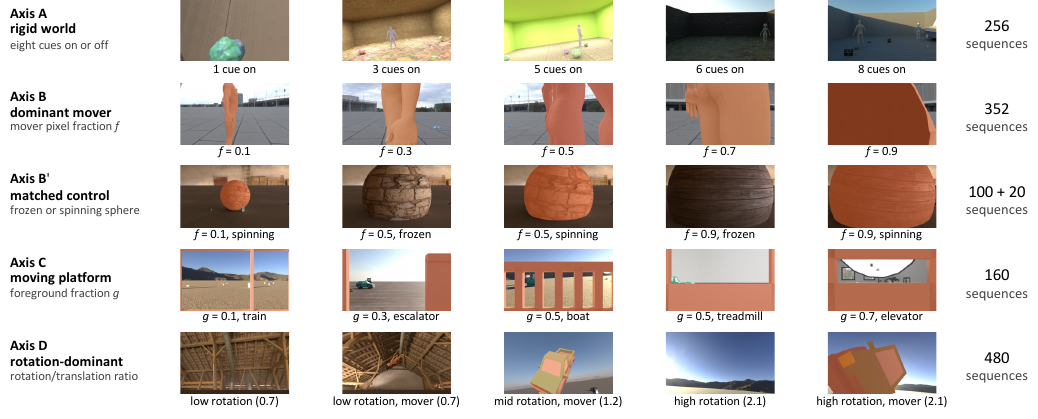}
  \caption{\bench{}, one row per axis and the matched control, frames across the lever with the ground-truth decomposition overlaid where one exists. Axis~A: the whole visible content moves rigidly, with one to eight cues on (tiles ordered by cue count); no decomposition is identifiable from the images (the generating one is recorded). Axis~B: one rigid mover covers a controlled pixel fraction $f$; Axis~B$'$: the same pixels frozen and spinning. Axis~C: a camera-attached foreground over a world seen from a moving platform, the environment frame defined by the ground truth. Axis~D: rotation-dominant flight.}
  \label{fig:axes}
\end{figure*}
\subsection{Dynamic SLAM and Motion Ambiguity}
Classical dynamic SLAM resolves the ambiguity by a rule: DynaSLAM~\cite{bescos2018dynaslam} removes a-priori dynamic classes before tracking, DROID-SLAM~\cite{teed2021droid} relies on the dominance of the static background, CasualSAM~\cite{zhang2022casualsam} and ParticleSfM~\cite{zhao2022particlesfm} learn a motion probability per pixel, LEAP-VO~\cite{chen2024leapvo} filters tracks~\cite{karaev2024cotracker} by temporal consistency, and consistent video depth methods~\cite{kopf2021rcvd} assume the camera explains most of the flow. Each is a prior made explicit, inherited by 4D representations such as Shape of Motion~\cite{wang2024shapeofmotion}; for learned feed-forward models the prior is implicit. 

\subsection{Dynamic Reconstruction Datasets}
\Cref{tab:benchmarks} sets \bench{} next to the datasets this paper evaluates on. Sintel~\cite{butler2012sintel}, TUM RGB-D~\cite{sturm2012tum}, Bonn~\cite{palazzolo2019bonn}, DyCheck~\cite{gao2022dycheck}, Kubric MOVi-F~\cite{greff2022kubric} and PointOdyssey~\cite{zheng2023pointodyssey} contain the ambiguous regime as a tail, movers are small and the camera dominates the flow, and none controls the lever that worsens it; \bench{} makes that regime the independent variable. KITTI~\cite{geiger2012kitti} (driving, with ego poses and object tracks) and Aria Digital Twin~\cite{pan2023adt} (egocentric video with a digital twin) are the real counterparts of Axes~B and C. StaDy4D~\cite{tsui2026stady4d} pairs static and dynamic renders of the same driving scenes for a pipeline that removes dynamic content before reconstruction; Axis~B$'$ keeps the foreground and its silhouette and changes its motion alone, so the visible background is preserved rather than revealed.

\subsection{Priors Versus Geometry}
Bratuli\'c et al.~\cite{bratulic2025geometry} asked how much of a static feed-forward model's output is geometry and how much is prior, probing DUSt3R, VGGT and Depth Anything~3 and testing VGGT under occlusion, repetition and focal-length change, and found learned geometric structure: epipolar geometry emerges in intermediate layers and is causally linked to correspondence patterns in attention. We ask the corresponding question for dynamic models, where the ambiguity is the central inference on every video, on axes that worsen it by a known amount. The study design follows the pre-registration practice of the empirical sciences~\cite{nosek2018preregistration}.

\section{\bench{}}
\label{sec:bench}

\subsection{Ambiguity and Metrics}
\label{sec:metrics}
\Cref{fig:teaser} is the object of measurement: a video is compatible with world~A, in which the camera moves, and world~B, in which the world moves, and a model's cameras, depth and mask each encode a choice. The camera-attribution ratio $r_\mathrm{cam}$ is the model's estimated camera travel over the true relative travel, made scale-free with its own depth of the placed objects (supplementary Section~B): 1 reads the motion as camera motion, 0 as world motion. Camera error is ATE after Sim(3) alignment~\cite{umeyama1991least} over the ground-truth path length (ATE/path) and mask quality is IoU; a sweep of the mover's pixel fraction $f$ gives the mask degradation point $f_\mathrm{mask}$, the smallest $f$ at which IoU falls below half its $f=0.10$ value or below 0.20, and the pose degradation threshold $f_\mathrm{deg}$, the smallest $f$ at which ATE/path exceeds three times its $f=0.10$ value (the pre-registered relative reading) or 0.10 (the absolute reading). On a camera-attached foreground each output gets a frame label. The mask attribution margin $m$ is the mean mask value on the foreground minus that on the background, physical when $m>0$ (the frame-choice rate is the share of sequences with $m>0$). The camera label reads the predicted camera travel against the predicted background depth, physical when the ratio exceeds half the ground truth's; on constructed reconstructions it reads the true one as physical and the object-frame one as egocentric, and it holds under a threefold error in the foreground depth, which the foreground-normalised ratio of Amendment~20 does not (supplementary Section~R). The label measures translation magnitude against background depth under the two frame hypotheses and not the trajectory's direction, timing or the reconstruction's consistency; agreement denotes matching labels under the two diagnostics. The downstream error is that of the foreground's world trajectory. Statistics are bootstrap 95\% intervals with the scene seed, recording or drive as the resampling unit.

\subsection{Synthetic Axes}
\label{sec:axes}
Each axis worsens the ambiguity with one lever (\Cref{fig:axes}; counts per axis and split in supplementary Table~3). Sequences have 30 frames at $512\times288$ pixels in metric scenes rendered with Blender's Cycles, with the camera, depth, instance ids, per-pixel 3D motion and the generating decomposition per frame. Every lever is measured on the ground-truth masks: the sequence mean of the realised share is within \leverBMaxDev{} of the nominal $f$ in every Axis~B cell and within \leverBMMaxDev{} on Axis~B$'$ (Amendment~25).

% Session 10, Part D: benchmark comparison. Properties of the other datasets from their papers (cited in the caption);
% sequence and ambiguous counts are the sets this paper evaluates (data/bench/existing_partitions.parquet,
% data/bench/realvideo_index.parquet), ambiguous = the top-20% ambiguity partition; \bench{} by construction.
\begin{table}[t]
\centering
\scriptsize
\setlength{\tabcolsep}{3pt}
\caption{\bench{} next to the datasets the paper evaluates on and Kubric~\cite{greff2022kubric}; properties from each paper~\cite{butler2012sintel,sturm2012tum,palazzolo2019bonn,gao2022dycheck,zheng2023pointodyssey,geiger2012kitti,pan2023adt}; counts are the evaluated sets, ambiguous is the top-20\% partition. \bench{} is the one set that controls the lever that worsens the ambiguity.}
\label{tab:benchmarks}
\begin{tabular}{lccccccrr}
\toprule
Dataset & \rotatebox{90}{\parbox{2.1cm}{\raggedright Real video}} & \rotatebox{90}{\parbox{2.1cm}{\raggedright Exact dynamic masks}} & \rotatebox{90}{\parbox{2.1cm}{\raggedright Ground-truth cameras}} & \rotatebox{90}{\parbox{2.1cm}{\raggedright Dominant mover, controlled $f$}} & \rotatebox{90}{\parbox{2.1cm}{\raggedright Camera-attached fg.\ on a platform}} & \rotatebox{90}{\parbox{2.1cm}{\raggedright Rotation-dominant paths}} & \rotatebox{90}{\parbox{2.1cm}{\raggedright Sequences}} & \rotatebox{90}{\parbox{2.1cm}{\raggedright Ambiguous (top 20\%)}} \\
\midrule
\bench{} &  & $\checkmark$ & $\checkmark$ & $\checkmark$ & $\checkmark$ & $\checkmark$ & 1248 & 1248 \\
Sintel &  &  & $\checkmark$ &  &  &  & 14 & 3 \\
TUM RGB-D & $\checkmark$ &  & $\checkmark$ &  &  & $\checkmark$ & 8 & 2 \\
Bonn & $\checkmark$ &  & $\checkmark$ &  &  &  & 5 & 1 \\
DyCheck & $\checkmark$ &  & $\checkmark$ &  &  &  & 7 & 2 \\
MOVi-F &  & $\checkmark$ & $\checkmark$ &  &  &  & 100 & 20 \\
PointOdyssey &  &  & $\checkmark$ &  &  &  & 52 & 11 \\
KITTI & $\checkmark$ &  & $\checkmark$ &  &  &  & 83 & 17 \\
Aria Digital Twin & $\checkmark$ & $\checkmark$ & $\checkmark$ &  &  &  & 216 & 44 \\
Kubric (generator) &  & $\checkmark$ & $\checkmark$ &  &  &  &  &  \\
\bottomrule
\end{tabular}
\end{table}

\paragraph{Axis A: Rigid World (256 Sequences).} The entire visible content moves rigidly relative to the camera, so the sequence is identical whether the camera moved or the world did; eight cues (ground plane, horizon, gravity-aligned roll, people, vehicles, indoors, known-size objects, texture density) are switched on or off in a resolution-IV factorial design, and $r_\mathrm{cam}$ reads the answer.

\paragraph{Axis B: Dominant Mover (352).} A static scene, a moving camera and one rigid mover covering $f \in \{0.05, 0.10, 0.20, \dots, 0.90, 0.95\}$ of the pixels in four object classes (person-shaped, vehicle-shaped, household, abstract) over eight scene seeds; the camera closes on the mover as $f$ grows, as in driving.

\paragraph{Axis B$'$: Matched Control (100 + 20).} Five Axis~B scenes with the camera path and background fixed at their $f=0.10$ configuration and a textured sphere covering $f \in \{0.1, 0.3, 0.5, 0.7, 0.9\}$ by its radius alone, rendered spinning, frozen and spun the other way; within a pair the silhouette and the visible background are identical (minimum mask IoU 0.999).

\paragraph{Axis C: Moving Platform (160).} A foreground rigidly attached to the camera (vehicle interior, railing, window frame) covers a nominal $g\in\{0.1,0.3,0.5,0.7\}$ of the pixels (realised shares in supplementary Table~1) over a world seen from a train, boat, escalator, elevator or treadmill; in the environment frame the surroundings are stationary and the foreground moves with the platform.

\paragraph{Axis D: Rotation-Dominant Motion (480).} Sixty FPV flight trajectories from a Betaflight-in-Gazebo simulation, rendered in four scenes with and without a mover and binned by rotation-to-translation ratio, read by static-pixel depth error and the mask false-positive rate.

\subsection{Existing Benchmarks and Real Video}
\label{sec:realsets}
An ambiguity score per sequence from ground truth (z-scored dynamic-pixel fraction plus rotation-to-translation ratio) partitions Sintel, TUM RGB-D dynamic, Bonn, DyCheck, MOVi-F and PointOdyssey at their top 20\%. The real driving set takes 83 clips of 30 frames from the KITTI tracking sequences~\cite{geiger2012kitti}: 41 with a moving vehicle ahead covering a box fraction of 0.05 to 0.30 (the Axis~B analogue), 40 random and 2 convoy windows, with ego poses from the GPS-inertial track and dynamic masks from the tracks that move in the world frame. The egocentric set takes 216 clips from 17 Aria Digital Twin recordings~\cite{pan2023adt}, with poses, depth and masks from the twin: 105 with a moving person or object (Axis~B analogue), 81 with the wearer's hands or a held object as a camera-attached foreground (Axis~C analogue, binned as $g$) and 30 random windows. Validation and test use disjoint scene seeds and asset halves; the pre-registered release of 1248 sequences (624 validation, 624 test) carries every verdict, and the extended release adds the 352 Axis~B sequences of Amendment~13 and the 120 of Axis~B$'$ (supplementary Section~N).

% generated by scripts/paper_leaderboard_main.py (Session 11; Session 12 moved the three threshold columns to the supplement); cells in results/phase4/leaderboard_main.json, traced in paper/NUMBERS.md (F14); do not edit
\begin{table*}[t]
  \centering
  \scriptsize
  \setlength{\tabcolsep}{3pt}
  \caption{Leaderboard (Axis B thresholds and every interval in supplementary Table~11). The mask degrades before the pose does wherever both thresholds exist, V-DPM excepted, and the mask reads the physical frame on 0.16 to 0.78 of moving-platform sequences. Axis B ATE/path at $f\ge0.5$; $\Delta$: change with the true static region, run on the eleven pre-registered pose models; FPR: worst rotation bin. Bold: best where a direction exists; n/a: no such output or run.; FPR: worst rotation bin. Bold: best where a direction exists; none: no crossing below 0.95; n/a: no such output.}
  \label{tab:leaderboard}
  \resizebox{\textwidth}{!}{\begin{tabular}{llllllllll}
    \toprule
    & & \multicolumn{2}{c}{Axis B, dominant mover} & \multicolumn{3}{c}{Axis C, moving platform} & Axis D & \multicolumn{2}{c}{real video} \\
    \cmidrule(lr){3-4} \cmidrule(lr){5-7} \cmidrule(lr){8-8} \cmidrule(lr){9-10}
    model & mask & ATE/path & $\Delta$ (\%) & phys. & ATE/path & $\Delta$ (\%) & FPR & hands phys. & KITTI ratio \\
    \midrule
    MonST3R & own & 0.118 & 0.0 & 0.62 & 0.174 & -10.6 & 0.40 & 0.93 & \textbf{1.62} \\
    CUT3R & derived & 0.123 & +23.8 & 0.42 & 0.168 & -17.5 & 0.42 & 0.85 & 3.22 \\
    Easi3R & own & 0.143 & +7.7 & 0.39 & 0.218 & -19.6 & 0.10 & 0.77 & 1.82 \\
    VGGT4D & own & 0.116 & \textbf{-8.2} & 0.16 & 0.145 & +6.0 & 0.32 & 0.33 & 4.06 \\
    $\pi^3$ & derived & 0.105 & +41.4 & 0.65 & \textbf{0.128} & +4.2 & 0.38 & 0.85 & 3.84 \\
    Any4D & derived & 0.121 & +25.2 & 0.70 & 0.194 & -5.0 & 0.33 & 0.86 & 2.97 \\
    MegaSaM & own & 0.091 & +29.4 & 0.37 & 0.184 & \textbf{-21.7} & 0.04 & 0.89 & 4.76 \\
    Depth Anything 3 & derived & \textbf{0.088} & +24.0 & 0.57 & 0.129 & -16.4 & 0.48 & 0.90 & 4.12 \\
    4RC & derived & 0.104 & +11.1 & \textbf{0.78} & 0.150 & -6.1 & 0.06 & 0.23 & 4.06 \\
    V-DPM & derived & 0.120 & +16.2 & 0.74 & 0.141 & -9.8 & 0.17 & \textbf{0.95} & 4.18 \\
    MoRe & own & 0.115 & +15.3 & 0.21 & 0.137 & -7.1 & 0.01 & 0.31 & 3.55 \\
    UniQuery4R & own & 0.107 &  & 0.41 & 0.143 &  & \textbf{0.00} & 0.94 & 3.86 \\
    Point4D & derived & 0.131 &  & 0.47 & 0.142 &  & 0.46 & 0.78 & 3.33 \\
    PAGE-4D & derived & 0.125 &  & 0.75 & 0.170 &  & 0.25 & 0.94 & 4.31 \\
    TraceAnything & own & n/a & n/a & 0.63 & n/a & n/a & 0.34 & n/a & n/a \\
    \bottomrule
  \end{tabular}}
\end{table*}

% generated by scripts/phase10_camlabel.py (Amendment 23, A23.1; the camera label of A21.1 in place of r_cd; no checkpoint rows, see Figure 4c and supplementary Table 13); traced in paper/NUMBERS.md (F16); do not edit
\begin{table}[t]
  \centering
  \footnotesize
  \setlength{\tabcolsep}{3pt}
  \caption{Attribution across outputs (Amendments~20 and 23). Per model on the Axis~C sequences and the real hands clips: the share whose dynamic mask reads the physical frame, whose camera reads it (predicted travel against predicted background depth, above half the ground truth's), and the share on which the two agree; every checkpoint, the foreground-normalised depth diagnostic and the downstream error are in supplementary Table~13.}
  \label{tab:agreement}
  \begin{tabular}{@{}lrrr|rrr@{}}
    \toprule
    & \multicolumn{3}{c|}{Axis C (synthetic)} & \multicolumn{3}{c}{real hands clips} \\
    model & mask & camera & agree & mask & camera & agree \\
    \midrule
    MonST3R & 0.62 & 0.48 & 0.64 & 0.93 & 1.00 & 0.93 \\
    CUT3R & 0.42 & 0.42 & 0.63 & 0.85 & 1.00 & 0.85 \\
    Easi3R & 0.39 & 0.66 & 0.38 & 0.77 & 1.00 & 0.77 \\
    VGGT4D & 0.16 & 0.70 & 0.37 & 0.33 & 1.00 & 0.33 \\
    $\pi^3$ & 0.65 & 0.85 & 0.67 & 0.85 & 1.00 & 0.85 \\
    Any4D & 0.70 & 0.60 & 0.64 & 0.86 & 1.00 & 0.86 \\
    MegaSaM & 0.37 & 0.37 & 0.88 & 0.89 & 0.98 & 0.89 \\
    Depth Anything 3 & 0.57 & 0.65 & 0.72 & 0.90 & 1.00 & 0.90 \\
    4RC & 0.78 & 0.55 & 0.65 & 0.23 & 1.00 & 0.23 \\
    V-DPM & 0.74 & 0.71 & 0.83 & 0.95 & 1.00 & 0.95 \\
    MoRe & 0.21 & 0.78 & 0.43 & 0.31 & 1.00 & 0.31 \\
    UniQuery4R & 0.41 & 0.43 & 0.67 & 0.94 & 1.00 & 0.94 \\
    Point4D & 0.47 & 0.60 & 0.77 & 0.78 & 1.00 & 0.78 \\
    PAGE-4D & 0.75 & 0.82 & 0.67 & 0.94 & 1.00 & 0.94 \\
    TraceAnything & 0.63 & n/a & n/a & n/a & n/a & n/a \\
    \bottomrule
  \end{tabular}
\end{table}

\subsection{Models and Protocol}
\label{sec:protocol}
Fifteen models with public code run through one interface at pinned commits with the authors' settings and resolution (supplementary Table~2); twelve were pre-registered and three added under Amendment~16 enter the tables and figures, not the verdicts. Models without a dynamic head get a mask derived from their own 3D-motion or depth residual with one threshold per model frozen on the validation split (\Cref{tab:leaderboard}). TraceAnything predicts no cameras and enters the mask metrics alone. Every pre-registered camera head and six mask heads pass the low-occupancy sanity check at $f\le0.10$ (supplementary Section~C). Hypotheses, thresholds, metrics and kill criteria were fixed before the first sequence was rendered (\Cref{sec:prereg}).

\section{Experiments and Results}
\label{sec:findings}

\subsection{Leaderboard}
\label{sec:leaderboard}
\Cref{tab:leaderboard} is the benchmark read once per model, and \Cref{tab:agreement} reads each model twice, by its mask and by its camera. On Axis~B the dynamic mask degrades at a model-specific fraction, from $f_\mathrm{mask}=0.30$ (VGGT4D) to 0.90 ($\pi^3$), six of the eleven pose models below $f=0.95$, and the mask degrades before the pose does wherever both thresholds exist, V-DPM excepted; person-shaped movers push ten of eleven models past the relative threshold, household objects none. On Axis~C the mask reads the physical frame on 0.16 (VGGT4D) to 0.78 (4RC) of the sequences and the camera on \camAxisCMinF{} to \camAxisCMaxF{}; the elevator, a textureless shaft behind a strong foreground, defeats eleven of fifteen masks. On Axis~A the visible ground plane raises $r_\mathrm{cam}$ for most models, and on Axis~D three derived-mask models mark 0.35 to 0.51 of a static scene as dynamic in every rotation bin (supplementary Section~F).

\subsection{Dominant Mover: Misattribution}
\label{sec:regimes}
\begin{figure*}[t]
  \centering
  \includegraphics[width=\textwidth]{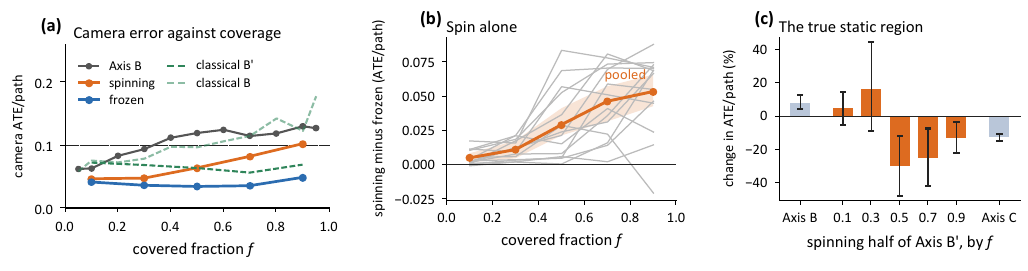}
  \caption{Under a dominant mover the camera absorbs the motion (Amendments~20 and 22). (a) Camera ATE/path against the covered fraction on Axis~B$'$ (the same pixels frozen and spinning on a fixed camera path), next to Axis~B and to an oracle-assisted classical estimator on the same static pixels (RAFT flow, five-point essential matrix, ground-truth static region and step length); dotted: the absolute threshold of 0.10. (b) Spinning minus frozen, per model and pooled with its cluster interval. (c) Change of camera error with the true static region on the spinning half of Axis~B$'$ by fraction and on Axes~B and C. }
  \label{fig:diagnostic}
\end{figure*}

Every model ran on the ambiguous validation subsets as released and with the ground-truth static region supplied (as a mask input or by filling the true mover; supplementary Figure~16). A model that resolves the two worlds of \Cref{fig:teaser} the wrong way should improve when told the right one. On Axis~B it does not: above its degradation threshold the four native-mask models change ATE/path by $+7.5\%$ [$+3.9$, $+12.8$] with the true static region, all eleven by $+17.1\%$ [$+14.4$, $+19.4$], and below the threshold the sign is the same ($+5.0\%$ [$+1.6$, $+8.6$]). That admits two readings, too few static pixels (the camera closes on the mover along Axis~B) or misattribution of the mover's motion, and the matched control separates them (\Cref{fig:diagnostic}).

\paragraph{The Matched Control.} On Axis~B$'$ the camera path and the background never change and the sphere covers the same pixels spinning and frozen; the frozen sphere is itself static texture, so the pair preserves the visible background rather than every static feature, and the classical control on the background alone answers that. The frozen sphere costs almost nothing (camera error between 0.034 and 0.048 across the sweep); the spinning one raises it at every coverage, by $+0.005$ [$+0.002$, $+0.008$] ATE/path at a tenth of the frame and $+0.053$ [$+0.043$, $+0.066$] at nine tenths, $+0.029$ [$+0.024$, $+0.034$] pooled over 699 pairs and fourteen models. The error has a direction, and it follows the spin: the component of the camera's rotation error along the spin axis, scaled so that 0 is the environment frame and $+1$ the camera that turns with the sphere, is $+0.20$ [$+0.16$, $+0.23$] with the forward spin and $-0.20$ [$-0.24$, $-0.16$] with the spin reversed (twenty cells at $f \in \{0.5, 0.9\}$; $0.00$ frozen), a paired difference of $+0.40$ [$+0.33$, $+0.47$] with the sign agreeing on fourteen of fourteen models, and the forward bias rises from $+0.07$ at $f=0.5$ to $+0.33$ at $f=0.9$ over the full set (Amendment~22, supplementary Section~R). Losing static pixels is not what hurts the camera; the object's motion is, and here the true static region recovers $-15.0\%$ [$-22.4$, $-7.3$] of the error for the native-mask models, $-30.5\%$ at $f=0.5$ (\Cref{fig:diagnostic}c). An oracle-assisted classical estimator on the same static pixels (RAFT flow restricted to the ground-truth static region, the five-point essential matrix chained frame to frame, the ground-truth step length; Amendment~20) stays below 0.10 ATE/path at every coverage on Axis~B$'$ (0.056 to 0.073), while the networks rise to 0.102 (\Cref{fig:diagnostic}a): the evidence to place the camera is there throughout, and the networks turn the camera with the sphere instead of using it.

\paragraph{Axis B and Its Regime.} On Axis~B, where the camera closes on the mover, the same estimator recovers the camera up to $f=0.5$ (0.097) and degrades beyond it (0.106 at $f=0.6$, 0.177 at $f=0.95$), and ten of the fourteen pose models cross their absolute threshold at or below $f=0.5$. So the networks fail where an oracle-assisted estimator succeeds; beyond $f=0.5$ that estimator degrades as well, and a scale-free chain sits at the threshold from the smallest coverage (Amendment~22), so these controls bound conditioning, not what any estimator could recover. None of the seven native masks reverses at its degradation point (four degrade to a partial mask, one vanishes; A20.1): on Axis~B the mask degrades and the camera absorbs the motion, and nothing hands the mover to the world.

%% A20.5 figure begin
%% A20.5 figure end

\subsection{Moving Platform: Frame Choice}
\label{sec:platform}
On Axis~C the environment frame is defined by the ground truth, and the models disagree with it and with themselves (\Cref{tab:agreement}). The mask reads the physical frame on 0.16 to 0.78 of the sequences and the camera on \camAxisCMinF{} to \camAxisCMaxF{}, and the two labels agree on \agreeAxisCMin{} to \agreeAxisCMax{}; the foreground-normalised ratio of Amendment~20 read the camera as egocentric on most sequences, and the constructed cases show that it answers to the foreground's depth, so it stays a depth diagnostic (supplementary Section~R). Here the true static region works, lowering ATE/path by $-12.8\%$ [$-14.9$, $-10.8$] over the four native-mask models and $-10.3\%$ [$-11.7$, $-8.6$] over all eleven: the models had the static context and used it for the wrong world. The choice depends on how much of the frame the platform fills, and not for every model: by the realised foreground share, CUT3R's mask reads the physical frame on \shareCutLow{} of the sequences at 0.1 and on \shareCutHigh{} at 0.5 and MonST3R's on \shareMonLow{} and \shareMonHigh{}, while V-DPM's stays between \shareVdpmMin{} and \shareVdpmMax{} across the range (supplementary Figure~5). On the real hands clips the camera reads the environment frame on \camHandsMinF{} to \camHandsMaxF{} of the clips, the mask on 0.23 to 0.95, so the mask is the output that carries the prior there; \cellSentence{}

\subsection{Training Versus Recalibration}
\label{sec:recipe}
% generated by scripts/paper_recipe_main.py (Session 11); cells in results/phase7/recipe_main.json, traced in paper/NUMBERS.md (F14); do not edit
\begin{table}[t]
  \centering
  % \scriptsize
  % \setlength{\tabcolsep}{0.8pt}
  \caption{The recipe on real video (intervals in supplementary Table~10). Hands: physical-frame rate on the 81 egocentric hands clips, released to after; the other columns: paired ATE/path change against the released checkpoint in percent, positive is worse, on the egocentric mover clips, the KITTI leading-vehicle clips, Sintel and TUM. Check: the row meets the pre-registered bar of its amendment, a hands gain with an interval excluding zero and no paired regression, of at least 45 points for the three rendered-subset rows (A17.2) and at least 15 points for the others. Control: dominant-mover sequences; held out: evaluated on the seven recordings not trained on; the rendered counts are moving-platform sequences.}
  \label{tab:recipe_main}
  \resizebox{\linewidth}{!}{
  \begin{tabular}{@{}lrrrrrc@{}}
    \toprule
    fine-tune & hands & mover & lead & Sintel & TUM & bar \\
    \midrule
    MoRe, 5 rendered & 0.31 to 0.94 & -3.1 & +2.5 & +1.1 & -3.9 & \checkmark \\
    MoRe, 25 rendered & 0.31 to 0.91 & -6.2 & -2.0 & -0.3 & -5.8 & \checkmark \\
    MoRe, 400 rendered & 0.31 to 0.91 & -6.7 & -1.3 & +0.8 & -2.9 & \checkmark \\
    MoRe, 400 control & 0.31 to 0.94 & -2.2 & +3.6 & +4.9 & +2.7 &  \\
    MoRe, 110 real (held out) & 0.13 to 1.00 & -5.0 & +11.4 & +4.7 & +0.1 &  \\
    VGGT4D head, 25 & 0.33 to 0.88 & +4.8 & +2.0 & +9.8 & +3.9 &  \\
    VGGT4D head, 25 control & 0.33 to 0.85 & +6.9 & -1.9 & +14.8 & +14.3 &  \\
    VGGT4D head, unfrozen & 0.33 to 0.42 & +24.8 & +6.7 & +76.9 & +15.4 &  \\
    MonST3R, 400 & 0.93 to 0.86 & +87.4 & +35.4 & +97.8 & +28.1 &  \\
    CUT3R, 400 & 0.85 to 0.25 & +82.5 & +20.7 & +42.1 & +69.5 &  \\
    \bottomrule
  \end{tabular}}
\end{table}

\begin{figure*}[t]
  \centering
  \includegraphics[width=\textwidth]{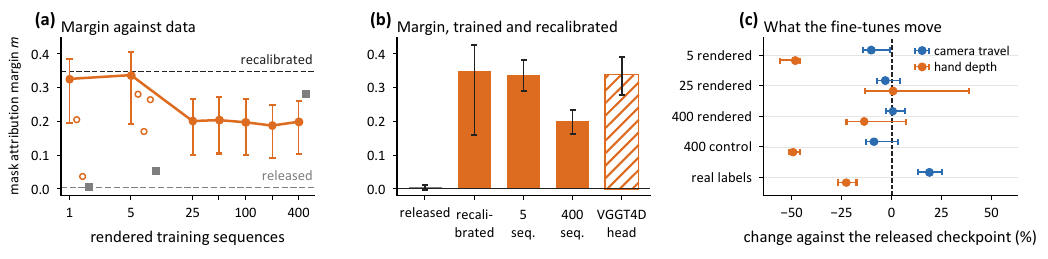}
  \caption{Training and the threshold (Amendments~17 to 22). (a) MoRe's mask attribution margin on the 81 real egocentric hands clips against the number of rendered sequences (recording-level intervals), with the five-sequence draws as open markers, the dominant-mover controls as squares and the released model recalibrated at the best threshold of the full sweep dashed. (b) The margin on the same clips: released, recalibrated, after five and 400 rendered sequences, and the VGGT4D head. (c) What the fine-tunes move: the camera's travel against the room and the predicted depth of the hands against the room, relative to the released checkpoint (recording-level intervals).}
  \label{fig:recipe}
\end{figure*}
If the mask's choice is a learned prior, training should move it. The recipe is a 3000-step fine-tune from the released weights with the authors' own training recipe on rendered moving-platform sequences whose labels mark the camera-attached foreground as the moving part (supplementary Section~L; a control uses dominant-mover sequences). The decisive test is real egocentric video with labels from the digital twin (\Cref{tab:recipe_main}, \Cref{fig:recipe}). On MoRe, five sequences raise the mask attribution margin on the 81 hands clips from 0.00 [$-0.01$, $+0.02$] to 0.34 [0.19, 0.41] and the physical-frame rate from 0.31 to 0.94 ($+63.0$ points [$+50.8$, $+85.4$]), with the mover mask IoU from 0.25 to 0.43 and the mover and leading-vehicle errors, Sintel, Bonn and TUM with no detected increase; 400 sequences give 0.20 and 0.91, and three further five-sequence draws replicate the rate in two of three, the third with the mover error $+5.1\%$ (A20.3). The dominant-mover control reaches 0.67 at five sequences and 0.94 at 400, so nothing in the gain is specific to moving-platform data: either rendered set teaches which side of \Cref{fig:teaser} is the world.

\paragraph{The Threshold.} The released MoRe already ranks the hands above the background and its threshold leaves them out of the mask. Sweeping that threshold over its full range (A20.2, A22.4) takes the physical-frame rate from 0.00 to 0.90, and at the best threshold for the margin it reaches 0.35 [0.16, 0.43] against the recipe's 0.34, a paired difference of $-0.01$ [$-0.05$, $+0.04$] over eight recordings; none of the four five-sequence draws exceeds it (A22.5), and a threshold selected on separate recordings reaches 0.35 on the held-out ones. By the margin the recipe is a transfer that recalibration matches. The mover mask is where training goes beyond a threshold: IoU 0.43 after five sequences against 0.30 at the margin's threshold ($+0.13$ [$+0.04$, $+0.21$] paired over 105 clips) and 0.34 at the threshold chosen for the mover IoU itself ($+0.09$ [$+0.06$, $+0.13$]). On MoRe the mask's prior on a moving platform is a calibration; training adds the mover (\Cref{fig:recipe}b).

\paragraph{The Camera Follows Neither.} The camera reads the environment frame on every hands clip, as released and under every fine-tune (\Cref{tab:agreement}, supplementary Table~13), the camera errors on the real mover and leading-vehicle clips move by less than 7\% under every rendered MoRe fine-tune (\Cref{tab:recipe_main}), and what the fine-tunes move is the predicted depth of the hands against the room, by up to half, while the camera's travel against the background stays within 19\% of the released checkpoint (Amendment~21, \Cref{fig:recipe}c). A head added to a model without one behaves the same way: a DPT head on VGGT4D's frozen aggregator, trained on 25 sequences (Amendment~19), gives it a mask with a hands margin of 0.34 [0.24, 0.47] and a rate of 0.88 from 0.33, while its camera label and error stay where they were (mover error $+4.8\%$). Fine-tunes whose geometry moved on the rendered sequences overfit the renderer (MonST3R, CUT3R, VGGT4D unfrozen; mover error up to $+87\%$), none moved the Axis~B degradation threshold, and a real-label variant repairs the held-out hands clips (0.13 to 1.00) at the price of the leading-vehicle clips ($+11.4\%$) and Sintel ($+4.7\%$) (supplementary Section~L).

\subsection{Real Video and Transfer}
\label{sec:realvideo}
On the 81 real hands clips (supplementary Section~K) the true static region lowers pooled ATE/path by $-6.7\%$ [$-19.7$, $+0.6$] over eight recordings and by $-14.6\%$ [$-26.8$, $-0.8$] in the top selection bin, the synthetic direction, and the ambiguous partition is 1.59 [1.29, 2.24] times harder than the rest for every model. On the 105 egocentric mover clips, all below the thresholds, the true static region gives $-6.0\%$ [$-14.8$, $+1.8$] where matched synthetic fractions give $+5.0\%$ [$+1.6$, $+8.6$]. On the 41 KITTI leading-vehicle clips the set's static region, the complement of the dilated moving-object boxes, changes ATE/path by $+2.1\%$ [$-3.8$, $+8.0$], the Axis~B sign with an interval that includes zero; the ambiguous partition is 3.11 [0.88, 6.59] times harder over all 83 clips and 1.33 [0.91, 1.65] over the 67 with the ego vehicle in motion. 

\subsection{Practical Guidance}
\label{sec:guidance}
Supplementary Table~12 reads \Cref{tab:leaderboard,tab:agreement} as advice, three rows per ambiguous situation (mask, camera, joint reconstruction) with the number that decides each. A model that leads on the mask's frame is not thereby the model for an accurate reconstruction: on a moving platform prefer V-DPM or PAGE-4D by the mask, Depth Anything~3, MoRe or $\pi^3$ by the camera, and $\pi^3$, Point4D or V-DPM by the foreground's trajectory, and recalibrate a released MoRe for the frame rather than retrain it; around a small mover prefer TraceAnything, MonST3R or V-DPM by the mask and MegaSaM, VGGT4D or Depth Anything~3 by the camera; under fast rotation UniQuery4R by both; under a large mover Depth Anything~3 or MegaSaM by the camera and 4RC by both, and no training repairs the threshold.

\subsection{Pre-Registered Outcomes}
\label{sec:prereg}
\looseness=1 Three hypotheses were registered with thresholds before rendering (supplementary Section~I): H1 and H2 fail as written; H3, that one of three pre-listed inference-time fixes cuts camera error by 25\% without regressing Sintel, was not tested because its kill criterion fired on validation; a pre-registered within-model ambiguity detector reached a pooled test AUROC of \detPooledAUROC{} against a bar of 0.75. The recipe, its controls and the audits of three external reviews were pre-registered with their readings and the wording for each outcome fixed in advance (Amendments~17 to 24). Twenty-five dated amendments changed a measurement, added an analysis or handled a clip that could not run as released; the original H1 to H3 thresholds were retained and later intervention criteria amended as documented, and a checkpoint evaluated once on the test split is not an untouched confirmation; the supplement separates the original hypotheses from the later follow-ups.

% \section{Limitations}
% The synthetic axes use rigid movers and an untextured mannequin; the matched control is rendered, and the real sets lack movers occupying over half the frame. The ambiguity score requires ground truth, while its inference-time proxy falls below the bar. Frame labels are scalar diagnostics validated on constructed cases, not world reconstructions. The recipe is limited to MoRe, the only architecture meeting the pre-registered bar, and camera misattribution remains unresolved without the true static-region input.

\section{Conclusion}
A monocular video does not uniquely reveal who moved, yet feed-forward 4D models resolve this ambiguity independently across their outputs. \bench makes this inconsistency measurable: under a dominant mover, the camera follows the object despite sufficient visual evidence to localize the camera; on a moving platform, the mask and camera select different worlds, revealing conflicting motion interpretations; and on MoRe, five labelled sequences validate and extend the mask's calibration to the mover in real-world settings, while the camera's motion attribution error remains unresolved.

\clearpage
% \section*{LLM Usage Statement}
% LLM-based tools were used solely to assist in editing and refining author-produced text, code, and figures. No LLM tools were used for literature review, experimental design or analysis, data preparation, or the development of research ideas and conclusions. Neither the full manuscript nor major portions were submitted to an LLM for reviewer-style feedback. All AI-assisted content was reviewed and verified by the authors, who take full responsibility for the manuscript.

{\small\bibliographystyle{ieeenat_fullname}\bibliography{main}}

@String(CVPR= {IEEE Conf. Comput. Vis. Pattern Recog.})

@String(ICCV= {Int. Conf. Comput. Vis.})

@String(ECCV= {Eur. Conf. Comput. Vis.})

@String(ICLR = {Int. Conf. Learn. Represent.})

@String(CVPRW= {IEEE Conf. Comput. Vis. Pattern Recog. Worksh.})

@String(CVPR  = {CVPR})

@String(ICCV  = {ICCV})

@String(ECCV  = {ECCV})

@String(ICLR  = {ICLR})

@String(CVPRW= {CVPRW})

@article{bescos2018dynaslam,
  title   = {{DynaSLAM}: Tracking, Mapping, and Inpainting in Dynamic Scenes},
  author  = {Bescos, Berta and F{\'a}cil, Jos{\'e} M. and Civera, Javier and Neira, Jos{\'e}},
  journal = {IEEE Robotics and Automation Letters (RA-L)},
  volume  = {3},
  number  = {4},
  pages   = {4076--4083},
  year    = {2018}
}

@inproceedings{bratulic2025geometry,
  title     = {On Geometric Understanding and Learned Priors in Feed-Forward {3D} Reconstruction Models},
  author    = {Bratuli{\'c}, Jelena and Mittal, Sudhanshu and Brox, Thomas and Rupprecht, Christian},
  booktitle = {European Conference on Computer Vision (ECCV)},
  pages     = {24--41},
  year      = {2026}
}

@inproceedings{butler2012sintel,
  title     = {A Naturalistic Open Source Movie for Optical Flow Evaluation},
  author    = {Butler, Daniel J. and Wulff, Jonas and Stanley, Garrett B. and Black, Michael J.},
  booktitle = {European Conference on Computer Vision (ECCV)},
  pages     = {611--625},
  year      = {2012}
}

@inproceedings{caesar2020nuscenes,
  title     = {{nuScenes}: A Multimodal Dataset for Autonomous Driving},
  author    = {Caesar, Holger and Bankiti, Varun and Lang, Alex H. and Vora, Sourabh and Liong, Venice Erin and Xu, Qiang and Krishnan, Anush and Pan, Yu and Baldan, Giancarlo and Beijbom, Oscar},
  booktitle = {IEEE/CVF Conference on Computer Vision and Pattern Recognition (CVPR)},
  pages     = {11621--11631},
  year      = {2020}
}

@article{uniquery4r2026,
  title   = {{UniQuery4R}: Unified {4D} Scene Reconstruction from a Single Query},
  author  = {Chen, Tiancheng and Tang, Sheng and Jin, Wenhua and Zhang, Weiqi and Fang, Juntong and Zhou, Junsheng and Li, Zesong},
  journal = {arXiv preprint arXiv:2608.17283},
  year    = {2026}
}

@inproceedings{chen2024leapvo,
  title     = {{LEAP-VO}: Long-Term Effective Any Point Tracking for Visual Odometry},
  author    = {Chen, Weirong and Chen, Le and Wang, Rui and Pollefeys, Marc},
  booktitle = {IEEE/CVF Conference on Computer Vision and Pattern Recognition (CVPR)},
  pages     = {19844--19853},
  year      = {2024}
}

@inproceedings{chen2025easi3r,
  title     = {{Easi3R}: Estimating Disentangled Motion from {DUSt3R} Without Training},
  author    = {Chen, Xingyu and Chen, Yue and Xiu, Yuliang and Geiger, Andreas and Chen, Anpei},
  booktitle = {IEEE/CVF International Conference on Computer Vision (ICCV)},
  pages     = {9158--9168},
  year      = {2025}
}

@inproceedings{more2025,
  title     = {{MoRe}: Motion-Aware Feed-Forward {4D} Reconstruction Transformer},
  author    = {Fang, Juntong and Chen, Zequn and Zhang, Weiqi and Di, Donglin and Zhang, Xuancheng and Yang, Chengmin and Liu, Yu-Shen},
  booktitle = {IEEE/CVF Conference on Computer Vision and Pattern Recognition (CVPR)},
  year      = {2026}
}

@inproceedings{gao2022dycheck,
  title     = {Monocular Dynamic View Synthesis: A Reality Check},
  author    = {Gao, Hang and Li, Ruilong and Tulsiani, Shubham and Russell, Bryan and Kanazawa, Angjoo},
  booktitle = {Advances in Neural Information Processing Systems (NeurIPS)},
  pages     = {33768--33780},
  year      = {2022}
}

@inproceedings{geiger2012kitti,
  title     = {Are We Ready for Autonomous Driving? {The} {KITTI} Vision Benchmark Suite},
  author    = {Geiger, Andreas and Lenz, Philip and Urtasun, Raquel},
  booktitle = {IEEE Conference on Computer Vision and Pattern Recognition (CVPR)},
  pages     = {3354--3361},
  year      = {2012}
}

@inproceedings{greff2022kubric,
  title     = {Kubric: A Scalable Dataset Generator},
  author    = {Greff, Klaus and Belletti, Francois and Beyer, Lucas and Doersch, Carl and Du, Yilun and Duckworth, Daniel and Fleet, David J. and Gnanapragasam, Dan and Golemo, Florian and Herrmann, Charles and others},
  booktitle = {IEEE/CVF Conference on Computer Vision and Pattern Recognition (CVPR)},
  pages     = {3749--3761},
  year      = {2022}
}

@inproceedings{vggt4d2025,
  title     = {{VGGT4D}: Mining Motion Cues in Visual Geometry Transformers for {4D} Scene Reconstruction},
  author    = {Hu, Yu and Cheng, Chong and Yu, Sicheng and Guo, Xiaoyang and Wang, Hao},
  booktitle = {IEEE/CVF Conference on Computer Vision and Pattern Recognition Findings (CVPR Findings)},
  pages     = {414--424},
  year      = {2026}
}

@inproceedings{point4d2026,
  title     = {{Point4D}: Long-Range {4D} Motion Reconstruction},
  author    = {Jeon, Minsik and Karhade, Jay and Ramanan, Deva and Tulsiani, Shubham},
  booktitle = {Advances in Neural Information Processing Systems (NeurIPS)},
  year      = {2026}
}

@inproceedings{karaev2024cotracker,
  title     = {{CoTracker}: It Is Better to Track Together},
  author    = {Karaev, Nikita and Rocco, Ignacio and Graham, Benjamin and Neverova, Natalia and Vedaldi, Andrea and Rupprecht, Christian},
  booktitle = {European Conference on Computer Vision (ECCV)},
  pages     = {18--35},
  year      = {2024}
}

@inproceedings{any4d2025,
  title     = {{Any4D}: Unified Feed-Forward Metric {4D} Reconstruction},
  author    = {Karhade, Jay and Keetha, Nikhil and Zhang, Yuchen and Gupta, Tanisha and Sharma, Akash and Scherer, Sebastian and Ramanan, Deva},
  booktitle = {IEEE/CVF Conference on Computer Vision and Pattern Recognition (CVPR)},
  pages     = {14578--14589},
  year      = {2026}
}

@inproceedings{kopf2021rcvd,
  title     = {Robust Consistent Video Depth Estimation},
  author    = {Kopf, Johannes and Rong, Xuejian and Huang, Jia-Bin},
  booktitle = {IEEE/CVF Conference on Computer Vision and Pattern Recognition (CVPR)},
  pages     = {1611--1621},
  year      = {2021}
}

@inproceedings{leroy2024mast3r,
  title     = {Grounding Image Matching in {3D} with {MASt3R}},
  author    = {Leroy, Vincent and Cabon, Yohann and Revaud, Jerome},
  booktitle = {European Conference on Computer Vision (ECCV)},
  pages     = {71--91},
  year      = {2024}
}

@inproceedings{li2025megasam,
  title     = {{MegaSaM}: Accurate, Fast, and Robust Structure and Motion from Casual Dynamic Videos},
  author    = {Li, Zhengqi and Tucker, Richard and Cole, Forrester and Wang, Qianqian and Jin, Linyi and Ye, Vickie and Kanazawa, Angjoo and Holynski, Aleksander and Snavely, Noah},
  booktitle = {IEEE/CVF Conference on Computer Vision and Pattern Recognition (CVPR)},
  pages     = {10486--10496},
  year      = {2025}
}

@inproceedings{da3_2025,
  title     = {{Depth Anything 3}: Recovering the Visual Space from Any Views},
  author    = {Lin, Haotong and Chen, Sili and Liew, Jun Hao and Chen, Donny Y. and Li, Zhenyu and Zhao, Yang and Peng, Sida and Guo, Hengkai and Zhou, Xiaowei and Shi, Guang and others},
  booktitle = {International Conference on Learning Representations (ICLR)},
  pages     = {141261--141285},
  year      = {2026}
}

@inproceedings{traceanything2025,
  title     = {{Trace Anything}: Representing Any Video in {4D} via Trajectory Fields},
  author    = {Liu, Xinhang and Xiao, Yuxi and Chen, Donny Y. and Feng, Jiashi and Tai, Yu-Wing and Tang, Chi-Keung and Kang, Bingyi},
  booktitle = {International Conference on Learning Representations (ICLR)},
  pages     = {37738--37767},
  year      = {2026}
}

@inproceedings{fourrc2025,
  title     = {{4RC}: {4D} Reconstruction via Conditional Querying Anytime and Anywhere},
  author    = {Luo, Yihang and Zhou, Shangchen and Lan, Yushi and Pan, Xingang and Loy, Chen Change},
  booktitle = {International Conference on Machine Learning (ICML)},
  pages     = {83198--83214},
  year      = {2026}
}

@article{nosek2018preregistration,
  title   = {The Preregistration Revolution},
  author  = {Nosek, Brian A. and Ebersole, Charles R. and DeHaven, Alexander C. and Mellor, David T.},
  journal = {Proceedings of the National Academy of Sciences (PNAS)},
  volume  = {115},
  number  = {11},
  pages   = {2600--2606},
  year    = {2018}
}

@inproceedings{palazzolo2019bonn,
  title     = {{ReFusion}: {3D} Reconstruction in Dynamic Environments for {RGB-D} Cameras Exploiting Residuals},
  author    = {Palazzolo, Emanuele and Behley, Jens and Lottes, Philipp and Gigu{\`e}re, Philippe and Stachniss, Cyrill},
  booktitle = {IEEE/RSJ International Conference on Intelligent Robots and Systems (IROS)},
  pages     = {7855--7862},
  year      = {2019}
}

@inproceedings{pan2023adt,
  title     = {{Aria Digital Twin}: A New Benchmark Dataset for Egocentric {3D} Machine Perception},
  author    = {Pan, Xiaqing and Charron, Nicholas and Yang, Yongqian and Peters, Scott and Whelan, Thomas and Kong, Chen and Parkhi, Omkar and Newcombe, Richard and Ren, Yuheng Carl},
  booktitle = {IEEE/CVF International Conference on Computer Vision (ICCV)},
  pages     = {20133--20143},
  year      = {2023}
}

@inproceedings{sturm2012tum,
  title     = {A Benchmark for the Evaluation of {RGB-D} {SLAM} Systems},
  author    = {Sturm, J{\"u}rgen and Engelhard, Nikolas and Endres, Felix and Burgard, Wolfram and Cremers, Daniel},
  booktitle = {IEEE/RSJ International Conference on Intelligent Robots and Systems (IROS)},
  pages     = {573--580},
  year      = {2012}
}

@inproceedings{vdpm2025,
  title     = {{V-DPM}: {4D} Video Reconstruction with Dynamic Point Maps},
  author    = {Sucar, Edgar and Insafutdinov, Eldar and Lai, Zihang and Vedaldi, Andrea},
  booktitle = {IEEE/CVF Conference on Computer Vision and Pattern Recognition (CVPR)},
  pages     = {14502--14511},
  year      = {2026}
}

@inproceedings{teed2021droid,
  title     = {{DROID-SLAM}: Deep Visual {SLAM} for Monocular, Stereo, and {RGB-D} Cameras},
  author    = {Teed, Zachary and Deng, Jia},
  booktitle = {Advances in Neural Information Processing Systems (NeurIPS)},
  pages     = {16558--16569},
  year      = {2021}
}

@inproceedings{tsui2026stady4d,
  title     = {{StaDy4D}: Towards Complete {4D} Static-Dynamic Reconstruction with {SIGMA}},
  author    = {Tsui, Hao-Tang and Tuan, Yu-Rou and Lai, Ethan and Wang, Chen-Yu},
  booktitle = {IEEE/CVF Conference on Computer Vision and Pattern Recognition Workshops (CVPRW)},
  pages     = {4697--4707},
  year      = {2026}
}

@article{umeyama1991least,
  title   = {Least-Squares Estimation of Transformation Parameters Between Two Point Patterns},
  author  = {Umeyama, Shinji},
  journal = {IEEE Transactions on Pattern Analysis and Machine Intelligence (TPAMI)},
  volume  = {13},
  number  = {4},
  pages   = {376--380},
  year    = {1991}
}

@article{wallach1953kinetic,
  title   = {The Kinetic Depth Effect},
  author  = {Wallach, Hans and O'Connell, D. N.},
  journal = {Journal of Experimental Psychology},
  volume  = {45},
  number  = {4},
  pages   = {205--217},
  year    = {1953}
}

@inproceedings{wang2025vggt,
  title     = {{VGGT}: Visual Geometry Grounded Transformer},
  author    = {Wang, Jianyuan and Chen, Minghao and Karaev, Nikita and Vedaldi, Andrea and Rupprecht, Christian and Novotny, David},
  booktitle = {IEEE/CVF Conference on Computer Vision and Pattern Recognition (CVPR)},
  pages     = {5294--5306},
  year      = {2025}
}

@inproceedings{wang2024shapeofmotion,
  title     = {{Shape of Motion}: {4D} Reconstruction from a Single Video},
  author    = {Wang, Qianqian and Ye, Vickie and Gao, Hang and Zeng, Weijia and Austin, Jake and Li, Zhengqi and Kanazawa, Angjoo},
  booktitle = {IEEE/CVF International Conference on Computer Vision (ICCV)},
  pages     = {9660--9672},
  year      = {2025}
}

@inproceedings{wang2025cut3r,
  title     = {Continuous {3D} Perception Model with Persistent State},
  author    = {Wang, Qianqian and Zhang, Yifei and Holynski, Aleksander and Efros, Alexei A. and Kanazawa, Angjoo},
  booktitle = {IEEE/CVF Conference on Computer Vision and Pattern Recognition (CVPR)},
  pages     = {10510--10522},
  year      = {2025}
}

@inproceedings{wang2024dust3r,
  title     = {{DUSt3R}: Geometric {3D} Vision Made Easy},
  author    = {Wang, Shuzhe and Leroy, Vincent and Cabon, Yohann and Chidlovskii, Boris and Revaud, Jerome},
  booktitle = {IEEE/CVF Conference on Computer Vision and Pattern Recognition (CVPR)},
  pages     = {20697--20709},
  year      = {2024}
}

@inproceedings{wang2025pi3,
  title     = {{$\pi^3$}: Permutation-Equivariant Visual Geometry Learning},
  author    = {Wang, Yifan and Zhou, Jianjun and Zhu, Haoyi and Chang, Wenzheng and Zhou, Yang and Li, Zizun and Chen, Junyi and Pang, Jiangmiao and Shen, Chunhua and He, Tong},
  booktitle = {International Conference on Learning Representations (ICLR)},
  pages     = {10481--10497},
  year      = {2026}
}

@inproceedings{d4rt2025,
  title     = {Efficiently Reconstructing Dynamic Scenes One {D4RT} at a Time},
  author    = {Zhang, Chuhan and Le Moing, Guillaume and Koppula, Skanda and Rocco, Ignacio and Momeni, Liliane and Xie, Junyu and Sun, Shuyang and Sukthankar, Rahul and Barral, Jo{\"e}lle K. and Hadsell, Raia and others},
  booktitle = {IEEE/CVF Conference on Computer Vision and Pattern Recognition (CVPR)},
  pages     = {7382--7392},
  year      = {2026}
}

@inproceedings{zhang2025monst3r,
  title     = {{MonST3R}: A Simple Approach for Estimating Geometry in the Presence of Motion},
  author    = {Zhang, Junyi and Herrmann, Charles and Hur, Junhwa and Jampani, Varun and Darrell, Trevor and Cole, Forrester and Sun, Deqing and Yang, Ming-Hsuan},
  booktitle = {International Conference on Learning Representations (ICLR)},
  pages     = {82863--82886},
  year      = {2025}
}

@inproceedings{zhang2022casualsam,
  title     = {Structure and Motion from Casual Videos},
  author    = {Zhang, Zhoutong and Cole, Forrester and Li, Zhengqi and Rubinstein, Michael and Snavely, Noah and Freeman, William T.},
  booktitle = {European Conference on Computer Vision (ECCV)},
  pages     = {20--37},
  year      = {2022}
}

@inproceedings{zhao2022particlesfm,
  title     = {{ParticleSfM}: Exploiting Dense Point Trajectories for Localizing Moving Cameras in the Wild},
  author    = {Zhao, Wang and Liu, Shaohui and Guo, Hengkai and Wang, Wenping and Liu, Yong-Jin},
  booktitle = {European Conference on Computer Vision (ECCV)},
  pages     = {523--542},
  year      = {2022}
}

@inproceedings{zheng2023pointodyssey,
  title     = {{PointOdyssey}: A Large-Scale Synthetic Dataset for Long-Term Point Tracking},
  author    = {Zheng, Yang and Harley, Adam W. and Shen, Bokui and Wetzstein, Gordon and Guibas, Leonidas J.},
  booktitle = {IEEE/CVF International Conference on Computer Vision (ICCV)},
  pages     = {19855--19865},
  year      = {2023}
}

@inproceedings{page4d2026,
  title     = {{PAGE-4D}: {VGGT-4D} Perception via Disentangled Pose and Geometry Estimation},
  author    = {Zhou, Kaichen and Wang, Yuhan and Chen, Grace and Beaudouin, Gaspard and Zhan, Fangneng and Liang, Paul Pu and Wang, Mengyu},
  booktitle = {International Conference on Learning Representations (ICLR)},
  pages     = {36401--36414},
  year      = {2026}
}

\clearpage
\appendix

\section{Reproducibility, Licences and Consent}
\label{app:repro}
\paragraph{Reproducibility.} The study was pre-registered in a protocol file fixed before the first sequence was rendered (hypotheses, thresholds, metrics, the three fix candidates, the kill criteria); every deviation is a dated amendment in the same file, listed in \cref{app:prereg}. Every evaluation run is logged with its configuration, commit hash and metrics; every number in this paper is traced to a results file in the released \texttt{NUMBERS.md}, and a \texttt{make check} target re-derives that file from the results. Models were evaluated at pinned commits with the authors' default settings; environment lock files and the install scripts, with the few one-line compatibility patches they apply, are released. The test split of the benchmark was opened for the pre-registered detector evaluation and for the single evaluations of the fine-tunes of Amendments~11, 12, 13, 17, 18, 19 and 20, and is released as the held-out split. Each of these was one evaluation per pre-registered reading; a checkpoint evaluated once on the test split is not an untouched confirmation. \Cref{app:prereg} separates the original hypotheses (H1 to H3) and the sanity check (H4), the validation-driven choices (the derived-mask thresholds, the fix candidates, the recipe's data sizes), the prospectively specified follow-ups (Amendments~10 to 22 and 24), the camera label of Amendment~23 and the measurement audit of Amendment~25. The benchmark description, the test-split sequence list, the evaluation script and output format, and a leaderboard regenerated from the results by \texttt{make site} are released upon publication as a benchmark page in the repository; the leaderboard table and the test-split list are reproduced in \cref{app:release} as they stand.

\paragraph{Licences and Consent.} The synthetic sequences use assets under permissive licences (CC BY 4.0 and CC0); the people in them are untextured base meshes. The real driving set is derived from KITTI under its licence and released as clip indices, not frames; the egocentric set is derived from Aria Digital Twin under its research licence and released as clip indices; web clips in the qualitative figure are referenced by public video identifier and timestamp, frames are not redistributed, and an opt-out contact is provided for clip owners. Faces in the pictured frames and in the supplementary video are blurred. No personal data beyond what the public clips show is collected and no downstream application beyond reconstruction evaluation is pursued. Existing benchmarks are used under their own licences and not redistributed.

\section{Metrics}
\label{app:metrics}
\paragraph{Camera Error.} Predicted camera-to-world trajectories are aligned to the ground truth with a Sim(3) Umeyama fit~\cite{umeyama1991least}; ATE is the root-mean-square position error after alignment, RPE is computed over 5-frame windows. Because the renderer reaches a pixel fraction $f$ by moving the camera closer and scales the camera path about the mover, the ground-truth path length shrinks with $f$ (1.00\,m at $f=0.10$ against 0.11\,m at $f=0.70$ in the pilot set), so absolute ATE falls with $f$ for every model regardless of any prior. The primary quantity is therefore ATE divided by the ground-truth camera path length (ATE/path); absolute ATE is kept as a secondary column (Amendment~2).

\paragraph{Camera-Attribution Ratio.} On Axis~A no decomposition is correct, and alignment to a ground-truth trajectory fits the scale to that trajectory, which makes the pre-registered ratio of path lengths close to 1 by construction (measured 1.31 to 1.48 for three models on 24 check sequences). The ratio is therefore computed scale-free within the model's own output: $r_\mathrm{cam} = (\text{estimated camera travel}/s)\,/\,\text{true relative travel}$, where $s$ is the pixel-wise median of predicted over true depth on the placed objects' pixels of the first frame (Amendment~4). The dynamic-pixel fraction on these sequences, where nothing moves relative to the world, is reported next to it as world attribution.

\paragraph{Thresholds.} The pose degradation threshold $f_\mathrm{deg}$ is the smallest $f$ at which the mean ATE/path exceeds three times its value at $f=0.10$ for that model and class (the pre-registered relative reading, called $f_\mathrm{pose}$ in the protocol) or exceeds 0.10 (the absolute reading of Amendment~6.3); $f_\mathrm{mask}$ is the smallest $f$ at which the mean mask IoU falls below half its $f=0.10$ value or below 0.20. Intervals come from 1000 bootstrap resamples of the sequences within each $f$ level, recomputing the threshold each time; the fraction of resamples without a crossing is reported as $P(\text{none})$. Amendment~9.2 renames the pose quantity from flip point to degradation threshold in the paper's prose because the pose head does not change its answer; the pre-registration record keeps the original wording and verdict. The main text likewise calls $f_\mathrm{mask}$ the mask degradation point, because Amendment~20 (A20.1, \cref{app:a20}) finds that no mask reverses at it; this supplement keeps the protocol's name, flip point, for the same quantity.

\paragraph{Frame Choice.} On Axis~C a sequence is read as physical when the model's dynamic mask covers the camera-attached foreground more than the moving-platform background, and egocentric otherwise; ATE/path is against the physical camera path and mask IoU against the physical labelling.

\paragraph{Realised Lever Values (Amendment~25).} Every lever is set by the generator and measured on the ground-truth masks, and the released index carries the realised share of each sequence. On Axis~C the realised foreground share is within \leverCMaxDev{} of the nominal $g$ in eighteen of the twenty platform-by-$g$ cells (\cref{tab:levers}). The escalator's camera-attached geometry reaches a third of the frame, so its sixteen sequences at nominal $g=0.5$ and $0.7$ realise \leverEscHalf{} and \leverEscSeven{}. The Axis~C rates of the main text are over all 160 sequences and do not condition on $g$. \Cref{fig:axisc} plots against the realised share, with the sixteen sequences in the 0.3 bin. Under the nominal binning, as pre-registered, a model's rate at 0.5 or 0.7 differs by up to \leverSensMask{} for the mask and \leverSensCam{} for the camera, and \leverSideChanges{} model-by-bin rates fall on the other side of one half; both binnings are in \texttt{results/phase10/levers/readings.json} and in the released \texttt{axis\_c.json}. On Axes~B and B$'$ the sequence mean of the mover and sphere share is within \leverBMaxDev{} and \leverBMMaxDev{} of the nominal $f$ in every cell.
% generated by scripts/phase10_levers.py (Amendment 25, A25.1); traced in paper/NUMBERS.md (F17); do not edit
\begin{table}[t]
  \centering
  \footnotesize
  \setlength{\tabcolsep}{5pt}
  \caption{Realised camera-attached foreground share on Axis~C by platform and nominal $g$ (mean over the eight sequences of a cell; Amendment~25). The escalator saturates.}
  \label{tab:levers}
  \begin{tabular}{lrrrr}
    \toprule
    platform & $g=0.1$ & $g=0.3$ & $g=0.5$ & $g=0.7$ \\
    \midrule
    train & 0.124 & 0.298 & 0.520 & 0.703 \\
    boat & 0.088 & 0.285 & 0.491 & 0.679 \\
    escalator & 0.094 & 0.288 & \textbf{0.318} & \textbf{0.320} \\
    elevator & 0.119 & 0.312 & 0.515 & 0.698 \\
    treadmill & 0.119 & 0.289 & 0.507 & 0.682 \\
    \bottomrule
  \end{tabular}
\end{table}

\paragraph{Derived Masks.} Models without a dynamic head get a mask from the residual of their predicted 3D motion (4RC, Any4D, V-DPM: motion magnitude over median depth, confidence-weighted where selected) or depth (CUT3R, Depth Anything~3, $\pi^3$: depth-reprojection or reprojection residual over a 5-frame window) after rigid alignment; the method and threshold per model were chosen in one round on the validation split (195 attempts logged) and frozen for the rest of the study (Amendment~3.5). Their validation IoU at $f\le0.10$ is 0.72 (4RC), 0.55 (Any4D), 0.26 (CUT3R), 0.31 (Depth Anything~3), 0.37 ($\pi^3$) and 0.77 (V-DPM).

\begin{table*}[t]
\centering
\footnotesize
\caption{Model zoo: fifteen feed-forward or optimisation-based 4D reconstruction models, evaluated as released (pinned commit, authors' default settings and recommended resolution). Derived masks use the residual method frozen on the validation split (A3.5). Training-data abbreviations used in the supplementary Table 4: PO PointOdyssey, DR Dynamic Replica, TA TartanAir, Spr Spring, Way Waymo, Kub Kubric, VK Virtual KITTI, SN++ ScanNet++, OWG OmniWorld-Game, St4D Stereo4D, BED BEDLAM2. UniQuery4R, Point4D and PAGE-4D were missed by the Phase 0 inventory and added under Amendment 16 after the hypotheses were decided on the first twelve; PAGE-4D's released training configuration includes the PointOdyssey test split, so its PointOdyssey numbers are on its own training scenes.}
\label{tab:zoo}
\resizebox{\textwidth}{!}{\begin{tabular}{lllll}
\toprule
Model & Commit & Outputs & Native res. & Dynamic mask \\
\midrule
MonST3R & \texttt{574cc77ad2} & poses, depth, dyn\_mask & 512$\times$208 & own head \\
CUT3R & \texttt{8bc15dc92a} & poses, depth & 512$\times$208 & derived (depth) \\
Easi3R & \texttt{146a77d77c} & poses, depth, dyn\_mask & 512$\times$208 & own head \\
VGGT4D & \texttt{c7af95a436} & poses, depth, dyn\_mask & 518$\times$224 & own head \\
$\pi^3$ & \texttt{9fa3ddb3f8} & poses, depth & 770$\times$322 & derived (reproj3d\_conf) \\
Any4D & \texttt{aa9f1b0d7e} & poses, depth, motion3d & 518$\times$252 & derived (motion3d) \\
TraceAnything & \texttt{54677b5e7b} & depth, dyn\_mask, motion3d (no cameras) & 512$\times$208 & own head \\
MegaSaM & \texttt{a27b4e633c} & poses, depth, dyn\_mask & 672$\times$288 & own head \\
Depth Anything 3 & \texttt{3d835ec1a5} & poses, depth & 504$\times$210 & derived (depth) \\
4RC & \texttt{a08ca0a539} & poses, depth, motion3d & 504$\times$210 & derived (motion3d\_conf) \\
V-DPM & \texttt{5e2a57cf60} & poses, depth, motion3d & 518$\times$224 & derived (motion3d\_conf) \\
MoRe & \texttt{f330a3499d} & poses, depth, dyn\_mask & 518$\times$224 & own head \\
UniQuery4R & \texttt{95119742e6} & poses, depth, dyn\_mask & 504$\times$280 & own head \\
Point4D & \texttt{4da8153711} & poses, depth, motion3d & 518$\times$294 & derived (motion3d+reproj3d) \\
PAGE-4D & \texttt{b8af462a13} & poses, depth & 518$\times$294 & derived (reproj3d\_conf) \\
\bottomrule
\end{tabular}}
\end{table*}

\begin{table*}[t]
\centering
\footnotesize
\caption{WhoMoved-Bench: the four synthetic axes (30 frames, 512$\times$288, exact ground truth), the dropped real-world panel and the ambiguity-partitioned existing benchmarks. The test split is held out; it was opened for the pre-registered detector evaluation (Amendment~7) and for the single evaluations of the fine-tunes of Amendments~11, 12, 13, 17, 18, 19 and 20, each listed in the supplement.}
\label{tab:bench}
\resizebox{\textwidth}{!}{\begin{tabular}{lp{5.2cm}rrp{4.0cm}}
\toprule
Part & Design & Sequences & Val / test & Ground truth \\
\midrule
A, rigid world & 32 cue combinations (resolution-IV) $\times$ 4 scenes $\times$ 2 motions & 256 & 128 / 128 & recorded generating decomposition \\
B, dominant mover & 11 $f$ $\times$ 4 object classes $\times$ 8 seeds & 352 & 176 / 176 & cameras, depth, masks, 3D motion \\
C, moving background & 4 $g$ $\times$ 5 platforms $\times$ 8 seeds & 160 & 80 / 80 & physical decomposition \\
D, rotation-dominant & 60 flight trajectories $\times$ 4 scenes $\times$ \{static, mover\} & 480 & 240 / 240 & cameras, depth, masks, 3D motion \\
Real-world panel (dropped, Amendment~9) & 6 situations $\times$ 15 labelled clips (A6.1), reduced by A9.4 and dropped with four of forty labels in place & 90 planned, 0 used & n/a & human static-region mask, frame choice (not collected) \\
Existing benchmarks & Sintel, Bonn, TUM, DyCheck, MOVi-F, PointOdyssey & 186 & n/a & own GT + ambiguity partition \\
\bottomrule
\end{tabular}}
\end{table*}

\begin{table*}[t]
\centering
\scriptsize
\setlength{\tabcolsep}{2pt}
\caption{Model properties next to the measured quantities. Family: FF = feed-forward pointmap, align = global alignment, OPT = optimisation-based. Dynamic supervision at training from the papers (abbreviations in Table~\ref{tab:zoo}; sources in the released \texttt{model\_properties.json}); mask source as used here; runtime is the median over the Axis B sequences (30 frames, one A6000); $f_\mathrm{mask}$ and $f_\mathrm{deg}$ pooled over classes at sixteen seeds per cell (none = no crossing up to $f=0.95$); Axis C physical-frame rate over 160 sequences.}
\label{tab:properties}
\resizebox{\textwidth}{!}{\begin{tabular}{llllrrrr}
\toprule
Model & Family & Dynamic supervision (data) & Mask source & s/clip & $f_\mathrm{mask}$ & $f_\mathrm{deg}$ & Axis C phys. \\
\midrule
MonST3R & FF + align & yes: PO, TA, Spr, Way & flow residual & 110 & 0.50 & none & 0.62 \\
Easi3R & FF + align (train.-free) & no: DUSt3R weights & attention & 122 & none & none & 0.39 \\
MegaSaM & OPT (BA + depth prior) & yes: Kub dyn.; TA, Kub static & learned head & 77 & 0.60 & 0.70 & 0.37 \\
VGGT4D & FF (training-free, VGGT) & partly: VGGT: PO, Kub, VK + static & attention & 35 & 0.30 & 0.60 & 0.16 \\
MoRe & FF + motion attention & yes: DR, PO, Spr, VK, TA, Way, OWG & learned head & 2 & 0.80 & none & 0.21 \\
UniQuery4R & FF + motion (query-cond.) & yes: Kub, PO, St4D, VK, Way + internal & own (scene-flow thr.) & 16 & 0.20 & none & 0.41 \\
Point4D & FF + 3D tracks & yes: PO, DR, BED, Kub, Way, OWG + static & derived (motion) & 45 & none & none & 0.47 \\
PAGE-4D & FF (VGGT ft., masked aggr.) & yes: PO (incl. test), DR, Kub, Spr, Way + Co3D & derived (depth) & 5 & 0.90 & none & 0.75 \\
CUT3R & FF, recurrent state & yes: PO, DR, Spr, TA, Way + 27 sets & derived (depth) & 4 & none & none & 0.42 \\
$\pi^3$ & FF, permutation-equiv. & yes: internal dyn. + 14 static & derived (reproj.) & 4 & 0.90 & none & 0.65 \\
Depth Anything 3 & FF depth-ray & no: static (PO for geometry) & derived (depth) & 5 & 0.80 & none & 0.57 \\
Any4D & FF + motion & yes: PO, DR, Kub; VK2, Way & derived (motion) & 36 & none & none & 0.70 \\
4RC & FF + motion & yes: PO, DR, Way, Kub + static & derived (motion) & 39 & none & none & 0.78 \\
V-DPM & FF + motion & yes: Kub, PO, Way + static & derived (motion) & 71 & none & 0.60 & 0.74 \\
TraceAnything & FF trajectory field & yes: Kub 20k, own platform 10k & own (traj. var.) & 3 & none & none & 0.63 \\
\bottomrule
\end{tabular}}
\end{table*}

\section{Sanity Check (H4)}
\label{app:h4}
On Axis~B at $f\le0.10$ (64 sequences per model) every pose model's ATE/path is within 1.5 times its own Sintel ATE/path measured by the same code path on the 14-sequence MonST3R-protocol subset, so the rendering is adequate on this low-coverage control for the camera heads. The mask clause (IoU $\ge0.6$) holds for 4RC, MegaSaM, MonST3R, MoRe, V-DPM and VGGT4D and fails for Any4D, CUT3R, Depth Anything~3 and $\pi^3$ (derived masks) and Easi3R (own head, 0.55); their mask results carry the pre-registered caveat. The published Sintel numbers of the models are in different units and protocols and could not serve as the reference (Amendment~1); they are recorded next to the measured ones.

\section{Axis B: Curves and Thresholds}
\label{app:sens}
\begin{figure*}[t]
  \centering
  \includegraphics[width=\textwidth]{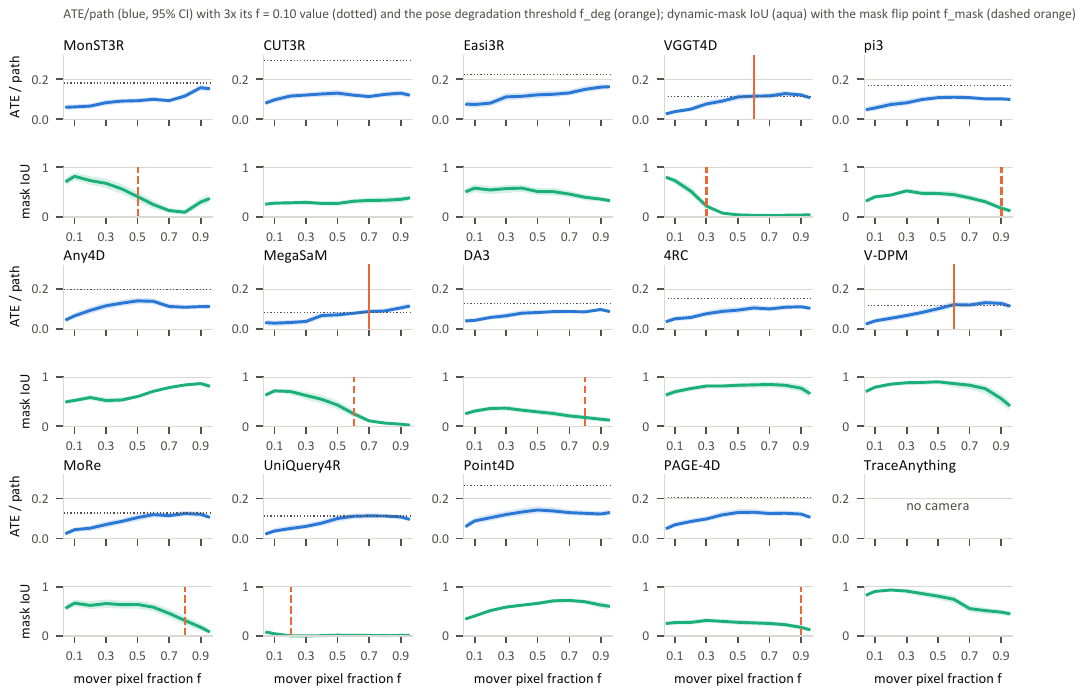}
  \caption{Axis~B curves. Camera ATE/path (blue, 95\% CI) with three times its $f=0.10$ value (dotted) and the pose degradation threshold $f_\mathrm{deg}$ (orange); dynamic-mask IoU (aqua) with the mask flip point $f_\mathrm{mask}$ (dashed orange). 64 sequences per point (sixteen seeds, four classes).}
  \label{fig:axisb}
\end{figure*}

\begin{figure}[t]
  \centering
  \includegraphics[width=\linewidth]{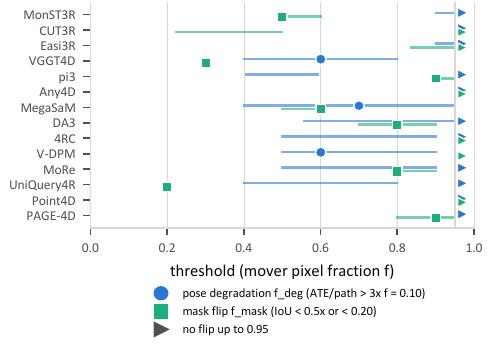}
  \caption{Pose degradation threshold (relative reading) and mask flip point per model with bootstrap intervals; a triangle at the right edge marks none up to $f=0.95$.}
  \label{fig:flips}
\end{figure}

\begin{figure}[t]
  \centering
  \includegraphics[width=\linewidth]{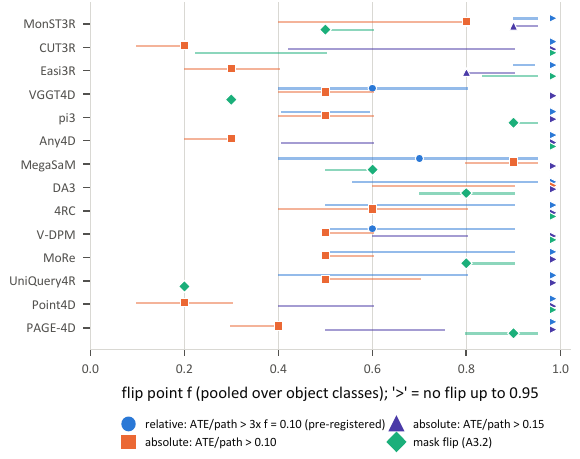}
  \caption{Thresholds per model under the pre-registered relative criterion, two absolute criteria and the mask criterion, pooled over classes, with bootstrap intervals; triangles mark none up to $f=0.95$.}
  \label{fig:sens}
\end{figure}

Every Axis~B value in this section uses sixteen scene seeds per cell: the eight of the pre-registered pass and eight added under Amendment~13 (Gate~6a of \cref{app:prereg}; 352 further sequences, 10554 evaluation runs over both batches and the fifteen models, six deterministic failures). The intervals narrow for most models (mask flip point 7 of 10, relative threshold 10 of 10, absolute threshold 9 of 11) and the pooled point estimates move for two mask flip points (V-DPM and TraceAnything from 0.95 to none) and four relative thresholds (MonST3R and MoRe to none, VGGT4D 0.50 to 0.60, MegaSaM 0.40 to 0.70); the eight-seed values stand in the pre-registration record and the verdicts of \cref{app:prereg} are not re-decided. \Cref{fig:axisb} plots ATE/path and mask IoU against $f$ for every model, \cref{fig:flips} the thresholds per model and \cref{tab:flips} the thresholds per class. Amendment~20 (A20.1) labels each mask flip point by the signed attribution margin $m$ (the mean mask value on the mover minus that on the background) and the mask's foreground false-negative and background false-positive rates: of the seven own-mask models, five have a flip point, four of them partial masks whose margin keeps its sign while the IoU criterion fires (MegaSaM, MonST3R, MoRe, VGGT4D) and one a collapse, the mask vanishing (UniQuery4R); none is a reversal, a margin that changes sign; Easi3R and TraceAnything have no flip below $f=0.95$ under either rule (\texttt{results/phase9/attribution/readings.json}). The relative criterion saturates when the $f=0.10$ error is already high, as it is for household movers (0.050 to 0.144 ATE/path across models against 0.012 to 0.088 for person movers); \cref{fig:sens} overlays three criteria. Under an absolute threshold of 0.10 ATE/path ten of eleven models cross between $f=0.2$ and $0.9$ (CUT3R 0.2, Easi3R and Any4D 0.3, MoRe, $\pi^3$, V-DPM and VGGT4D 0.5, 4RC 0.6, MonST3R 0.8, MegaSaM 0.9; Depth Anything~3 has no pooled crossing, interval [0.6, 0.9]) and household movers cross in all eleven; person-versus-abstract differences of 0.10 or more exist for two models. Under 0.15 two models cross, Easi3R at 0.8 and MonST3R at 0.9. H2 is decided on the pre-registered criterion; this analysis is exploratory (Amendment~6.3).

\begin{table*}[t]
\centering
\footnotesize
\caption{Pose degradation thresholds and mask flip points on Axis B (A9.2). $f_\mathrm{deg}$: smallest $f$ at which ATE/path exceeds $3\times$ its $f=0.10$ value (pre-registered, A3.1); $f_\mathrm{mask}$: IoU below $0.5\times$ its $f=0.10$ value or below 0.20 (A3.2); $f_{>0.10}$: first $f$ with ATE/path above 0.10 (A6.3, exploratory). Bootstrap 95\% CIs over sequences; n/a = no flip up to $f=0.95$. Sixteen scene seeds per cell (eight of the pre-registered batch and eight added under Amendment~13); the intervals are those at $n = 16$.}
\label{tab:flips}
\begin{tabular}{lrrrrrr}
\toprule
Model & $f_\mathrm{deg}$ all [CI] & person & abstract & household & $f_\mathrm{mask}$ all [CI] & $f_{>0.10}$ \\
\midrule
MonST3R & n/a & 0.90 & 0.90 & n/a & 0.50 [0.50, 0.60] & 0.80 \\
CUT3R & n/a & n/a & n/a & n/a & n/a & 0.20 \\
Easi3R & n/a & 0.90 & n/a & n/a & n/a & 0.30 \\
VGGT4D & 0.60 [0.40, 0.80] & 0.40 & 0.40 & n/a & 0.30 [0.30, 0.30] & 0.50 \\
$\pi^3$ & n/a & 0.40 & 0.50 & n/a & 0.90 [0.90, 0.95] & 0.50 \\
Any4D & n/a & 0.40 & 0.40 & n/a & n/a & 0.30 \\
MegaSaM & 0.70 [0.40, 0.95] & 0.40 & 0.30 & n/a & 0.60 [0.50, 0.60] & 0.90 \\
Depth Anything 3 & n/a & 0.70 & 0.90 & n/a & 0.80 [0.70, 0.90] & n/a \\
4RC & n/a & 0.60 & 0.80 & n/a & n/a & 0.60 \\
V-DPM & 0.60 [0.50, 0.90] & 0.30 & 0.40 & n/a & n/a & 0.50 \\
MoRe & n/a & 0.40 & 0.40 & n/a & 0.80 [0.80, 0.90] & 0.50 \\
UniQuery4R & n/a & 0.50 & 0.40 & n/a & 0.20 [0.20, 0.20] & 0.50 \\
Point4D & n/a & n/a & n/a & n/a & n/a & 0.20 \\
PAGE-4D & n/a & n/a & n/a & n/a & 0.90 [0.80, 0.95] & 0.40 \\
\bottomrule
\end{tabular}
\end{table*}

\begin{figure*}[t]
  \centering
  \includegraphics[width=\textwidth]{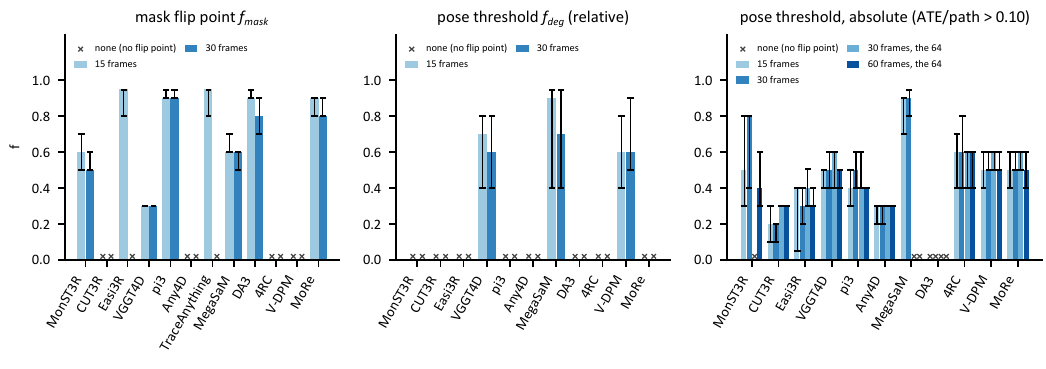}
  \caption{Temporal context (Amendment~13, Gate 6c). Mask flip point (left) and relative pose threshold (middle) per model at 15 and 30 frames over all 704 Axis~B sequences, and the absolute pose threshold (right) at 15 and 30 frames over the 704 and at 30 and 60 frames over the 64 sequences of the 60-frame set, with bootstrap intervals. None marks no crossing up to $f=0.95$ (left, middle, and the 704-sequence bars at right) or within $f\in\{0.3,\dots,0.6\}$ (the 64-sequence bars).}
  \label{fig:context}
\end{figure*}

\paragraph{Temporal Context (Amendment~13).} The thresholds above are read at 30 frames. A13.3 asks whether they move with the length of the clip: every Axis~B sequence of both batches was evaluated again at 15 frames (every second frame of the same render; 8448 runs, matching the 8448 at 30 frames), and 64 sequences were rendered again at 60 frames over the same scene duration (the $f$ levels bracketing each model's pooled threshold, $f\in\{0.3, 0.4, 0.5, 0.6\}$, person and abstract movers, eight seeds; 768 runs). TraceAnything refused the 60-frame clips (its released script down-samples sequences longer than 40 frames, so the 64 runs are recorded as failures rather than evaluated at another length) and one Easi3R run failed the solver assertion. Between 15 and 30 frames the mask flip point moves for four of eleven pose models (MonST3R 0.60 to 0.50, Depth Anything~3 and MoRe 0.90 to 0.80, Easi3R flips at 0.95 at 15 frames and not at 30) and the relative pose threshold for two (VGGT4D 0.70 to 0.60, MegaSaM 0.90 to 0.70); the other seven and nine are unchanged. The 60-frame set spans $f$ from 0.3 to 0.6 and has no $f=0.10$ reference, so the absolute threshold (ATE/path above 0.10) is the reading available at 60 frames: on the 64 sequences it moves for five of eleven models between 30 and 60 frames (Easi3R, VGGT4D, V-DPM and MoRe one level lower at 60 frames; MonST3R crosses at 0.40 at 60 frames and not within the four levels at 30) and for three of eleven between 15 and 30 frames over all sequences (Easi3R 0.40 to 0.30, $\pi^3$ 0.40 to 0.50, MonST3R 0.50 to 0.80 with intervals [0.30, 0.80] and [0.40, 0.80]). The reading is descriptive, as A13.3 pre-registered: neither threshold is fixed across contexts for every model and neither moves for most, and the expected pattern (the flip moving with context, the threshold not) is not what the data show, since both move for a minority (\cref{fig:context}).

\section{Object-Motion Stratification}
\label{app:rho}
Let $\rho$ be the mover's displacement over the sequence divided by the camera's. Of the 352 Axis~B sequences, 88.6\% have $\rho\ge0.5$ (40 below 0.5, 188 between 0.5 and 2, 124 above 2; median 1.39), so the axis can show a pose effect; $\rho$ rises with $f$ because the mover's speed scales with the viewing distance. ATE/path rises from $f\le0.10$ to $f\ge0.50$ in every $\rho$ bin for every pose model except CUT3R and $\pi^3$ in the lowest and highest bins and Easi3R in the lowest, and the smallest rises are in the lowest bin, as expected.

\section{Axis A Cue Effects and Axis D}
\label{app:axisa}
\begin{figure*}[t]
  \centering
  \includegraphics[width=\textwidth]{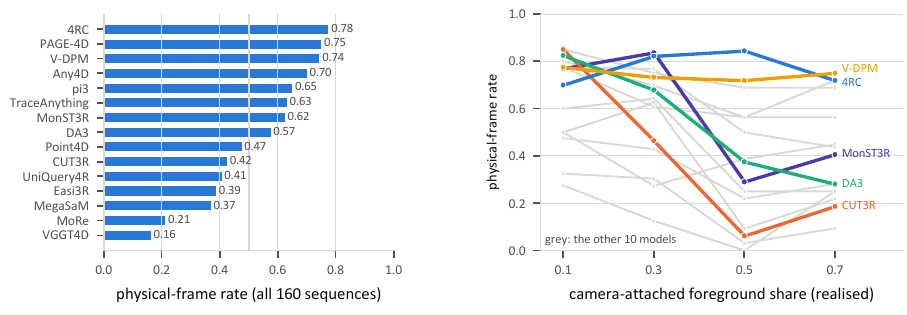}
  \caption{Axis~C frame choice. Left: rate of choosing the physical frame per model over all 160 sequences. Right: the same rate against the realised camera-attached foreground share (bins of 40, 56, 32 and 32 sequences; the sixteen escalator sequences of nominal $g=0.5$ and $0.7$ sit in the 0.3 bin, \cref{tab:levers}); five models highlighted, the other ten in grey.}
  \label{fig:axisc}
\end{figure*}

\begin{figure*}[t]
  \centering
  \includegraphics[width=\textwidth]{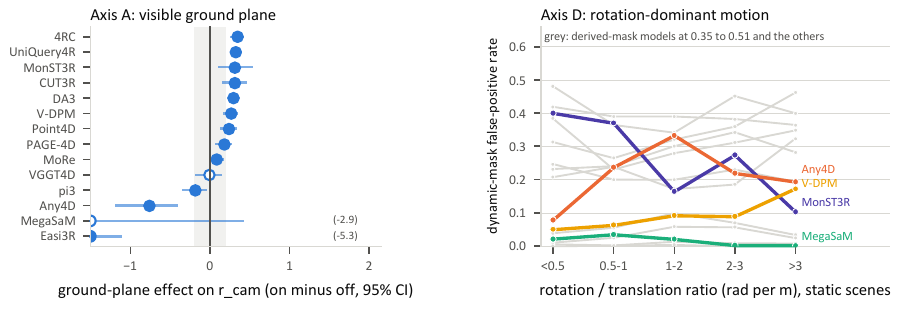}
  \caption{Left, Axis~A: main effect of the visible ground plane on the camera-attribution ratio $r_\mathrm{cam}$ per model (on minus off, 95\% CI, filled when the interval excludes zero; grey band, the pre-registered 0.20 threshold; Easi3R and MegaSaM clipped, values printed). Right, Axis~D: dynamic-mask false-positive rate on static scenes against the rotation-to-translation ratio.}
  \label{fig:axisad}
\end{figure*}

Main effects of the eight cues on $r_\mathrm{cam}$ (mean on minus mean off over the resolution-IV design, 256 sequences per model, 2000 bootstrap resamples) are in the released \texttt{axis\_a\_effects.parquet}; \cref{fig:axisad} plots the ground-plane effect and the Axis~D false-positive rates. Cues shifting $r_\mathrm{cam}$ by 0.20 or more with an interval excluding zero: 4RC ground plane $+0.34$, indoor $+0.21$; Any4D gravity roll $-0.67$, ground plane $-0.76$, indoor $-0.43$, sky and horizon $-0.49$; CUT3R ground plane $+0.31$; Depth Anything~3 ground plane $+0.30$; MonST3R gravity roll $-0.32$, ground plane $+0.31$; MoRe vehicles $+0.20$; V-DPM ground plane $+0.26$; VGGT4D vehicles $+0.25$; $\pi^3$ none. Easi3R (ground plane $-5.28$, sky $+5.76$) and MegaSaM (indoor $+3.63$, texture $+3.37$) are driven by a handful of sequences with $r_\mathrm{cam}$ in the tens or hundreds; with $r_\mathrm{cam}$ capped at 5 their significant cues are gravity roll $-0.54$ (Easi3R) and ground plane $+0.30$, indoor $+0.41$, vehicles $+0.22$ (MegaSaM), and the H1 verdict is unchanged (four of eleven models with two cues). World attribution on these sequences is 0.02 to 0.10 for the dynamic-head models, 4RC and V-DPM, and 0.20 to 0.24 for Any4D, CUT3R and TraceAnything. \Cref{fig:axisc} plots the Axis~C frame choice per model and against the realised foreground share; per-model Axis~C rates by foreground fraction and by platform, and Axis~D depth error, mask false-positive rate and ATE/path per rotation bin with and without the mover, are in the released \texttt{axis\_c.json} and \texttt{axis\_d.json}; the Axis~D bins contain 40, 52, 100, 36 and 12 sequences per mover condition.

\paragraph{The Ground-Plane Cue and Rotation.} On Axis~A the models cannot be right or wrong, just consistent or not, and the cue analysis reads their priors directly (\cref{fig:axisad}). Of the eight cues, the visible ground plane is the one that moves attribution the same way for most models: switching it on raises $r_\mathrm{cam}$ by 0.26 to 0.34 for V-DPM, Depth Anything~3, CUT3R, MonST3R and 4RC (intervals excluding zero) and lowers it for Any4D ($-0.76$) and $\pi^3$ ($-0.19$); the other cues act by less than 0.20 and in model-specific directions. Axis~D separates the two heads a second way: as the rotation-to-translation ratio grows from below 0.5 to above 3 on static scenes, the mask false-positive rate rises for V-DPM (0.051 to 0.173) and Any4D (0.078 to 0.193), falls for MonST3R (0.401 to 0.104) and MegaSaM (0.022 to 0.002), while camera ATE/path is flat across bins for all eleven pose models. Rotation-dominant motion is a problem of attribution, not of geometry, and it lands on the mask head.

\section{Existing Benchmarks}
\label{app:existing}
\begin{figure*}[t]
  \centering
  \includegraphics[width=0.82\textwidth]{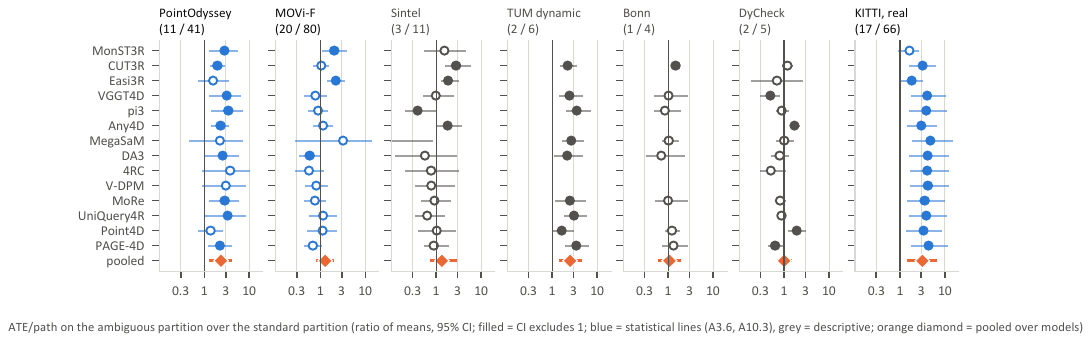}
  \caption{Transfer of the ambiguity score: ratio of mean ATE/path on the ambiguous (top-20\%) partition over the standard partition, per model and pooled, with 95\% bootstrap intervals (filled when the interval excludes 1). MOVi-F and PointOdyssey are the pre-registered statistical lines; the other sets have one to three ambiguous sequences and are descriptive; KITTI is the real-video test.}
  \label{fig:transfer}
\end{figure*}

\Cref{tab:existing} gives ATE/path per model and benchmark at the standard frame count. Five models (4RC, Any4D, Easi3R, MonST3R, V-DPM) exceed the 48\,GB of the GPU on the 90-frame TUM, 110-frame Bonn and 55-frame DyCheck clips at their released settings even when running alone, and TraceAnything's wrapper refuses clips longer than 40 frames because the authors' script would silently down-sample them; these pairs were re-run once at the smallest uniform frame stride in \{2, 3, 4\} that fits and are reported in the separate last column, never mixed into the standard numbers (Amendment~5). MonST3R and TraceAnything assert landscape input on the portrait DyCheck clips (14 pairs not evaluable). The ambiguity-partition statistics (\cref{fig:transfer}) use the standard-frame-count runs of the pose models.

\begin{table*}[t]
\centering
\scriptsize
\caption{Existing benchmarks, camera ATE/path at the standard frame count (mean, $n$ sequences). The last column lists the Amendment~5 reduced-frame re-runs (stride $s$) for clips that exceed 48\,GB or TraceAnything's 40-frame limit; they are never mixed into the standard columns.}
\label{tab:existing}
\begin{tabular}{lrrrrrrp{5.2cm}}
\toprule
Model & Sintel & TUM & Bonn & DyCheck & MOVi-F & PointOdyssey & Reduced frames (A5) \\
\midrule
MonST3R & 0.071 (14) & n/a & n/a & n/a & 0.044 (100) & 0.027 (52) & TUM 0.076 (s2, 8); Bonn 0.080 (s2, 5) \\
CUT3R & 0.100 (14) & 0.052 (8) & 0.100 (5) & 0.021 (7) & 0.085 (100) & 0.061 (52) & n/a \\
Easi3R & 0.170 (13) & n/a & n/a & 0.021 (7) & 0.071 (99) & 0.053 (52) & TUM 0.034 (s2, 8); Bonn 0.069 (s2, 5) \\
VGGT4D & 0.070 (14) & 0.016 (8) & 0.060 (5) & 0.008 (7) & 0.015 (100) & 0.014 (52) & n/a \\
$\pi^3$ & 0.050 (14) & 0.021 (8) & 0.052 (5) & 0.006 (7) & 0.036 (100) & 0.028 (52) & n/a \\
Any4D & 0.124 (14) & n/a & n/a & 0.025 (7) & 0.045 (100) & 0.039 (52) & TUM 0.039 (s2, 8); Bonn 0.100 (s2, 5) \\
MegaSaM & 0.024 (14) & 0.017 (8) & 0.050 (5) & 0.003 (7) & 0.008 (100) & 0.012 (52) & n/a \\
Depth Anything 3 & 0.051 (14) & 0.022 (8) & 0.052 (5) & 0.003 (7) & 0.025 (100) & 0.011 (52) & n/a \\
4RC & 0.058 (14) & n/a & n/a & 0.006 (7) & 0.009 (100) & 0.009 (52) & TUM 0.016 (s2, 8); Bonn 0.044 (s2, 5) \\
V-DPM & 0.061 (14) & n/a & n/a & n/a & 0.021 (100) & 0.011 (52) & TUM 0.021 (s2, 8); Bonn 0.060 (s2, 5); DyCheck 0.007 (s2, 7) \\
MoRe & 0.042 (14) & 0.019 (8) & 0.062 (5) & 0.007 (7) & 0.019 (100) & 0.015 (52) & n/a \\
UniQuery4R & 0.021 (14) & 0.018 (8) & n/a & 0.005 (7) & 0.030 (100) & 0.010 (52) & Bonn 0.054 (s2, 5) \\
Point4D & 0.100 (14) & 0.084 (8) & 0.076 (5) & 0.046 (7) & 0.028 (100) & 0.053 (52) & n/a \\
PAGE-4D & 0.066 (14) & 0.025 (8) & 0.080 (5) & 0.011 (7) & 0.031 (100) & 0.021 (52) & n/a \\
\bottomrule
\end{tabular}
\end{table*}

\paragraph{Score Check (Amendment~20, A20.7).} The ambiguity score could be a proxy for how far or how fast the camera moves. The partial Spearman correlation of the score with ATE/path, controlling for the ground-truth camera path length and the rotation-to-translation ratio (pooled over the eleven pose models, 2000 bootstrap resamples over sequences), is PointOdyssey (52 sequences) $+0.13$ [$-0.07$, $+0.31$] against the plain $+0.29$ [$+0.10$, $+0.45$], 10 of 11 models positive; the ambiguous over standard ratio with rotation RPE as the outcome is 1.08 [0.89, 1.39] against 2.34 [1.52, 3.42] on ATE/path; MOVi-F (100 sequences) $+0.14$ [$+0.07$, $+0.22$] against the plain $+0.13$ [$+0.01$, $+0.24$], 11 of 11 models positive; the ambiguous over standard ratio with rotation RPE as the outcome is 4.47 [2.09, 8.18] against 1.28 [0.93, 1.73] on ATE/path; KITTI (83 sequences) $+0.15$ [$+0.02$, $+0.26$] against the plain $+0.37$ [$+0.24$, $+0.49$], 10 of 11 models positive; the ambiguous over standard ratio with rotation RPE as the outcome is 0.88 [0.50, 1.22] against 3.11 [1.74, 5.73] on ATE/path; the egocentric set (216 sequences) $+0.25$ [$+0.18$, $+0.33$] against the plain $+0.37$ [$+0.30$, $+0.42$], 11 of 11 models positive; the ambiguous over standard ratio with rotation RPE as the outcome is 1.20 [0.82, 1.56] against 1.59 [1.36, 1.85] on ATE/path. The score keeps a positive partial correlation on every set, with an interval above zero on three of the four (PointOdyssey's includes zero), and the ambiguous partition is harder in rotation on MOVi-F alone: what the score measures is not the camera's path length or its rotation (\texttt{results/phase9/score\_check/readings.json}).

\section{Ambiguity Detector}
\label{app:detector}
\begin{figure}[t]
  \centering
  \includegraphics[width=\linewidth]{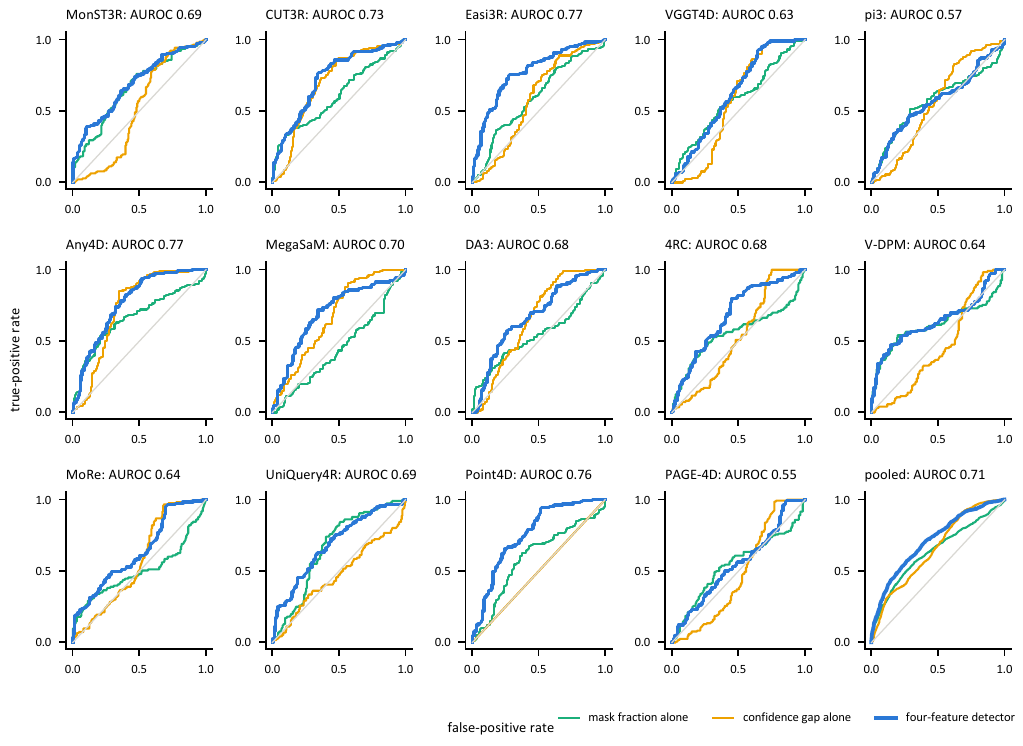}
  \caption{ROC curves of the ambiguity detector (blue) and the two one-feature baselines per model and pooled, on the test split.}
  \label{fig:roc}
\end{figure}

The detector of Amendment~7 takes features per sequence from the model's own outputs: the fraction of pixels its dynamic mask flags; its camera travel divided by the median predicted depth of the first frame; their product; and the mean confidence on the flagged region minus the mean confidence on the rest (derived masks recomputed with the frozen method for the six derived-mask models). A logistic regression with $C=1$ on standardised features, fitted per model on the validation split of Axes~B and C and the validation partitions of MOVi-F and PointOdyssey (the ambiguous validation lists plus the standard sequences with even hash parity), predicts ATE/path above 0.10; the baselines are one-feature fits of the mask fraction and of the confidence gap. The test split was opened once for this evaluation and the script refuses to run it again. \Cref{fig:roc} shows the ROC curves; per-model AUROC runs from \detAUROCmin{} to \detAUROCmax{}, the pooled AUROC is \detPooledAUROC{} \detPooledAUROCci{} over \detNtest{} test sequences, and the detector beats both baselines with an interval excluding zero for \detNbeat{} of \detNmodels{} models, short of the pre-specified bar. The three models added under Amendment~16 had their one test evaluation each after the pre-registered reading was written: their AUROC runs from \detAUROCminAdded{} to \detAUROCmaxAdded{} and \detNbeatAdded{} of \detNmodelsAdded{} (Point4D) beats both baselines; they enter the figure and not the pooled reading.

\section{Pre-Registration and Amendments}
\label{app:prereg}
\paragraph{Protocol.} The protocol fixed the claim, the four axes, the metrics, three hypotheses (H1 cues, H2 flip point, H3 method) and a sanity check (H4), three kill criteria (K0 fewer than three runnable models, K1 no finding, K2 no fix meets H3 on validation), the renderer, the wrapper, the alignment, the three fix candidates and the splits before any sequence was rendered. The study ran in numbered phases, each closed by a gate, a decision record with the decision on its own line (Phase~0 is the model inventory, Phase~2 the diagnostic pass over the released models, Phase~3 the inference-time candidates and Phase~5 the real-video and training passes); the verdicts below cite the gates by number.

\paragraph{Amendments (Dated, Signed by the PI).} A1: the H4 reference is the model's own Sintel error measured by the same code path, because published numbers are in incomparable units or absent. A2: ATE normalised by the camera path length, because the path shrinks with $f$. A3: definitions of the thresholds, the $\rho$ stratification, a scale-free $r_\mathrm{cam}$, one derived-mask round on validation, and descriptive treatment of partitions with fewer than ten ambiguous sequences. A4: the depth scale of $r_\mathrm{cam}$ taken on the placed objects' pixels. A5: reduced-frame re-runs for clips that exceed the GPU or the model's own frame limit, reported separately. A6: the real-world set reduced to six situations, the oracle ceiling and the flip-criterion sensitivity added as analyses, and K2 read on the validation split with the test split kept unopened. A7: a within-model ambiguity detector with a one-shot test evaluation and the wording of the headline finding fixed before the test. A8: the ground-truth-mask ceiling and the decision rule for building a method, with PointOdyssey excluded from that pass because it has no per-pixel dynamic mask. A9: the two-regime reading of that ceiling as an added analysis, the renaming of the pose quantity, the geometric correlation analysis, and the real-world panel reduced to four situations with the fallback used here. A10: the real-video replication with ground-truth poses and masks and its three pre-specified readings. A11: the training-side intervention (one model, two fine-tunes, one test evaluation) and its pre-specified reading. A12: one replay-mixed fine-tune with its reading restated for the observed sample sizes, and the forgetting curve. A13: robustness additions with no verdict change: a second architecture for the training-side test, Axis~B extended to sixteen scene seeds per cell, a temporal-context evaluation at 15 and 60 frames, a model sweep for releases after the zoo was built, a rigid-world real-video analogue on nuScenes-mini, and a benchmark page. A14: the ground-truth-mask pass extended below the degradation threshold, the matched-fraction re-reading of the egocentric mover result, and the wording of the real frame-choice result fixed before the pass. A15: the real-video transfer of the fine-tunes with its three readings fixed in advance. A16: three models missed by the zoo check, added as released with a scope rule by GPU time; their results change no verdict. A17: the attributional repair as a data recipe (data scaling, a scope test on a third and fourth architecture, a real-label instantiation, a downstream metric from cached outputs), each reading fixed in advance. A18: two further data-scaling points at one and five sequences, with no reading attached. A19: two bounds on the recipe, the sequence dependence of the one-sequence point and a second architecture chosen by a rule fixed before the check, with the readings and the paper's wording for each outcome fixed in advance. A20: in response to an external review, attribution read across outputs, a calibration control for the recipe, a replication of the five-sequence point over three further draws, a classical estimator on the remaining static pixels, a matched occluder control (Axis~B$'$), cluster bootstrap intervals and a partial-correlation check of the ambiguity score, each reading fixed before the analysis ran (\cref{app:a20}). A21: the hands camera-depth attribution decomposed across every recipe checkpoint, descriptive, with the paper's wording for each outcome fixed in advance. A22: in response to a second external review, measurement audits (a directional analysis of the matched control, a validation of the camera-depth ratio on constructed predictions, the full calibration envelope, per-draw margins, one uncertainty policy, ties, the oracle residual, a scale-free classical chain) and wording corrections, each reading and its wording fixed before the analysis ran (\cref{app:a22}). A23: the background-normalised camera label, validated under A22.3, replaces the foreground-normalised ratio as the main text's camera label; descriptive, no verdict changes. A24: in response to a third external review, the mover-IoU difference against the IoU-optimal threshold with its reading fixed before the analysis ran, and three wording corrections (the binary-mask sentence withdrawn, the downstream-cell sentence scoped to its eligibility rule and marked exploratory, the camera label's scope stated at its definition); no verdict changes. A25: realised lever values, the mean ground-truth share of every Axis~B, B$'$ and C sequence against the nominal value in its name, with the sensitivity of the by-$g$ readings to the escalator's saturated cells (\cref{app:metrics}); a measurement audit, no hypothesis, threshold, sequence set or verdict rule changes; the Axis~C panel is plotted by realised share (A25.3). The original H1 to H3 thresholds were retained; later intervention criteria and measurements were amended as documented.

\paragraph{Verdicts.} K0 not fired (11 of 12 models runnable). H4: six models pass, five fail the mask clause and carry the caveat. H1: fail (five of eleven models with two significant cues). H2: fail (five of eleven pooled crossings; class difference for five of eleven). K1 not fired (Axes A to D all show model dependence). H3: not tested; K2 fired on validation. Gate 3b (detector): negative result. Gate 3b-2 (ground-truth mask): no method. Gate 5 (driving set): reading (i) Axis~B part replicated, Axis~C part not testable, reading (ii) replicated, reading (iii) not testable. Gate 5 addendum (egocentric set): reading (i) Axis~B part not replicated, Axis~C part replicated, reading (ii) replicated, reading (iii) replicated. Gate 5b (training side): not supported as pre-specified (the frame-choice gain and the unchanged Axis~B threshold hold; the Sintel regression bound does not). Gate 5c (replay fine-tune): not a targeted repair as pre-specified (Bonn and TUM within the bound, Sintel not). Gate 6a (sixteen seeds per Axis~B cell): no pre-registered verdict changes; the flip tables and figures use the sixteen-seed values. Gate 6b (second architecture): the dissociation does not replicate as pre-specified on MoRe (the Axis~C gain, $+81.2$ points, and the unchanged Axis~B threshold hold; the dominant-mover control raises the rate as well, $+66.2$ points). Gate 6d (ground-truth mask below the threshold): the Axis~B statement stays as written ($+5.0\%$ [$+2.0$, $+8.1$] below the threshold); the egocentric mover reading at matched fractions: not replicated. Gate 6c (temporal context, descriptive): no pre-registered verdict changes; between 15 and 30 frames the mask flip point moves for four of eleven pose models and the relative pose threshold for two; between 30 and 60 frames the absolute threshold moves for five of eleven on the 64-sequence set. Gate 6e (real-video transfer): MoRe's moving-platform tune transfers ($+60.5$ points on the egocentric hands clips, mover and leading-vehicle error with no detected increase) and the gain is attributed to physical-frame motion supervision in general; MonST3R's does not ($-6.2$ points, mover error $+87\%$). Gate 7 (the recipe): A17.2 data requirement 25 rendered sequences; A17.3 motion-head half no second instance available, no-motion-head half supported (CUT3R $-60.5$ points on the hands clips, mover error $+82\%$, Sintel and TUM worse); A17.4 real-label variant does not meet its bar (held-out hands $+86.7$ points, leading-vehicle error $+11\%$, Sintel $+5\%$); A17.5 descriptive; no pre-registered verdict changes. Gate 8 (the recipe's two bounds): A19.2 one of three single moving-platform sequences meets the one-sequence bar, and the one-sequence control attributes the gain at one sequence to moving-platform data; A19.4 not a second positive instance (the head added to VGGT4D: hands $+54.3$ points, mover error $+4.8\%$, TUM $+3.9\%$); no pre-registered verdict changes. Gate 9 (Amendment~20): A20.1 no mask reversal on Axis~B (four partial masks, one collapse); A20.2 the hands rate gain is attributable to calibration under the pre-registered rate reading, the margin and the mover IoU are not; A20.3 three of four five-sequence draws meet the bar; A20.4 the classical estimator recovers the camera up to $f=0.5$ on Axis~B and at every coverage on Axis~B$'$; A20.5 the moving member of every matched pair costs more camera error than the frozen one; A20.6 every pooled interval keeps its sign under the cluster unit; A20.7 the score keeps a positive partial correlation on every set; the geometric reading of \cref{app:phase3} (Amendment~9.3) is replaced in the main text by the attribution reading, and no pre-registered verdict changes.

\paragraph{Attempts.} Every failure was logged with its traceback and its re-runs (the full tables are in the released \texttt{RESULTS.md}). \Cref{tab:attempts} compresses them to one row per class.

\begin{table}[t]
\centering
\footnotesize
\caption{Failures over the study by class, with how each class was closed. Environment failures were re-run after the fix; model failures are results.}
\label{tab:attempts}
\begin{tabular}{p{3.0cm}rp{3.4cm}}
\toprule
Class & Runs & Closed by \\
\midrule
Install: CUDA dispatch API break in four models' kernels & 4 & one-line \texttt{scalar\_type()} patches, recorded per model \\
GPU out of memory, third process on a GPU (Phases 0 to 2) & 24 & re-run; rule: two per GPU with a 24\,GB launch guard \\
Root disk full from per-run outputs (Phase 2) & 4 & outputs moved to the store volume; re-run \\
Existing-benchmark clips over 48\,GB alone on the GPU & 72 & reduced-frame re-run, separate column (A5) \\
TraceAnything over-40-frame refusals & 76 & reduced-frame re-run, separate column (A5) \\
Portrait DyCheck input asserted by two models & 14 & not evaluable \\
Machine reboot mid-Phase 3 & 4 & re-run \\
GitHub outage in a model's \texttt{torch.hub} call (Phase 3) & 14 & cached-repository patch in the environment; re-run \\
Memory collision with a 60-frame clip run (Phase 3) & 5 & serialised; re-run \\
Model crash: deterministic solver assertions (MonST3R, Easi3R), clustering abort (VGGT4D) & 34 & kept as results \\
Fine-tune out of memory beside an evaluation lane (Phase 5) & 2 & lane paused; re-run alone \\
Real-video pass out of memory beside a fine-tune (Phase 5) & 235 & re-run after the fine-tune \\
Store disk full during the egocentric pass (Phase 5) & 1307 & Phase 2 work directories removed with approval; re-run \\
Egocentric pass: backend exit beside a 32 to 35\,GB process (Phase 5) & 142 & lanes re-paired, V-DPM alone on its GPU; re-run \\
Egocentric pass: two lanes re-running the same failed clip at once (Phase 5) & 2 & harness skip rule fixed; re-run \\
\bottomrule
\end{tabular}
\end{table}

\section{Inference-Time Ceilings}
\label{app:phase3}
\begin{figure}[t]
  \centering
  \includegraphics[width=\linewidth]{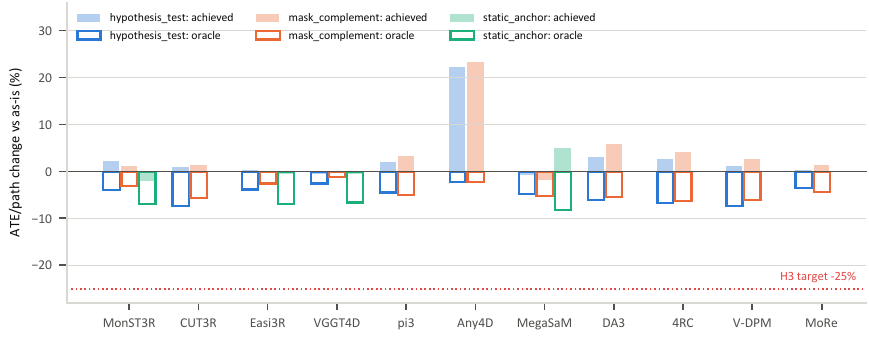}
  \caption{Achieved (filled) and oracle (outlined) ATE/path change per model and candidate on the ambiguous validation subsets; the dotted line is the pre-registered H3 target.}
  \label{fig:oracle}
\end{figure}

Candidates were implemented for every model with a mask hook (every backend accepts a static-world mask that is resized to its working resolution; static-anchor conditioning applies to the four models with a native mask input, MonST3R, Easi3R, VGGT4D and MegaSaM). The selection score is the reprojection consistency of the decomposition's static part plus the gravity alignment of the recovered camera, as pre-registered. Validation subsets: Axis~B sequences at $f\ge f_\mathrm{deg}$ (or $f\ge0.5$ where no pooled crossing exists), Axis~C, and the ambiguous validation partitions of MOVi-F and PointOdyssey; the regression check is the Sintel validation partition (7 sequences) with a bound of $+2\%$ ATE. Six masked passes crashed deterministically in Easi3R's pose solver and four pairs had no as-is output; everything else ran. \Cref{tab:selection} lists the achieved and oracle changes; \cref{fig:oracle} plots them. The oracle re-scores the cached masked output of every run against ground truth and keeps the better decomposition per sequence; it agrees with the harness wherever both exist (2044 runs, maximum difference 0).

\begin{table*}[t]
\centering
\scriptsize
\caption{Phase 3 selection record on the validation split (ambiguous subsets: Axis B at $f \ge f_\mathrm{deg}$, Axis C, ambiguous partitions of MOVi-F and PointOdyssey) with the oracle-selection ceiling (A6.2). Achieved: the candidate's own consistency-plus-gravity selection; oracle: the better of as-is and masked per sequence. Paired bootstrap 95\% CIs. H3 required $-25\%$; K2 fired.}
\label{tab:selection}
\begin{tabular}{llrrrrrl}
\toprule
Model & Candidate & $n$ & Achieved $\Delta$ATE/path & Oracle $\Delta$ATE/path & Masked better & Sintel val $\Delta$ATE & Within bound \\
\midrule
MonST3R & hypothesis test & 123 & +2.4 [+0.3, +4.8] & -3.9 [-5.9, -2.2] & 0.43 & -0.1 & yes \\
MonST3R & mask complement & 123 & +1.1 [-0.5, +2.7] & -3.0 [-5.0, -1.4] & 0.37 & -4.9 & yes \\
MonST3R & static anchor & 123 & -2.1 [-7.1, +2.9] & -6.9 [-11.1, -3.7] & 0.46 & +12.5 & no \\
CUT3R & hypothesis test & 190 & +1.0 [-0.1, +2.4] & -7.3 [-9.7, -5.1] & 0.29 & -0.1 & yes \\
CUT3R & mask complement & 190 & +1.4 [+0.3, +2.9] & -5.7 [-7.6, -3.7] & 0.31 & +0.0 & yes \\
Easi3R & hypothesis test & 185 & +0.3 [-1.2, +1.9] & -3.7 [-5.3, -2.4] & 0.45 & +8.4 & no \\
Easi3R & mask complement & 187 & -0.0 [-1.8, +1.5] & -2.5 [-4.1, -1.5] & 0.45 & -4.0 & yes \\
Easi3R & static anchor & 183 & -0.5 [-3.9, +2.9] & -6.9 [-9.5, -4.6] & 0.43 & +20.6 & no \\
VGGT4D & hypothesis test & 190 & -0.4 [-1.3, +0.4] & -2.5 [-3.2, -1.8] & 0.55 & +0.2 & yes \\
VGGT4D & mask complement & 190 & +0.1 [-0.3, +0.6] & -1.1 [-1.5, -0.7] & 0.47 & -0.1 & yes \\
VGGT4D & static anchor & 190 & -0.5 [-3.5, +2.6] & -6.6 [-8.5, -4.6] & 0.46 & +0.6 & yes \\
$\pi^3$ & hypothesis test & 190 & +2.0 [+0.0, +4.5] & -4.4 [-6.1, -3.0] & 0.32 & +4.0 & no \\
$\pi^3$ & mask complement & 190 & +3.3 [+1.2, +5.9] & -5.1 [-7.0, -3.3] & 0.29 & +9.3 & no \\
Any4D & hypothesis test & 190 & +22.1 [+16.2, +28.9] & -2.2 [-3.4, -1.3] & 0.16 & +54.5 & no \\
Any4D & mask complement & 190 & +23.4 [+17.5, +30.7] & -2.2 [-3.3, -1.3] & 0.14 & +54.9 & no \\
MegaSaM & hypothesis test & 206 & -0.8 [-2.2, +0.5] & -4.8 [-7.0, -2.8] & 0.43 & -9.5 & yes \\
MegaSaM & mask complement & 206 & -1.7 [-4.0, +0.2] & -5.3 [-7.6, -3.4] & 0.42 & +0.0 & yes \\
MegaSaM & static anchor & 206 & +5.1 [-0.1, +11.0] & -8.2 [-11.1, -5.6] & 0.42 & +119.7 & no \\
Depth Anything 3 & hypothesis test & 190 & +3.0 [+0.6, +6.5] & -6.0 [-8.4, -3.8] & 0.34 & +0.0 & yes \\
Depth Anything 3 & mask complement & 190 & +5.8 [+2.1, +10.4] & -5.5 [-8.0, -3.3] & 0.31 & -0.1 & yes \\
4RC & hypothesis test & 190 & +2.7 [-0.5, +6.7] & -6.7 [-8.9, -4.7] & 0.33 & -0.1 & yes \\
4RC & mask complement & 190 & +4.0 [+0.8, +8.2] & -6.2 [-8.3, -4.2] & 0.33 & +3.9 & no \\
V-DPM & hypothesis test & 174 & +1.1 [-1.1, +3.7] & -7.3 [-10.7, -4.6] & 0.41 & -0.2 & yes \\
V-DPM & mask complement & 174 & +2.7 [+0.5, +5.1] & -6.1 [-9.2, -3.6] & 0.36 & -0.2 & yes \\
MoRe & hypothesis test & 174 & +0.4 [-1.1, +1.9] & -3.5 [-5.5, -1.9] & 0.29 & -0.7 & yes \\
MoRe & mask complement & 174 & +1.4 [-0.7, +4.1] & -4.3 [-6.6, -2.4] & 0.29 & -0.2 & yes \\
\emph{pooled} & hypothesis test & 2002 & +3.3 [+2.5, +4.3] & -4.7 [-5.3, -4.1] & 0.36 & n/a & n/a \\
\emph{pooled} & mask complement & 2004 & +4.0 [+3.1, +5.0] & -4.2 [-4.8, -3.6] & 0.34 & n/a & n/a \\
\emph{pooled} & static anchor & 702 & +0.6 [-1.4, +2.7] & -7.2 [-8.6, -5.8] & 0.44 & n/a & n/a \\
\bottomrule
\end{tabular}
\end{table*}

\begin{figure}[t]
  \centering
  \includegraphics[width=\linewidth]{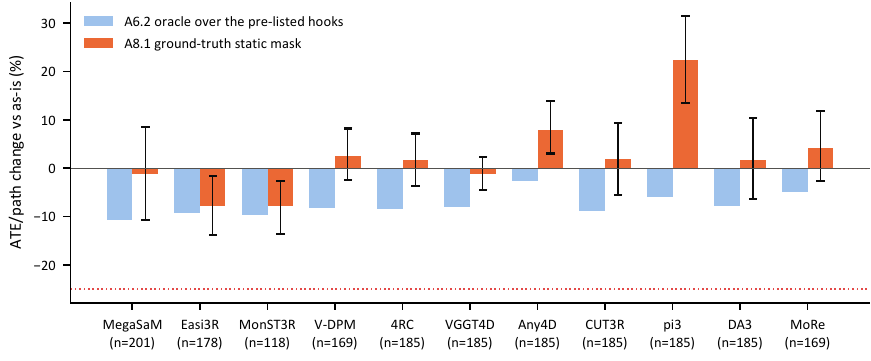}
  \caption{Two ceilings per model on the ambiguous validation subsets: the best-of-candidates oracle over the pre-listed hooks (blue) and the ground-truth static mask (orange, 95\% CI); the dotted line is the pre-registered H3 target.}
  \label{fig:gtmask}
\end{figure}

\paragraph{The Ground-Truth Decomposition (Amendment~8).} The oracle above bounds selection between passes; it does not bound what a correct decomposition would buy. We therefore gave every pose model the ground-truth static region on the same validation subsets (Axis~B at $f\ge f_\mathrm{deg}$, Axis~C and the MOVi-F ambiguous validation partition; PointOdyssey has no per-pixel dynamic mask and is excluded): MonST3R, Easi3R, VGGT4D and MegaSaM received it as their mask input, the other seven received frames with the ground-truth mover filled by the hypothesis-test masking operator. Over 1945 paired sequences the pooled ATE/path change is $+1.9\%$ [$-0.3$, $+4.1$]; two models improve with a CI excluding zero (MonST3R $-7.9\%$, Easi3R $-7.7\%$) and two get worse (Any4D $+8.0\%$, $\pi^3$ $+22.5\%$). The split by axis is the finding of Sections~4.2 and 4.3 of the main text: $-10.3\%$ [$-12.7$, $-7.7$] on Axis~C against $+17.1\%$ [$+13.9$, $+20.3$] on Axis~B (cluster intervals in \cref{tab:cluster}). Ten of 1955 scheduled pairs failed (six carried a Phase~2 failure forward, four were Easi3R's pose solver asserting on the masked input). Under the pre-registered decision rule (a pooled reduction of at least 25\% with six models' CIs excluding zero to build the WorldMask candidate; below 15\% no method) no method was built. \Cref{fig:gtmask} plots both ceilings per model.

\begin{figure}[t]
  \centering
  \includegraphics[width=\linewidth]{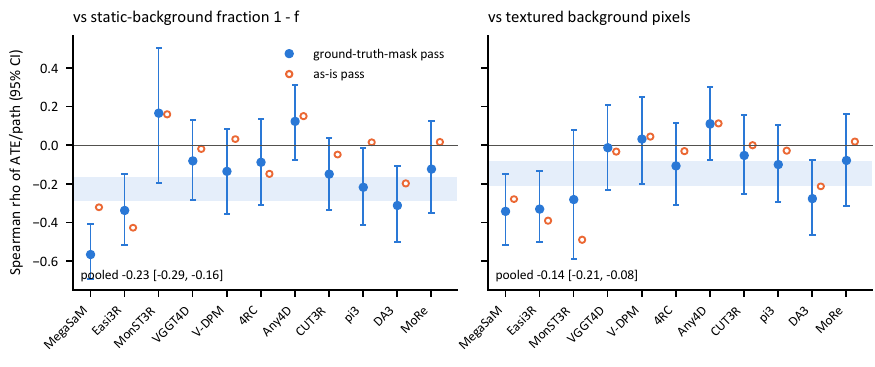}
  \caption{Geometric reading. Spearman correlation of the Axis~B ATE/path with the static-background fraction $1-f$ (left) and with the textured background pixel count (right), per model with 95\% intervals, for the ground-truth-mask pass (filled) and the as-is pass (open); the band is the pooled interval.}
  \label{fig:geo}
\end{figure}

\paragraph{The Geometric Reading (Amendment~9.3).} For every Axis~B pair of the ground-truth-mask pass (973 pairs, 89 sequences, 11 models) we correlated the masked-pass ATE/path with the static-background fraction $1-f$ and with the number of textured background pixels (Sobel gradient above 0.1 on the grey frame within the ground-truth static region, frames 0, 15 and 29 averaged), Spearman with 2000 bootstrap resamples over sequences (\cref{fig:geo}). Pooled: $-0.23$ [$-0.29$, $-0.16$] with the static fraction and $-0.14$ [$-0.21$, $-0.08$] with texture; four and three of the eleven models have their own interval below zero; the as-is pass gives $-0.15$ and $-0.12$. The two covariates correlate at 0.62. Less static context means more error whichever mask the model is given.

\section{Real Video}
\label{app:realvideo}
\begin{figure*}[t]
  \centering
  \includegraphics[width=\textwidth]{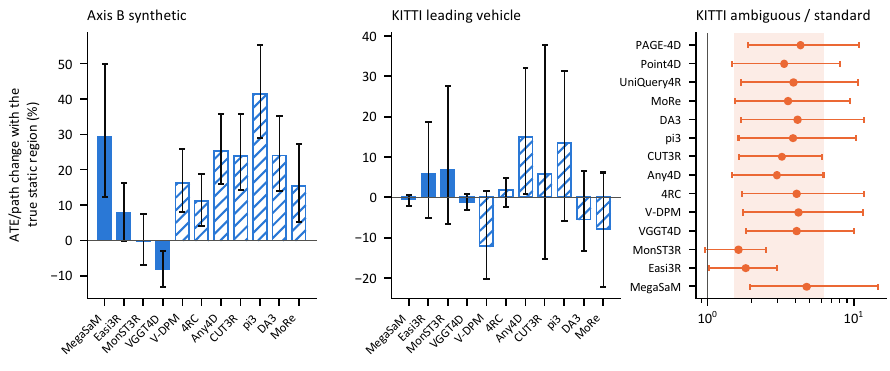}
  \caption{Real-video replication. Left: the synthetic Axis~B change with the true static region (as in \cref{fig:regimes_supp}). Middle: the same on the KITTI leading-vehicle clips. Right: per-model ratio of ATE/path on the ambiguous over the standard KITTI partition with 95\% intervals (the band is the pooled interval).}
  \label{fig:rv}
\end{figure*}

The real-video set of Amendment~10 takes the 21 training sequences of the KITTI tracking benchmark~\cite{geiger2012kitti} (left colour camera at 10\,Hz, OXTS ego poses, 2D and 3D object tracks; CC BY-NC-SA 3.0). Every 30-frame window at a stride of 5 frames (1487 windows) is classified: lead, a tracked vehicle ahead that moves faster than 0.5\,m/s in the world frame and whose box covers a fraction of the frame binned as 0.05 to 0.10, 0.10 to 0.20 and 0.20 to 0.30 (no window reaches 0.30 or more, a fact of roof-mounted driving footage); convoy, a vehicle moving with the ego vehicle within 1\,m/s (an Axis~A analogue, descriptive); generic, random windows. Windows with annotation gaps are dropped and counted (53). The selected 83 clips (41 lead, 2 convoy, 40 generic) are exported at 512 pixels long side with the camera-2 pose from the OXTS track and the calibration, and a dynamic mask that is the union of the boxes of moving tracks dilated by 2 pixels; the ambiguity score is computed exactly as for the existing benchmarks and the top 20\% (17 clips) is the ambiguous partition. No public driving set shows the ego vehicle's hood, so the Axis-C readings (i-C and iii) are not testable on the driving set; the egocentric set below supplies them. All eleven pose models ran as released and with the ground-truth static region through the same hooks as the Amendment~8 pass. Readings: (i) on the lead clips the complement of the dilated moving-object boxes, the set's static region, changes ATE/path by $+2.1\%$ [$-2.2$, $+6.6$] pooled (451 pairs), native models $+3.2\%$ [$-2.1$, $+10.4$], input-fill models $+1.4\%$ [$-4.7$, $+8.2$]: replicated (the interval includes zero, as pre-specified for the dominant-mover regime); (ii) the ambiguous partition's ATE/path over the standard partition is 3.11 [1.52, 6.29] (17 against 66 clips, cluster bootstrap), eleven of eleven models above 1 and ten with their own interval above 1: replicated; a sensitivity line that was not pre-specified keeps the 67 clips whose ego vehicle moves faster than 1\,m/s (ten of the seventeen ambiguous clips are near-stationary, where ATE divides by a path of about 1\,m) and gives 1.33 [0.92, 1.90] with eight models above 1 on their own; (iii) not testable on the driving set. \Cref{fig:rv} plots the per-model numbers next to the synthetic Axis~B panel. Runs: 1662 ok after the re-runs of 235 out-of-memory failures that occurred while a fine-tune shared the GPU; no model failure.

\begin{figure*}[t]
  \centering
  \includegraphics[width=\textwidth]{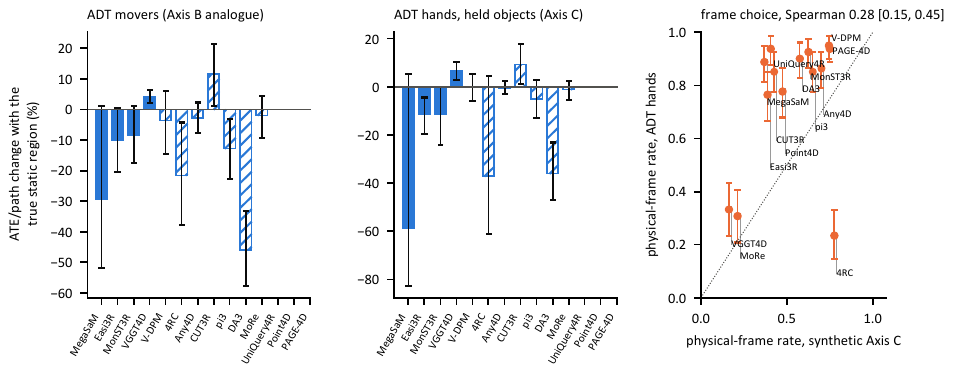}
  \caption{Egocentric set (Aria Digital Twin). Left: ATE/path change with the true static region on the mover clips (Axis~B analogue), per model with 95\% intervals (filled: native mask input; hatched: input fill). Middle: the same on the hands clips (Axis~C analogue). Right: per-model physical-frame rate on the hands clips against the synthetic Axis~C rate, with the identity line.}
  \label{fig:adt}
\end{figure*}

\paragraph{Egocentric Set (Aria Digital Twin).} The Aria Digital Twin set~\cite{pan2023adt} (236 recordings of two indoor scenes with a digital-twin ground truth; licence accepted by the principal investigator, download links not redistributed) supplies both analogues. Every 30-frame window at 10\,Hz (stride 15 recorded frames, 43290 windows over all recordings, 461 dropped for annotation gaps) is scored from the twin: mover, a person or object whose scene-frame displacement in the window exceeds 5\,cm, binned by its fraction of the frame; hands, the wearer's skeleton segmentation and any held object (an object that moves at least 0.3\,m in the world while moving less than 0.3 of that in the device frame) as a camera-attached foreground binned as $g$; generic, random windows. Box fractions on the fisheye over-estimate the measured mask fractions on the exported pinhole by 2.4 times, so a second selection round on measured fractions added 6 clips; measured mover fractions reach the 0.30 to 0.50 bin for 5 clips and none above, measured $g$ reaches the 0.20 to 0.40 bin for 11 clips and none above (the hands bins in \cref{fig:adt} are by selection fraction), a property of the set recorded under the `whatever the data support' clause of Amendment~10.2. The 216 clips (105 mover, 81 hands, 30 generic, from 17 recordings) are re-projected from the fisheye RGB to a 512 pixel pinhole with a 90 degree field of view, rotated upright, with the device trajectory as pose and the instance segmentation dilated by 2 pixels as the dynamic mask (on the hands clips the mask is the camera-attached foreground); the ambiguous partition is the top 20\% (44 clips) of the same score. All eleven models ran as released and with the ground-truth static region (4422 runs, none failed after the re-runs of \cref{tab:attempts}). Readings (\cref{fig:adt}): (i-B) on the mover clips the true static region changes ATE/path by $-6.0\%$ [$-9.3$, $-2.9$] pooled (1155 pairs), native models $-6.2\%$ [$-11.1$, $-1.6$], input fill $-5.9\%$ [$-9.9$, $-2.0$]: not replicated under the pre-specified rule (the interval excludes zero on the helping side); the movers sit below every degradation threshold, where the synthetic ground-truth-mask pass (run at $f\ge f_\mathrm{deg}$) has no counterpart, and the change is concentrated in one bin (0.20 to 0.30 by selection fraction, $-19.2\%$ [$-25.5$, $-12.8$]; the other four bins include zero or are positive). (i-C) On the hands clips the true static region changes ATE/path by $-6.7\%$ [$-10.7$, $-2.5$] pooled (891 pairs), native models $-10.8\%$ [$-19.4$, $-2.2$], input fill $-4.3\%$ [$-8.5$, $-0.4$], growing with $g$ from $-1.6\%$ [$-5.4$, $+1.8$] below 0.2 to $-14.6\%$ [$-24.7$, $-4.4$] at 0.6 and above: replicated (the synthetic Axis~C reads $-10.3\%$ [$-12.7$, $-7.7$]). (ii) The ambiguous partition's ATE/path over the standard partition is 1.59 [1.30, 1.97] (44 against 172 clips), eleven of eleven models above 1 and nine on their own: replicated; both real sets pooled give 1.95 [1.52, 2.52]. (iii) On the hands clips a model chooses the physical frame when its dynamic mask covers the camera-attached foreground more than the rest; the rates span 0.72 [0.64, 0.81] (0.23 for 4RC, 0.31 for MoRe, 0.33 for VGGT4D, 0.77 to 0.95 for the other eight; pooled 0.72) and correlate with the synthetic Axis~C rates at Spearman 0.28 [0.15, 0.45] over clip resamples: replicated as pre-specified (the interval excludes zero and the rates differ), read as model-specific on real video with weak rank agreement (Amendment~14.4); two orderings reverse, 4RC at 0.78 synthetic against 0.23 real and MegaSaM at 0.37 against 0.89, so the synthetic ordering is not claimed to predict the real one. None of the four readings changes a pre-registered verdict. The hands bins above are selection-box bins; by the measured foreground fraction of the twin's mask (\texttt{results/phase9/hands\_measured\_bins.json}), the true static region changes ATE/path by $+2.0\%$ [$-1.5$, $+5.7$] on the 46 clips below 0.05 of the frame, by $-5.8\%$ [$-9.4$, $-2.5$] on the 24 clips at 0.05 to 0.20 and by $-26.0\%$ [$-39.2$, $-12.9$] on the 11 clips at 0.20 to 0.40 (none above 0.40); the top selection bin (box fraction at least 0.6) holds 20 clips whose measured foreground spans 0.01 to 0.36 of the frame.

\begin{figure*}[t]
  \centering
  \includegraphics[width=\textwidth]{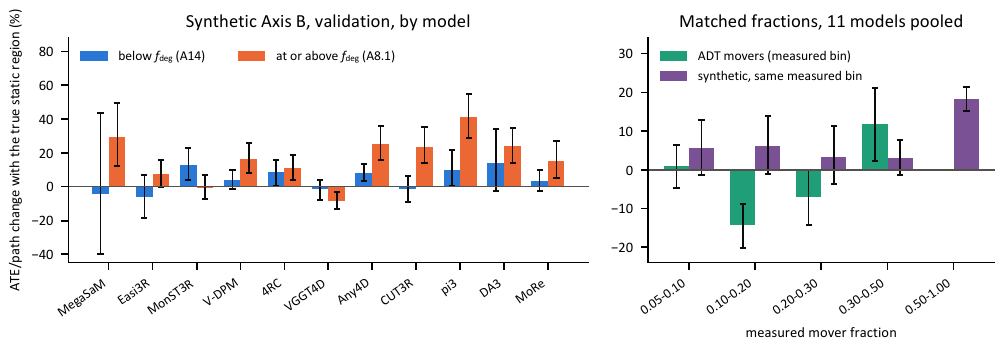}
  \caption{Ground-truth mask below the degradation threshold (Amendment~14). Left: ATE/path change with the true static region on the Axis~B validation sequences below each model's threshold (blue) and at or above it (orange, the Amendment~8 pass), per model with 95\% intervals. Right: the same change in matched bins of measured mover fraction, egocentric movers against synthetic sequences, eleven models pooled.}
  \label{fig:gtbelow}
\end{figure*}

\paragraph{Below the Degradation Threshold (Amendment~14).} The egocentric movers sit below every model's degradation threshold, where the Amendment~8 pass had no synthetic counterpart, so that pass was extended to the Axis~B validation sequences with $f<f_\mathrm{deg}$ per model ($f<0.5$ for the models without a threshold; 960 pairs, 959 ok, one Easi3R solver assertion), with the same hooks and masking operator. Below the threshold the true static region changes ATE/path by $+5.0\%$ [$+2.0$, $+8.1$] pooled (native-mask models $+3.8\%$ [$-2.6$, $+10.1$], input fill $+5.6\%$ [$+2.3$, $+9.3$]) against $+17.1\%$ [$+13.9$, $+20.3$] at or above it (native $+7.5\%$ [$+2.4$, $+13.0$]); no model's interval lies below zero (\cref{fig:gtbelow}, left). Under the pre-specified rule the Axis~B statement stays as written. At matched measured mover fractions (\cref{fig:gtbelow}, right) the egocentric clips and the synthetic sequences disagree in sign: synthetic $+5.6\%$ [$-1.2$, $+12.9$], $+6.3\%$ [$-1.1$, $+14.0$], $+3.4\%$ [$-3.7$, $+11.2$] and $+3.1\%$ [$-1.3$, $+7.9$] in the bins 0.05 to 0.10, 0.10 to 0.20, 0.20 to 0.30 and 0.30 to 0.50, against $+0.9\%$ [$-4.5$, $+6.4$], $-14.3\%$ [$-20.1$, $-8.8$], $-7.0\%$ [$-14.3$, $-0.1$] and $+11.9\%$ [$+2.3$, $+21.2$] on the egocentric movers (21, 32, 20 and 5 clips; the 27 clips whose measured fraction is under 0.05 change by $-0.2\%$ [$-4.0$, $+3.6$]), so the matched-fraction reading is not replicated and the egocentric mover result is reported as a discrepancy between real and synthetic movers at low fractions. The Gate~5 addendum verdict is unchanged.

\begin{figure}[t]
  \centering
  \includegraphics[width=\linewidth]{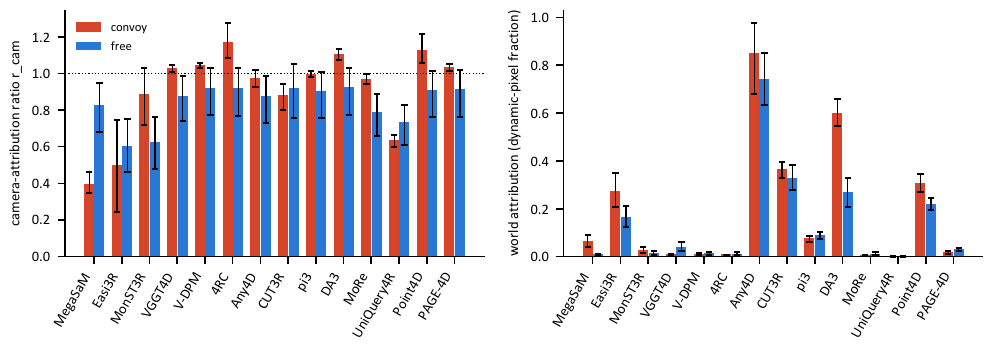}
  \caption{Rigid-world analogue on nuScenes-mini (Amendment~13, A13.5). Camera-attribution ratio (left, LiDAR depth scale; dotted line at 1) and world attribution (right) per model on the seven convoy clips (red) and the twenty free-driving clips (blue), with 95\% intervals over clips.}
  \label{fig:nuscenes}
\end{figure}

\paragraph{Rigid-World Analogue (Amendment~13, A13.5).} Axis~A has no counterpart with ground truth in the two sets above. nuScenes-mini~\cite{caesar2020nuscenes} (ten scenes with ego poses, LiDAR and 3D boxes, downloaded under its terms of use without a login) was scanned with the rules of the driving set for windows in which the vehicles ahead move with the ego vehicle: 30 convoy windows and no traffic-jam window met the criteria in the ten scenes, and seven convoy clips and twenty free-driving clips of 30 frames were exported at the synthetic resolution (measured dynamic-pixel fraction 0.041 and 0.035, ego speed 13.3 and 8.4\,m/s). The fourteen pose models ran as released (378 runs, none failed; the three models added under Amendment~16 enter the figure and not the pooled numbers, which are over the eleven pre-registered models and their 297 runs); $r_\mathrm{cam}$ takes its depth scale from the frame-0 LiDAR returns on the vehicle pixels (Amendment~4). Pooled over models the camera-attribution ratio is 0.91 [0.84, 0.96] on the convoy clips against 0.84 [0.79, 0.88] on free driving, and world attribution 0.21 against 0.15. Per model (\cref{fig:nuscenes}) the ratio is higher on the convoy clips with an interval excluding zero for six of eleven (MonST3R $+0.26$, 4RC $+0.25$, Depth Anything~3 and MoRe $+0.18$, VGGT4D $+0.15$, V-DPM $+0.12$) and lower for one, MegaSaM (0.40 against 0.83, $-0.43$ [$-0.56$, $-0.27$]), which hands part of the ego travel to the world in the way the synthetic rigid-world axis anticipates; world attribution rises on the convoy clips for four of eleven (Depth Anything~3 by the most, 0.27 to 0.60). The reading is descriptive, as A13.5 pre-registered: on real convoy clips most feed-forward models attribute more travel to the camera, not less, and the one optimisation-based model attributes less.

\begin{figure}[t]
  \centering
  \includegraphics[width=\columnwidth]{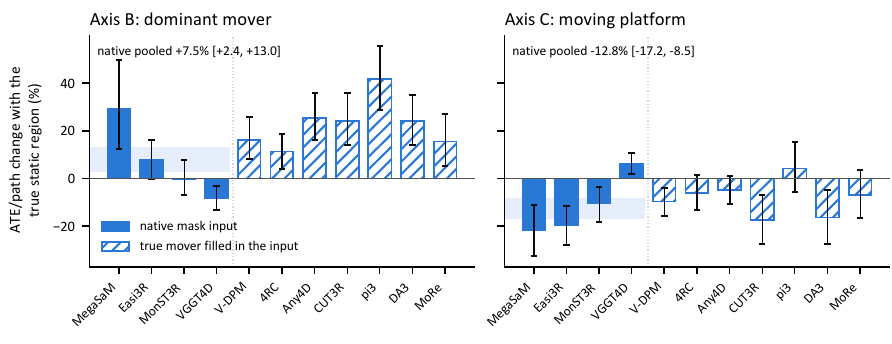}
  \caption{The true static region as input. Change of camera ATE/path when the ground-truth static region is supplied, per model with 95\% intervals, on Axis~B above each model's degradation threshold (left) and on Axis~C (right); filled bars are the four models with a native mask input, hatched bars the seven in which the true mover is filled in the input; the band is the native-mask pooled interval.}
  \label{fig:regimes_supp}
\end{figure}

\section{Training-Side Intervention}
\label{app:finetune}
\begin{figure}[t]
  \centering
  \includegraphics[width=\linewidth]{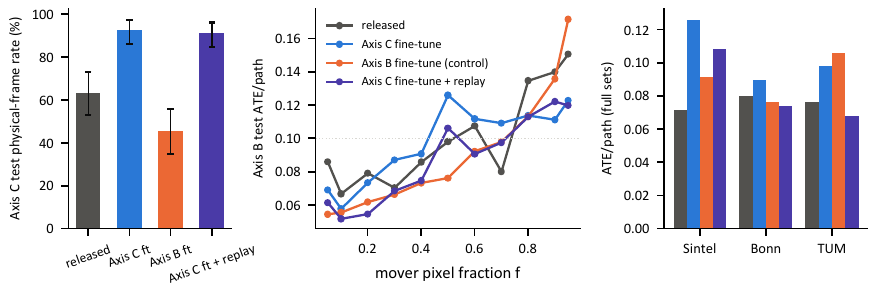}
  \caption{Training-side test (MonST3R). Released, fine-tuned on moving-platform sequences, and fine-tuned on dominant-mover sequences (control), evaluated once on the test splits: Axis~C physical-frame rate (left), Axis~B ATE/path against $f$ (middle), and the full Sintel, Bonn and TUM sets (right).}
  \label{fig:finetune}
\end{figure}

Under Amendment~11, MonST3R was chosen because its training code is public, its inference is fast enough for the test splits and it has a native mask input. Two training sets of 400 sequences each: the Axis~C validation split (80) plus 320 Axis~C-style sequences rendered with new scene seeds from the validation asset pool, and, for the control, the Axis~B validation split plus Axis~B-style renders (440, cut to 400 by a seeded permutation). Recipe: the authors' training command (DUSt3R trainer, encoder frozen, ConfLoss over the Regr3D loss, AdamW with betas 0.9 and 0.95 and weight decay 0.05, learning rate $5\times10^{-5}$ with cosine decay to $10^{-6}$, one warm-up epoch, ColorJitter and the three 512-pixel crop resolutions of their command, frame strides 1 to 9), batch 4 with 4 accumulation steps per process as in their per-GPU setting (effective batch 16 on one GPU against their 32 on two), 3000 optimizer steps in 10 epochs, seed 0, no selection: the last checkpoint is the fine-tune, and a 20-sequence slice of the training set logs a monitor loss and selects nothing. Each fine-tune took about 4\,h on one A6000. The three checkpoints were evaluated once on the Axis~B and Axis~C test splits (176 and 80 sequences) and on the full Sintel, Bonn and TUM sets through the unchanged harness (the fine-tuned weights substituted through an environment variable recorded in every run). \Cref{tab:ft} gives the numbers; \cref{fig:finetune} plots them. Reading: the Axis~C fine-tune raises the Axis~C test physical-frame rate by $+29.1$ points [$+19.0$, $+40.5$] and leaves the Axis~B degradation threshold within the released interval, as the two-regime reading predicts, but raises Sintel ATE/path by $98\%$ [$+20$, $+285$], Bonn by $12\%$ [$-3$, $+24$] and TUM by $28\%$ [$-26$, $+161$], so the pre-registered support criterion (all three within 2\%) is not met and the result is reported as observed. The control lowers the Axis~C physical-frame rate by $17.1$ points [$3.9$, $31.6$], lowers Axis~B ATE/path by $9.2\%$ [$1.7$, $16.2$] and moves Sintel by $+26\%$ [$-16$, $+138$]. Eight evaluation runs failed deterministically in MonST3R's pose initialiser (SQPnP assertion or singular matrix) and are kept as results; paired comparisons use the sequences both checkpoints completed.

\begin{figure}[t]
  \centering
  \includegraphics[width=0.8\linewidth]{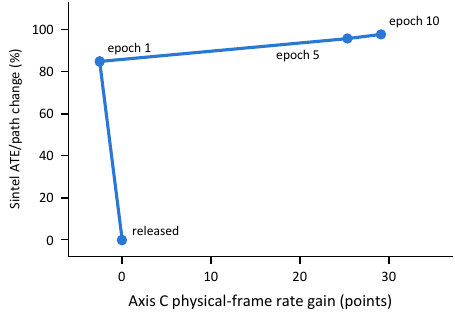}
  \caption{Forgetting curve of the Axis~C fine-tune: Sintel ATE/path change against the Axis~C test physical-frame gain, both paired against the released checkpoint, at epochs 1, 5 and 10.}
  \label{fig:forget}
\end{figure}

\paragraph{Replay-Mixed Fine-Tune (Amendment~12).} The regression is the forgetting expected of a short fine-tune on rendered data with no replay, so one further run, the last training run of the study, repeats the Axis~C fine-tune with half of every epoch's samples drawn from PointOdyssey's training split through the authors' own loader (a seeded subset of 40 of the 109 usable training videos; v1.2, CC BY-NC-SA 4.0); recipe, steps and seed are unchanged. Its pre-specified reading restates the regression bound for the sample sizes at hand (a point estimate of at most $+10\%$ with an interval including zero on each of Sintel, Bonn and TUM, with the 20-point frame-choice gain and the unchanged threshold as before). Result: Axis~C test physical-frame rate 0.91 [0.85, 0.96] ($+28.2$ points [$+17.9$, $+39.7$] over the released checkpoint), Axis~C ATE/path $-39\%$, Axis~B ATE/path $-13\%$ [$-19$, $-6$] with the degradation threshold unchanged (relative reading none, absolute 0.5), mask flip point 0.80; Sintel $+52\%$ [$+15$, $+111$], Bonn $-8\%$ [$-17$, $+0$], TUM $-11\%$ [$-43$, $+26$]. Bonn and TUM meet the bound and Sintel does not, so the run is reported as observed and not as a repair (\cref{tab:ft}, \cref{fig:finetune}). Two evaluation runs failed deterministically (one SQPnP assertion, one singular matrix). The forgetting curve of the un-replayed Axis~C fine-tune (\cref{fig:forget}; epochs 1, 5 and 10 on the Axis~C test split and Sintel) shows the Sintel regression in place after the first epoch ($+85\%$ [$+15$, $+214$] at a frame-choice gain of $-2.5$ points) and the frame-choice gain arriving later ($+25.3$ points at epoch 5, $+29.1$ at epoch 10, Sintel $+96\%$ and $+98\%$): the two effects are not one trade-off along the run, the forgetting comes first.

\begin{figure}[t]
  \centering
  \includegraphics[width=\linewidth]{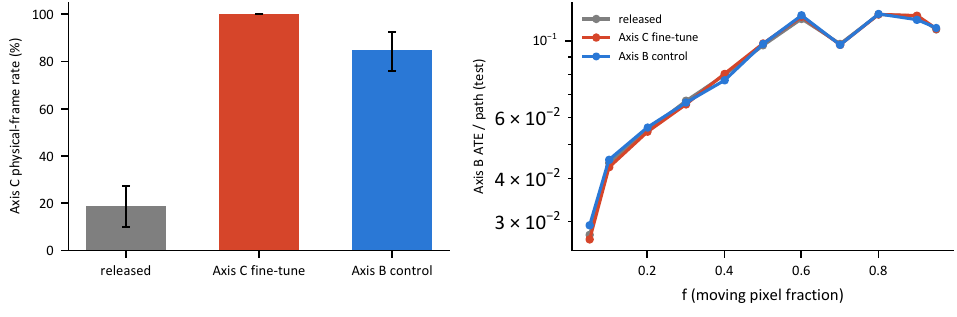}
  \caption{Second architecture (MoRe). Left: Axis~C test physical-frame rate for the released checkpoint, the moving-platform fine-tune and the dominant-mover control (95\% CIs). Right: Axis~B test ATE/path against the mover fraction for the three checkpoints; the curves coincide.}
  \label{fig:finetune2}
\end{figure}

\paragraph{Second Architecture (Amendment~13).} The replication asks for a model whose training code is public, whose dynamic mask is usable at small movers (IoU at $f \le 0.10$ of at least 0.6) and whose family differs from MonST3R's pointmap regression. In the pre-registered preference order 4RC and V-DPM have no released training code; MoRe~\cite{more2025} (a frozen VGGT-style aggregator with camera, depth, point and motion heads; IoU 0.644 at $f \le 0.10$) is the first candidate that qualifies. The same two 400-sequence sets, the authors' recipe (aggregator frozen, AdamW at $10^{-6}$ with warm-up and cosine decay, weight decay 0.05, bf16, camera, depth, point, motion and attention losses), 3000 steps and seed 0 were used, from the released weights with a fresh optimizer; on a 48\,GB GPU the authors' 36-image chunks do not fit and 12-image chunks with two accumulation chunks were used (the one deviation, recorded in the run config). Three in-memory shims over the released code (an f-string fix for Python 3.11, the backend's attention-saving shim, the attention loss restricted to patch-token keys) leave the checkout untouched. Both checkpoints and the released one were evaluated once on the Axis~B and C test splits and on Sintel, Bonn and TUM (\cref{fig:finetune2}). The moving-platform tune raises the Axis~C test physical-frame rate from 0.19 [0.10, 0.28] to 1.00 ($+81.2$ points [$+72.5$, $+90.0$], paired over 80 sequences) with the Axis~C error unchanged ($-0.2\%$ [$-1.4$, $+0.8$]); the dominant-mover control raises it to 0.85 ($+66.2$ [$+56.2$, $+76.2$]). Neither checkpoint moves the Axis~B degradation threshold (relative reading: no crossing for the released checkpoint, interval [0.20, 0.90], and no crossing for either fine-tune; absolute reading 0.60 for all three; Axis~B error $-0.1\%$ [$-1.0$, $+0.7$] and $+0.1\%$ [$-0.8$, $+1.0$]). Regression, reported without a bound as A13.1 asks: the Axis~C tune changes Sintel by $+0.8\%$ [$-1.6$, $+3.8$], Bonn by $-0.8\%$ [$-2.1$, $+1.2$] and TUM by $-2.9\%$ [$-7.1$, $+4.0$]; the control changes Sintel by $+4.9\%$ [$+0.6$, $+9.7$], Bonn by $-0.6\%$ [$-4.7$, $+1.7$] and TUM by $+2.7\%$ [$-2.4$, $+15.9$]. Under the pre-specified reading the dissociation does not replicate: the gain and the unchanged threshold hold, the control raises the rate as well. The pattern is consistent with the frame choice of MoRe living in its motion head, which either rendered set teaches to mark the camera-attached foreground as the moving part; the aggregator, frozen in the authors' recipe, cannot be the carrier.

\begin{table*}[t]
\centering
\footnotesize
\caption{Training-side test (Amendment~11), one evaluation on the test splits and the full existing sets. Axis C physical-frame rate with its bootstrap interval and ATE/path over 80 test sequences; Axis B degradation threshold $f_\mathrm{deg}$ (relative reading with interval, absolute reading) and mask flip point over 176 test sequences; ATE/path on Sintel, Bonn and TUM (the released checkpoint's existing-set numbers are the Phase 2 runs).}
\label{tab:ft}
\begin{tabular}{lllllllll}
\toprule
Checkpoint & Axis C phys. [CI] & Axis C ATE/path & $f_\mathrm{deg}$ rel. [CI] & $f_\mathrm{deg}$ abs. & $f_\mathrm{mask}$ & Sintel & Bonn & TUM \\
\midrule
released & 0.63 [0.53, 0.73] & 0.174 & none [0.50, 0.95] & 0.60 & 0.50 & 0.071 & 0.080 & 0.076 \\
Axis C fine-tune & 0.93 [0.86, 0.97] & 0.102 & none [0.30, 0.95] & 0.50 & 0.95 & 0.126 & 0.089 & 0.098 \\
Axis B fine-tune (control) & 0.45 [0.35, 0.56] & 0.181 & 0.95 [0.40, 0.95] & 0.80 & none & 0.092 & 0.076 & 0.106 \\
Axis C fine-tune + replay & 0.91 [0.85, 0.96] & 0.106 & none [0.40, 0.95] & 0.50 & 0.80 & 0.108 & 0.074 & 0.068 \\
\bottomrule
\end{tabular}
\end{table*}

\begin{figure}[t]
  \centering
  \includegraphics[width=\linewidth]{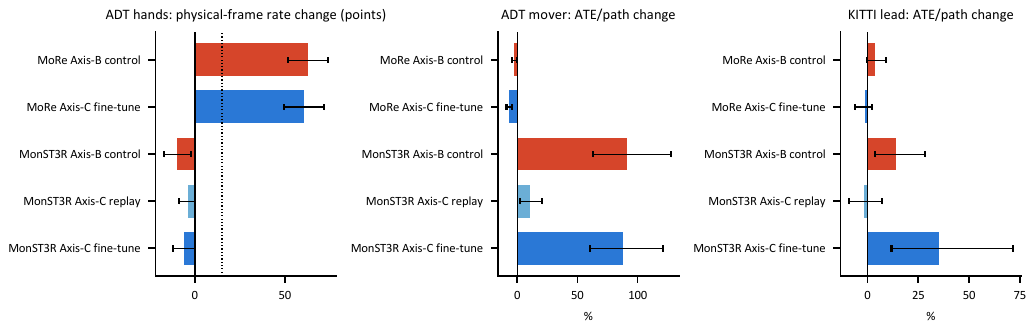}
  \caption{Real-video transfer of the fine-tunes (Gate 6e). Change against the released checkpoint of the same model, paired over clips with 95\% intervals: egocentric hands physical-frame rate in points (left; the dotted line is the 15-point margin of A15.3), egocentric mover ATE/path (middle) and leading-vehicle ATE/path (right), both relative, positive is worse.}
  \label{fig:finetune_transfer}
\end{figure}

\paragraph{Real-Video Transfer (Amendment~15).} A15.2 evaluates the five fine-tuned checkpoints (MonST3R moving-platform, replay and dominant-mover control; MoRe moving-platform and control) as released checkpoints would be, through the unchanged real-video code and without ground-truth masks, on the 81 egocentric hands clips (the wearer's hands as a camera-attached foreground, physical-frame rate), the 105 egocentric mover clips and the 41 leading-vehicle clips (ATE/path), each once, against the released checkpoint of the same model (paired bootstrap over clips, \cref{fig:finetune_transfer}). A15.3 reads the gain as transferring when the moving-platform tune raises the hands physical-frame rate by at least 15 points with the interval excluding zero and the mover and leading-vehicle error shows no detected increase, and attributes it to physical-frame motion supervision in general when the control rises by the same margin, else to moving-platform data. MoRe: the moving-platform tune raises the hands rate from 0.31 to 0.91 ($+60.5$ points [$+49.4$, $+71.6$]) with mover $-6.7\%$ [$-9.3$, $-4.4$] and leading-vehicle $-1.3\%$ [$-6.2$, $+2.0$] (transfers), the control raises it to 0.94 ($+63.0$ [$+51.9$, $+74.1$]) with mover $-2.2\%$ [$-4.1$, $-0.0$] and leading-vehicle $+3.6\%$ [$-0.1$, $+8.7$]; both raise the mover mask IoU ($+9.8$ and $+13.6$ points). MonST3R: the moving-platform tune changes the hands rate from 0.93 by $-6.2$ points [$-12.3$, $+0.0$] with mover $+87.4\%$ [$+60.2$, $+120.5$] and leading-vehicle $+35.4\%$ [$+11.6$, $+71.7$] (does not transfer), the control by $-9.9$ [$-17.3$, $-2.5$] with mover $+91.0\%$ [$+62.7$, $+127.4$] and leading-vehicle $+14.0\%$ [$+3.5$, $+28.3$], and the replay variant by $-3.7$ [$-8.6$, $+0.0$] with mover $+10.4\%$ [$+1.7$, $+20.6$] and leading-vehicle $-1.8\%$ [$-9.3$, $+7.2$]. The released MonST3R already chooses the physical frame on 0.93 of the hands clips, so there is little room above it, and its rendered-data tunes raise its real-video camera error; the replay mix removes most of that cost. No pre-registered verdict changes (A15.3).

\paragraph{The Recipe (Amendment~17).} Gate 6e scoped the repair to models whose decomposition is a learned, supervised motion head; Amendment~17 bounds that claim with three pre-registered experiments, each fine-tune evaluated once (the single evaluation of A15.2 plus the Axis~C test split), with the paired bootstrap of Gate 6e. \emph{Data scaling (A17.2).} Fine-tunes with the Gate 6b recipe (3000 steps, seed 0) on seeded nested subsets of 25, 50, 100 and 200 of the 400 Axis~C sequences (one permutation, the first $n$); the reading is the smallest size at which the gain on the 81 egocentric hands clips is at least 45 points with an interval excluding zero and no paired ATE/path change on the real mover and leading-vehicle clips or on Sintel, Bonn and TUM has an interval above zero. The reading is 25 sequences: the hands gain is $+60.5$ points at 25 and at 400 (74 of 81 clips physical) and $+59.3$ at 50, 100 and 200 (73 of 81), the Axis~C test rate 1.00 at every size, and no paired change on the mover and leading-vehicle clips or on Sintel, Bonn and TUM has an interval above zero (the largest movements are improvements: the mover error $-6.2\%$ to $-7.2\%$, TUM $-5.8\%$ [$-9.7$, $-0.6$] at 25). One run per size at seed 0 is the pre-registered design, so differences between sizes are read against their intervals, which overlap throughout. Amendment~18 (2026-10-15) adds 1 and 5 sequences from the same permutation as two further points on the curve, with no reading attached: one sequence raises the hands rate to 0.93 ($+61.7$ points [$+51.8$, $+71.6$]) and five to 0.94 ($+63.0$ [$+51.9$, $+74.1$]), with no paired change elsewhere above zero (one sequence: mover $+1.2\%$ [$-0.5$, $+3.1$], leading vehicle $+0.5\%$ [$-3.5$, $+5.4$], Sintel $+1.6\%$ [$-0.8$, $+3.5$]; five: mover $-3.1\%$ [$-5.1$, $-0.9$], leading vehicle $+2.5\%$ [$-0.1$, $+6.2$], Sintel $+1.1\%$ [$-1.0$, $+3.4$]); \cref{tab:recipe} and Figure~4 of the main text carry them. \emph{Scope (A17.3).} The pre-registered rule takes the first zoo model with public training code and a learned motion output (Any4D first, then the Amendment~16 models) and the first with training code and no learned motion output ($\pi^3$ or CUT3R, whichever trains first). On the check date (2026-10-14) Any4D's repository states that its training code will be released later; UniQuery4R, V-DPM, 4RC and TraceAnything ship inference code and no training code; Point4D ships inference code and has no motion head; PAGE-4D ships training code but its dynamics-aware spatial mask is an unsupervised attention mask inside the aggregator with no mask term in its loss. The motion-head half is therefore reported as no second instance available. CUT3R trains at its pinned commit (its authors' fine-tuning entry for the 512 checkpoint: encoder frozen, confidence-weighted 3D regression and RGB losses, AdamW at $10^{-6}$ with half an epoch of warm-up and cosine decay to $10^{-7}$, weight decay 0.05, batch 4 with 4 accumulation steps, bf16, 4 to 30 views per sample, truncated back-propagation in 4-frame chunks), 3000 steps in 10 epochs on the same 400 sequences; three in-memory shims over the untouched checkout (a stub for the rendering library its training code imports for tensorboard images, the monitor loader's batches moved to the device, the tensorboard renders replaced) are recorded in the run config. Evaluated once as the MoRe fine-tunes: the Axis~C test rate rises from 0.45 to 0.62 ($+17.5$ points [$+6.2$, $+30.0$]) and the test ATE/path falls by half ($-50.1\%$ [$-59.6$, $-39.4$]); on the egocentric hands clips the physical-frame rate falls from 0.85 to 0.25 ($-60.5$ points [$-71.6$, $-48.1$]), the mover error rises $+82.5\%$ [$+62.6$, $+108.4$] and the mover mask IoU falls from 0.20 to 0.13 ($-7.1$ points [$-8.8$, $-5.4$]), the leading-vehicle error $+20.7\%$ [$-3.1$, $+60.2$], Sintel $+42.1\%$ [$+4.5$, $+108.4$], Bonn $+33.7\%$ [$-1.7$, $+76.5$] and TUM $+69.5\%$ [$+25.8$, $+154.7$]. The model without a motion head does not gain 15 points and regresses, so this half of the scope condition is supported: the same data that repair MoRe's motion head move CUT3R's point-map regression toward the renderer and away from real video. \emph{Real labels (A17.4).} One MoRe fine-tune with the same recipe on the hands and mover clips of ten of the seventeen egocentric recordings (51 hands and 59 mover clips; physical-frame labels, poses and depth from the digital twin, no rendered data; the ten recordings hold 62 mover clips, and the three in recordings whose depth archive the dataset server answered with a 404 on two attempts are listed in the manifest and left out; the split chosen by a seeded rule among random 10/7 splits so that the held-out seven hold a representative share of the hands clips, 30 of 81, and of the mover clips, 43 of 105), evaluated once on the other seven recordings' clips, the Axis~C test split, the leading-vehicle clips and Sintel, Bonn and TUM. The held-out hands clips go from 0.13 to 1.00 physical ($+86.7$ points [$+73.3$, $+96.7$], 30 clips; the training recordings' own hands clips, in sample, from 0.41 to 0.94), the held-out mover mask IoU from 0.06 to 0.52 ($+46.5$ points [$+39.1$, $+54.0$]) and the held-out mover error improves ($-5.0\%$ [$-8.1$, $-2.5$]); the Axis~C test rate rises to 0.57 ($+38.8$ [$+27.5$, $+50.0$]). Three paired changes have intervals above zero: the leading-vehicle error $+11.4\%$ [$+5.0$, $+26.3$], Sintel $+4.7\%$ [$+0.9$, $+10.2$] and the Axis~C test error $+0.7\%$ [$+0.1$, $+1.4$] (Bonn $-1.1\%$ [$-3.2$, $+0.4$], TUM $+0.1\%$ [$-4.4$, $+11.2$]), so the reading is that the real-data variant does not meet the pre-specified bar. \Cref{tab:recipe} lists every evaluation.

\begin{figure}[t]
  \centering
  \includegraphics[width=\linewidth]{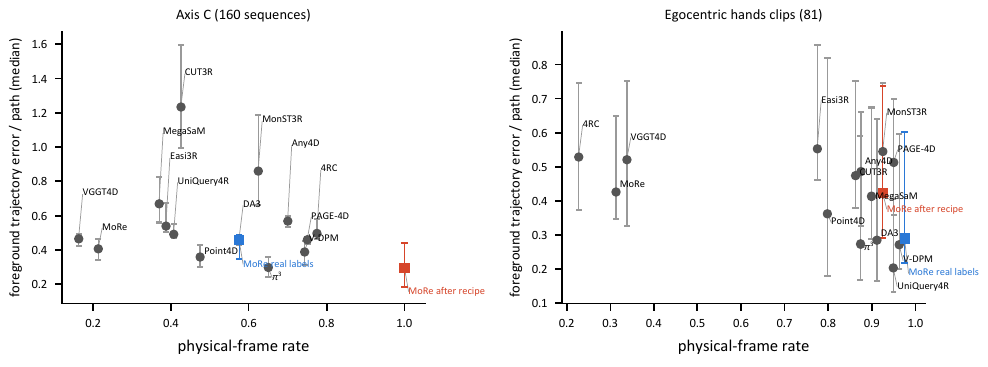}
  \caption{The downstream consequence of frame choice (A17.5). Median error of the camera-attached foreground's world trajectory (displacement from the first frame, over the camera path length, 95\% bootstrap intervals) against the physical-frame rate, per model, on Axis~C (left) and on the egocentric hands clips (right); the MoRe checkpoints after the recipe (rendered and real labels) in colour.}
  \label{fig:downstream}
\end{figure}

\paragraph{The Downstream Consequence (A17.5).} For every model on Axis~C and on the egocentric hands clips, and for the MoRe checkpoints before and after the recipe, the camera-attached foreground's trajectory in world coordinates: the ground-truth foreground pixels back-projected with the model's own depth and intrinsics and carried into the world frame by its poses, their centroid per frame, against the same centroid from the twin's depth and poses (for the egocentric clips the twin's depth images were added for this purpose). The error is the foreground's displacement from its first frame under the scale and rotation of the Sim(3) camera alignment that every ATE number uses, against the ground-truth displacement, mean over frames over the ground-truth camera path length; the median over sequences per model with a bootstrap interval (\cref{fig:downstream}). No reading is attached; the definition was fixed before the per-model numbers were read, after one change recorded in the results log (a first draft set the scale on the foreground depth of the first frame, which the depth heads get wrong by factors of 2 to 7 on the close, textureless foreground). Over the 160 Axis~C validation and test sequences, eleven of the fourteen models with poses and depth have a larger median error among their egocentric choices than among their physical choices (the exceptions are MoRe, VGGT4D and Easi3R, whose two medians are within 0.1 path lengths of each other); the largest gaps are V-DPM (1.38 against 0.26), MonST3R (1.70 against 0.71), CUT3R (1.53 against 0.83) and 4RC (1.21 against 0.44). MoRe's released checkpoint has a test-split median of 0.44 [0.33, 0.47]; the 400-sequence fine-tune 0.30 [0.18, 0.44], the 25 to 200-sequence fine-tunes 0.29 to 0.32, the dominant-mover control 0.46 [0.37, 0.48] and the real-label fine-tune 0.46 [0.35, 0.49]. On the 80 hands clips with a visible camera-attached foreground (one of the 81 has fewer than five frames with twenty valid foreground pixels and is listed in the output file), ten of fourteen models have the larger median among egocentric choices; the released MoRe's median is 0.43 [0.35, 0.65], the 400-sequence fine-tune's 0.42 [0.29, 0.74], the 25 to 200-sequence fine-tunes' 0.47 to 0.57 and the real-label fine-tune's 0.29 [0.22, 0.60]: on real video the error is dominated by the depth of the close hands, which the rendered-data recipe does not change and the real-label variant, trained with the twin's depth, does. The CUT3R fine-tune (A17.3) moves with its frame choice on both sets: on the Axis~C test split its median falls from 1.39 [1.01, 2.02] (released, physical-frame rate 0.45; 1.28 among physical choices, 1.95 among egocentric) to 0.12 [0.09, 0.15] (rate 0.62), and on the hands clips it rises from 0.47 [0.38, 0.75] (rate 0.86; 0.45 among physical choices, 0.98 among egocentric) to 1.13 [0.55, 1.85] (rate 0.25; 0.32 among the 20 physical choices, 1.43 among the 60 egocentric).

\begin{figure}[t]
  \centering
  \includegraphics[width=\linewidth]{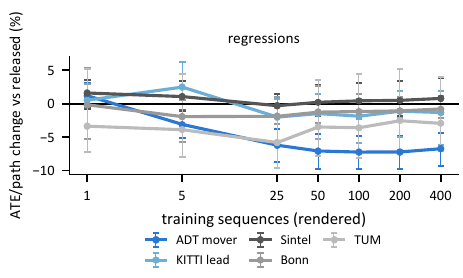}
  \caption{The recipe's regressions (A17.2, A18): the paired ATE/path changes against the released checkpoint on the real mover and leading-vehicle clips and on Sintel, Bonn and TUM after fine-tuning MoRe on 1 to 400 rendered moving-platform sequences, 95\% intervals, positive is worse.}
  \label{fig:recipe_regressions}
\end{figure}

\begin{table*}[t]
\centering
\scriptsize
\setlength{\tabcolsep}{3pt}
\caption{The recipe (Amendment~17): each fine-tune evaluated once against its released checkpoint, paired bootstrap 95\% intervals. Physical-frame rates with the change in points; ATE/path changes in percent (positive is worse). The 400-sequence MoRe row is the Gate 6b / 6e run; the rows marked A18 (1 and 5 sequences, the same permutation) extend the A17.2 curve under Amendment~18 and carry no reading; the real-label row is evaluated on the seven held-out recordings; the A19.2 rows are single sequences from other seeded permutations and the A19.3 rows the VGGT4D motion head (Amendment~19).}
\label{tab:recipe}
\resizebox{\textwidth}{!}{\begin{tabular}{lllllllll}
\toprule
Fine-tune & Axis C test phys. & ADT hands phys. & ADT mover ATE & KITTI lead ATE & Sintel & Bonn & TUM \\
\midrule
MoRe, 1 rendered (A18) & 1.00 (+81.2 [+72.5, +90.0]) & 0.93 (+61.7 [+51.8, +71.6]) & +1.2 [-0.5, +3.1] & +0.5 [-3.5, +5.4] & +1.6 [-0.8, +3.5] & -0.2 [-5.2, +3.0] & -3.3 [-7.2, +5.2] \\
MoRe, 5 rendered (A18) & 1.00 (+81.2 [+72.5, +90.0]) & 0.94 (+63.0 [+51.9, +74.1]) & -3.1 [-5.1, -0.9] & +2.5 [-0.1, +6.2] & +1.1 [-1.0, +3.4] & -1.9 [-5.7, +0.5] & -3.9 [-7.9, +4.4] \\
MoRe, 25 rendered & 1.00 (+81.2 [+72.5, +90.0]) & 0.91 (+60.5 [+49.4, +71.6]) & -6.2 [-8.7, -3.8] & -2.0 [-6.1, +0.8] & -0.3 [-1.8, +1.5] & -1.9 [-3.1, +1.1] & -5.8 [-9.7, -0.6] \\
MoRe, 50 rendered & 1.00 (+81.2 [+72.5, +90.0]) & 0.90 (+59.3 [+48.1, +70.4]) & -7.1 [-9.8, -4.6] & -1.5 [-5.3, +1.4] & +0.2 [-1.9, +2.8] & -1.3 [-2.9, +2.7] & -3.5 [-7.8, +3.5] \\
MoRe, 100 rendered & 1.00 (+81.2 [+72.5, +90.0]) & 0.90 (+59.3 [+48.1, +70.4]) & -7.2 [-9.8, -4.8] & -1.9 [-6.0, +0.8] & +0.4 [-1.6, +3.1] & -1.2 [-2.3, +1.5] & -3.6 [-8.1, +4.5] \\
MoRe, 200 rendered & 1.00 (+81.2 [+72.5, +90.0]) & 0.90 (+59.3 [+48.1, +70.4]) & -7.2 [-9.8, -4.9] & -1.1 [-4.9, +1.9] & +0.5 [-1.8, +3.5] & -1.1 [-2.1, +1.0] & -2.5 [-6.9, +5.2] \\
MoRe, 400 rendered & 1.00 (+81.2 [+72.5, +90.0]) & 0.91 (+60.5 [+49.4, +71.6]) & -6.7 [-9.3, -4.4] & -1.3 [-6.2, +2.0] & +0.8 [-1.6, +3.8] & -0.8 [-2.1, +1.2] & -2.9 [-7.1, +4.0] \\
MoRe, 110 real clips (ten recordings) & 0.57 (+38.8 [+27.5, +50.0]) & 1.00 (+86.7 [+73.3, +96.7]) (held out) & -5.0 [-8.1, -2.5] (held out) & +11.4 [+5.0, +26.3] & +4.7 [+0.9, +10.2] & -1.1 [-3.2, +0.4] & +0.1 [-4.4, +11.2] \\
CUT3R, 400 rendered & 0.62 (+17.5 [+6.2, +30.0]) & 0.25 (-60.5 [-71.6, -48.1]) & +82.5 [+62.6, +108.4] & +20.7 [-3.1, +60.2] & +42.1 [+4.5, +108.4] & +33.7 [-1.7, +76.5] & +69.5 [+25.8, +154.7] \\
MoRe, 1 rendered, seed-1 sequence (A19.2) & 0.90 (+71.2 [+61.2, +81.2]) & 0.93 (+61.7 [+50.6, +72.8]) & +2.7 [+0.5, +5.3] & +0.5 [-3.6, +3.8] & +6.5 [+1.6, +13.2] & -4.0 [-11.1, +1.9] & +1.0 [-7.8, +12.7] \\
MoRe, 1 rendered, seed-2 sequence (A19.2) & 0.54 (+35.0 [+23.8, +46.2]) & 0.56 (+24.7 [+14.8, +34.6]) & -4.2 [-7.4, -0.9] & +0.6 [-3.3, +4.3] & +3.0 [-2.9, +10.4] & -1.7 [-10.7, +4.6] & +3.7 [-5.4, +22.0] \\
MoRe, 1 Axis B sequence, control (A19.2) & 0.35 (+16.2 [+7.5, +25.0]) & 0.37 (+6.2 [-1.2, +13.6]) & +6.4 [+3.0, +10.2] & +2.8 [-0.3, +5.6] & -1.1 [-3.2, +0.5] & -4.4 [-13.7, +2.0] & -2.6 [-7.1, +4.9] \\
VGGT4D, 25 rendered, motion head (A19.3) & 1.00 (+86.2 [+77.5, +93.8]) & 0.88 (+54.3 [+43.2, +65.4]) & +4.8 [+2.8, +7.1] & +2.0 [-2.4, +6.2] & +9.8 [-3.8, +29.7] & +5.3 [-4.8, +28.4] & +3.9 [+1.8, +7.8] \\
VGGT4D, 1 rendered, motion head (A19.3) & 0.85 (+71.2 [+58.8, +82.5]) & 0.86 (+53.1 [+40.7, +64.2]) & +7.1 [+4.6, +10.1] & -5.1 [-12.1, +4.3] & +11.6 [-3.3, +32.9] & +13.2 [-0.8, +49.8] & +20.8 [+12.6, +30.6] \\
VGGT4D, 25 Axis B, motion head, control (A19.3) & 0.96 (+82.5 [+73.8, +90.0]) & 0.85 (+51.9 [+39.5, +64.2]) & +6.9 [+2.4, +11.9] & -1.9 [-5.9, +2.4] & +14.8 [+0.3, +43.0] & +7.7 [-2.6, +40.4] & +14.3 [+7.1, +24.6] \\
VGGT4D, 25 rendered, head and backbone (A19.3) & 1.00 (+86.2 [+77.5, +93.8]) & 0.42 (+8.6 [-6.2, +22.2]) & +24.8 [+17.1, +32.8] & +6.7 [+0.5, +14.0] & +76.9 [+29.3, +167.6] & +14.0 [-11.7, +59.9] & +15.4 [+8.8, +27.2] \\
MoRe, 5 rendered, seed-1 draw (A20.3) & 0.97 (+78.8 [+70.0, +87.5]) & 0.91 (+60.5 [+49.4, +71.6]) & +5.1 [+3.1, +7.7] & -2.1 [-5.0, +0.2] & +0.8 [-2.1, +4.4] & -3.7 [-16.3, +3.7] & -3.8 [-8.2, +3.6] \\
MoRe, 5 rendered, seed-2 draw (A20.3) & 0.86 (+67.5 [+57.5, +77.5]) & 0.90 (+59.3 [+49.4, +70.4]) & +0.8 [-1.1, +2.8] & +0.3 [-1.1, +2.1] & +2.1 [-1.2, +6.0] & -1.9 [-8.7, +2.3] & +1.2 [-6.8, +17.8] \\
MoRe, 5 rendered, seed-3 draw (A20.3) & 0.99 (+80.0 [+70.0, +88.8]) & 0.91 (+60.5 [+49.4, +71.6]) & -9.7 [-12.5, -7.1] & +0.4 [-3.9, +3.4] & +1.3 [-1.5, +5.0] & -1.1 [-3.9, +4.2] & -1.1 [-6.4, +8.2] \\
MoRe, 5 Axis B sequences, control (A20.3) & 0.68 (+48.8 [+37.5, +60.0]) & 0.67 (+35.8 [+25.9, +46.9]) & -0.4 [-2.9, +2.4] & +5.1 [+1.5, +11.9] & +3.5 [-0.4, +7.5] & -1.8 [-8.4, +1.9] & +1.5 [-7.1, +19.5] \\
\bottomrule
\end{tabular}}
\end{table*}

\paragraph{The Recipe's Two Bounds (Amendment~19).} Amendment~19 (2026-10-15) bounds the recipe on the two sides a reader will probe, with the readings and the paper's wording for each outcome fixed before the runs. \emph{Sequence dependence (A19.2).} The one-sequence point of Amendment~18 is the first sequence of one seeded permutation. Three further MoRe fine-tunes with the Gate 6b recipe train on one sequence each: the first Axis~C sequence of the seed-1 and of the seed-2 permutation of the 400 (a train scene at $g=0.70$ and an escalator scene at $g=0.30$; the Amendment~18 sequence is an elevator scene at $g=0.30$) and, as the one-sequence control, the first sequence of the seed-0 permutation of the Gate 6b dominant-mover set (a person at $f=0.30$); each is evaluated once as the A17.2 runs are. The reading is the number of the three single Axis~C sequences that meet the A17.2 bar (a hands gain of at least 45 points with an interval excluding zero and no detected regression): one of three. The elevator sequence gives $+61.7$ points [$+51.8$, $+71.6$] (0.31 to 0.93) with no detected regression; the train sequence $+61.7$ [$+50.6$, $+72.8$] (0.93) with the mover error $+2.7\%$ [$+0.5$, $+5.3$] and Sintel $+6.5\%$ [$+1.6$, $+13.2$]; the escalator sequence $+24.7$ [$+14.8$, $+34.6$] (0.56) with no detected regression. The control moves the hands rate by $+6.2$ points [$-1.2$, $+13.6$] (0.37) and the Axis~C test rate by $+16.2$ [$+7.5$, $+25.0$], so at one sequence the A15.3(ii) attribution is to moving-platform data, where at 400 sequences the same control reached 0.94. The pre-specified consequence applied: the abstract states five sequences (the Amendment~18 point) and the main text reports the count. \emph{Second architecture (A19.3).} The choice was made by a rule fixed before the check and recorded before any training: 4RC if its repository carried public training code for its motion output on the check date, otherwise VGGT4D with a motion head added. On 2026-10-15 4RC's repository lists its training code as to be released, so VGGT4D is the instance. The head is a DPT head of VGGT's depth-head architecture on the aggregator tokens, initialised from the released depth head with a fresh one-channel output, trained with a per-pixel cross-entropy on the rendered physical-frame motion labels (MoRe's loss, with its confidence filter) under the A11.2 schedule (AdamW at $10^{-6}$, 3000 steps, 12-image chunks, seed 0) with VGGT's weights frozen; at inference its mask (logit above zero, untuned) replaces the attention-derived mask at the entry VGGT4D consumes, the one the ground-truth-mask pass of Gate 3b used, with everything else as released, and frame choice, mask IoU, ATE/path and the downstream error come from the unchanged evaluation code. Four runs: the head on the 25-sequence and the one-sequence subsets of A17.2 and Amendment~18, the head on 25 dominant-mover sequences as the control, and one optional run with the aggregator unfrozen on the 25 (it fits a 48\,GB card at 24.5\,GB). Readings against the released VGGT4D, paired over clips: (i) at 25 sequences the hands physical-frame rate rises from 0.33 to 0.88 ($+54.3$ points [$+43.2$, $+65.4$]); (ii) the mover mask IoU rises $+7.4$ points [$+2.5$, $+12.2$] (0.15 to 0.22), the mover error $+4.8\%$ [$+2.8$, $+7.1$], the leading-vehicle error $+2.0\%$ [$-2.4$, $+6.2$], Sintel $+9.8\%$ [$-3.8$, $+29.7$], Bonn $+5.3\%$ [$-4.8$, $+28.4$] and TUM $+3.9\%$ [$+1.8$, $+7.8$], two intervals above zero, so (ii) is not met; (iii) at one sequence the hands rate reaches 0.86 ($+53.1$ [$+40.7$, $+64.2$]) with the mover error $+7.1\%$ [$+4.6$, $+10.1$] and TUM $+20.8\%$ [$+12.6$, $+30.6$]; (iv) the control head reaches 0.85 ($+51.9$ [$+39.5$, $+64.2$]), so the attribution is to physical-frame motion supervision in general, as for MoRe; (v) the Axis~C test rate is 1.00 at 25 sequences ($+86.2$ [$+77.5$, $+93.8$]) with the test error $+11.9\%$ [$+3.8$, $+21.7$], and the downstream medians (A17.5) are 0.49 on the test split and 0.51 on the hands clips (released 0.52). The verdict is not a second positive instance, and the claim stays stated for MoRe. The pattern matches the ground-truth-mask pass, where the true static region raised VGGT4D's error by $+6.0\%$ on Axis~C and $+6.4\%$ on the hands clips: VGGT4D's camera stage does not profit from a better mask, so a head that fixes the decomposition leaves the pose where it was or a few percent worse. The unfrozen run separates the two carriers: with the aggregator trained under the same mask loss the Axis~C test rate reaches 1.00 and the test error rises $+51.2\%$ [$+34.4$, $+73.9$], the hands rate stays at 0.42 ($+8.6$ [$-6.2$, $+22.2$]), the mover error rises $+24.8\%$ [$+17.1$, $+32.8$] and Sintel $+76.9\%$ [$+29.3$, $+167.6$], the renderer overfit of CUT3R (A17.3) on a third architecture. \Cref{tab:recipe} lists every evaluation and Figure~4 of the main text carries the single-sequence markers.

\section{Attention and Object Motion}
\label{app:analysis}
\paragraph{Where Does the Flip Happen?} For VGGT4D, the one attention-based model whose camera token we can read, we recorded the camera token's attention over image patches on two Axis~B sequences below its mask flip and two above. The attention is close to uniform in every global block (normalised entropy 0.996 to 0.997); in the middle blocks (10 to 16) the camera token attends less per mover patch than per static patch, and more so above the flip (minimum ratio 0.12 to 0.26 against 0.27 to 0.29 below), and no head attends preferentially to the mover (maximum per-head ratio 1.13 to 1.33). The flip is not a failure of attention to find the mover; it is a decision taken with the mover in view. A pose effect can show in ATE just when the mover's displacement differs materially from the camera's; 89\% of Axis~B sequences have a displacement ratio above 0.5 and the error rises with $f$ in every stratum (\cref{app:rho}).

\clearpage
\section{Release: Leaderboard and Test Split}
\label{app:release}
The benchmark page of the repository (description, evaluation script and output format, leaderboard regenerated from the results, opt-out contact) is released upon publication. Its leaderboard table (\cref{tab:leaderboard}) and its test-split sequence list are reproduced here as they stand. Two versions of the benchmark exist: the pre-registered release (1248 sequences, 624 validation and 624 test; every verdict of the paper) and the extended release, whose test split below holds 800 sequences (the 624 plus the test half of the 352 Axis~B sequences added under Amendment~13) and which also holds the 120 sequences of Axis~B$'$ (the 100 of Amendment~20 and their 20 opposite-spin renders of Amendment~22).

% generated by scripts/make_supp_release.py from docs/leaderboard.json; do not edit
\begin{table*}[t]
  \centering
  \scriptsize
  \setlength{\tabcolsep}{3pt}
  \caption{The leaderboard of the benchmark page as it stands (generated 2026-10-14 from 10554 Axis~B, 2392 Axis~C and 2774 existing-benchmark runs; Axis~B flip points at sixteen seeds per cell, bootstrap intervals in brackets; none marks no crossing up to $f = 0.95$). Commits are the pinned evaluation commits.}
  \label{tab:leaderboard}
  \resizebox{\textwidth}{!}{\begin{tabular}{llllllllllll}
    \toprule
    model & commit & $f_\mathrm{mask}$ & $f_\mathrm{deg}$ rel. & $f_\mathrm{deg}$ abs. & ATE/path $f \le 0.10$ & ATE/path $f \ge 0.50$ & physical frame & Axis C ATE/path & Sintel & Bonn & TUM \\
    \midrule
    MonST3R & \texttt{574cc77} & 0.50 [0.50, 0.60] & none [0.90, 0.95] & 0.80 [0.40, 0.80] & 0.060 [0.051, 0.069] & 0.118 [0.112, 0.124] & 62.4 [54.8, 70.1] \% & 0.174 [0.158, 0.189] & 0.071 [0.038, 0.110] & 0.080 [0.053, 0.100] & 0.076 [0.041, 0.114] \\
    CUT3R & \texttt{8bc15dc} & none [0.23, 0.50] & none [none, none] & 0.20 [0.10, 0.20] & 0.091 [0.082, 0.101] & 0.123 [0.118, 0.129] & 42.5 [34.4, 50.0] \% & 0.168 [0.156, 0.181] & 0.100 [0.059, 0.146] & 0.100 [0.086, 0.119] & 0.052 [0.034, 0.069] \\
    Easi3R & \texttt{146a77d} & none [0.84, 0.95] & none [0.90, 0.94] & 0.30 [0.20, 0.40] & 0.075 [0.064, 0.088] & 0.143 [0.136, 0.149] & 38.7 [31.0, 46.5] \% & 0.218 [0.204, 0.231] & 0.170 [0.119, 0.218] & 0.069 [0.043, 0.096] & 0.034 [0.019, 0.053] \\
    VGGT4D & \texttt{c7af95a} & 0.30 [0.30, 0.30] & 0.60 [0.40, 0.80] & 0.50 [0.40, 0.60] & 0.033 [0.027, 0.040] & 0.116 [0.111, 0.122] & 16.2 [10.6, 21.9] \% & 0.145 [0.129, 0.162] & 0.070 [0.038, 0.111] & 0.060 [0.028, 0.109] & 0.016 [0.009, 0.023] \\
    pi3 & \texttt{9fa3ddb} & 0.90 [0.90, 0.95] & none [0.41, 0.59] & 0.50 [0.40, 0.60] & 0.053 [0.044, 0.063] & 0.105 [0.101, 0.110] & 65.0 [57.5, 72.5] \% & 0.128 [0.114, 0.144] & 0.050 [0.023, 0.084] & 0.052 [0.029, 0.082] & 0.021 [0.010, 0.033] \\
    Any4D & \texttt{aa9f1b0} & none [none, none] & none [0.60, 0.60] & 0.30 [0.20, 0.30] & 0.056 [0.048, 0.065] & 0.121 [0.115, 0.126] & 70.0 [62.5, 76.9] \% & 0.194 [0.180, 0.209] & 0.124 [0.078, 0.172] & 0.100 [0.078, 0.123] & 0.039 [0.021, 0.060] \\
    MegaSaM & \texttt{a27b4e6} & 0.60 [0.50, 0.60] & 0.70 [0.40, 0.95] & 0.90 [0.80, 0.95] & 0.029 [0.019, 0.039] & 0.091 [0.086, 0.096] & 36.9 [30.0, 43.8] \% & 0.184 [0.167, 0.202] & 0.024 [0.002, 0.054] & 0.050 [0.034, 0.064] & 0.017 [0.009, 0.024] \\
    Depth Anything 3 & \texttt{3d835ec} & 0.80 [0.70, 0.90] & none [0.56, 0.95] & none [0.60, 0.90] & 0.042 [0.035, 0.049] & 0.088 [0.084, 0.092] & 57.5 [49.4, 65.0] \% & 0.129 [0.114, 0.145] & 0.051 [0.017, 0.094] & 0.052 [0.020, 0.102] & 0.022 [0.012, 0.034] \\
    4RC & \texttt{a08ca0a} & none [none, none] & none [0.50, 0.90] & 0.60 [0.40, 0.80] & 0.044 [0.037, 0.052] & 0.104 [0.099, 0.109] & 77.5 [71.2, 83.8] \% & 0.150 [0.134, 0.166] & 0.058 [0.022, 0.105] & 0.044 [0.021, 0.071] & 0.016 [0.009, 0.023] \\
    V-DPM & \texttt{5e2a57c} & none [0.95, 0.95] & 0.60 [0.50, 0.90] & 0.50 [0.50, 0.60] & 0.032 [0.026, 0.040] & 0.120 [0.114, 0.126] & 74.4 [67.5, 81.2] \% & 0.141 [0.126, 0.155] & 0.061 [0.028, 0.102] & 0.060 [0.025, 0.111] & 0.021 [0.010, 0.033] \\
    MoRe & \texttt{f330a34} & 0.80 [0.80, 0.90] & none [0.50, 0.90] & 0.50 [0.50, 0.60] & 0.035 [0.028, 0.042] & 0.115 [0.110, 0.121] & 21.2 [15.0, 27.5] \% & 0.137 [0.122, 0.153] & 0.042 [0.025, 0.061] & 0.062 [0.030, 0.103] & 0.019 [0.011, 0.030] \\
    UniQuery4R & \texttt{9511974} & 0.20 [0.20, 0.20] & none [0.40, 0.80] & 0.50 [0.50, 0.70] & 0.032 [0.024, 0.040] & 0.107 [0.102, 0.112] & 40.6 [33.1, 48.1] \% & 0.143 [0.128, 0.158] & 0.021 [0.011, 0.033] & 0.054 [0.022, 0.104] & 0.018 [0.010, 0.027] \\
    Point4D & \texttt{4da8153} & none [none, none] & none [none, none] & 0.20 [0.10, 0.30] & 0.076 [0.066, 0.086] & 0.131 [0.125, 0.138] & 47.5 [39.4, 55.0] \% & 0.142 [0.128, 0.156] & 0.100 [0.055, 0.150] & 0.076 [0.058, 0.094] & 0.084 [0.055, 0.117] \\
    PAGE-4D & \texttt{b8af462} & 0.90 [0.80, 0.95] & none [none, none] & 0.40 [0.30, 0.40] & 0.060 [0.052, 0.070] & 0.125 [0.119, 0.130] & 75.0 [68.1, 81.2] \% & 0.170 [0.155, 0.184] & 0.066 [0.039, 0.097] & 0.080 [0.044, 0.115] & 0.025 [0.013, 0.038] \\
    \bottomrule
  \end{tabular}}
\end{table*}

% generated by scripts/make_supp_release.py from docs/test_split.txt; do not edit
\noindent The held-out test split as released (800 sequences: Axis~A 128, Axis~B 352, Axis~C 80, Axis~D 240; the 624 of the pre-registered split and the test half of the 352 Axis~B sequences added under Amendment~13). Sequence names encode the axis, the condition and the scene seed.

\paragraph{Axis A (128).}
{\tiny\raggedright axisA\_run00\_rt\_sc2, axisA\_run00\_rt\_sc3, axisA\_run00\_tr\_sc2, axisA\_run00\_tr\_sc3, axisA\_run01\_rt\_sc2, axisA\_run01\_rt\_sc3, axisA\_run01\_tr\_sc2, axisA\_run01\_tr\_sc3, axisA\_run02\_rt\_sc2, axisA\_run02\_rt\_sc3, axisA\_run02\_tr\_sc2, axisA\_run02\_tr\_sc3, axisA\_run03\_rt\_sc2, axisA\_run03\_rt\_sc3, axisA\_run03\_tr\_sc2, axisA\_run03\_tr\_sc3, axisA\_run04\_rt\_sc2, axisA\_run04\_rt\_sc3, axisA\_run04\_tr\_sc2, axisA\_run04\_tr\_sc3, axisA\_run05\_rt\_sc2, axisA\_run05\_rt\_sc3, axisA\_run05\_tr\_sc2, axisA\_run05\_tr\_sc3, axisA\_run06\_rt\_sc2, axisA\_run06\_rt\_sc3, axisA\_run06\_tr\_sc2, axisA\_run06\_tr\_sc3, axisA\_run07\_rt\_sc2, axisA\_run07\_rt\_sc3, axisA\_run07\_tr\_sc2, axisA\_run07\_tr\_sc3, axisA\_run08\_rt\_sc2, axisA\_run08\_rt\_sc3, axisA\_run08\_tr\_sc2, axisA\_run08\_tr\_sc3, axisA\_run09\_rt\_sc2, axisA\_run09\_rt\_sc3, axisA\_run09\_tr\_sc2, axisA\_run09\_tr\_sc3, axisA\_run10\_rt\_sc2, axisA\_run10\_rt\_sc3, axisA\_run10\_tr\_sc2, axisA\_run10\_tr\_sc3, axisA\_run11\_rt\_sc2, axisA\_run11\_rt\_sc3, axisA\_run11\_tr\_sc2, axisA\_run11\_tr\_sc3, axisA\_run12\_rt\_sc2, axisA\_run12\_rt\_sc3, axisA\_run12\_tr\_sc2, axisA\_run12\_tr\_sc3, axisA\_run13\_rt\_sc2, axisA\_run13\_rt\_sc3, axisA\_run13\_tr\_sc2, axisA\_run13\_tr\_sc3, axisA\_run14\_rt\_sc2, axisA\_run14\_rt\_sc3, axisA\_run14\_tr\_sc2, axisA\_run14\_tr\_sc3, axisA\_run15\_rt\_sc2, axisA\_run15\_rt\_sc3, axisA\_run15\_tr\_sc2, axisA\_run15\_tr\_sc3, axisA\_run16\_rt\_sc2, axisA\_run16\_rt\_sc3, axisA\_run16\_tr\_sc2, axisA\_run16\_tr\_sc3, axisA\_run17\_rt\_sc2, axisA\_run17\_rt\_sc3, axisA\_run17\_tr\_sc2, axisA\_run17\_tr\_sc3, axisA\_run18\_rt\_sc2, axisA\_run18\_rt\_sc3, axisA\_run18\_tr\_sc2, axisA\_run18\_tr\_sc3, axisA\_run19\_rt\_sc2, axisA\_run19\_rt\_sc3, axisA\_run19\_tr\_sc2, axisA\_run19\_tr\_sc3, axisA\_run20\_rt\_sc2, axisA\_run20\_rt\_sc3, axisA\_run20\_tr\_sc2, axisA\_run20\_tr\_sc3, axisA\_run21\_rt\_sc2, axisA\_run21\_rt\_sc3, axisA\_run21\_tr\_sc2, axisA\_run21\_tr\_sc3, axisA\_run22\_rt\_sc2, axisA\_run22\_rt\_sc3, axisA\_run22\_tr\_sc2, axisA\_run22\_tr\_sc3, axisA\_run23\_rt\_sc2, axisA\_run23\_rt\_sc3, axisA\_run23\_tr\_sc2, axisA\_run23\_tr\_sc3, axisA\_run24\_rt\_sc2, axisA\_run24\_rt\_sc3, axisA\_run24\_tr\_sc2, axisA\_run24\_tr\_sc3, axisA\_run25\_rt\_sc2, axisA\_run25\_rt\_sc3, axisA\_run25\_tr\_sc2, axisA\_run25\_tr\_sc3, axisA\_run26\_rt\_sc2, axisA\_run26\_rt\_sc3, axisA\_run26\_tr\_sc2, axisA\_run26\_tr\_sc3, axisA\_run27\_rt\_sc2, axisA\_run27\_rt\_sc3, axisA\_run27\_tr\_sc2, axisA\_run27\_tr\_sc3, axisA\_run28\_rt\_sc2, axisA\_run28\_rt\_sc3, axisA\_run28\_tr\_sc2, axisA\_run28\_tr\_sc3, axisA\_run29\_rt\_sc2, axisA\_run29\_rt\_sc3, axisA\_run29\_tr\_sc2, axisA\_run29\_tr\_sc3, axisA\_run30\_rt\_sc2, axisA\_run30\_rt\_sc3, axisA\_run30\_tr\_sc2, axisA\_run30\_tr\_sc3, axisA\_run31\_rt\_sc2, axisA\_run31\_rt\_sc3, axisA\_run31\_tr\_sc2, axisA\_run31\_tr\_sc3\par}

\paragraph{Axis B (352).}
{\tiny\raggedright axisB\_f05\_abstract\_s204, axisB\_f05\_abstract\_s205, axisB\_f05\_abstract\_s206, axisB\_f05\_abstract\_s207, axisB\_f05\_abstract\_s4, axisB\_f05\_abstract\_s5, axisB\_f05\_abstract\_s6, axisB\_f05\_abstract\_s7, axisB\_f05\_household\_s204, axisB\_f05\_household\_s205, axisB\_f05\_household\_s206, axisB\_f05\_household\_s207, axisB\_f05\_household\_s4, axisB\_f05\_household\_s5, axisB\_f05\_household\_s6, axisB\_f05\_household\_s7, axisB\_f05\_person\_s204, axisB\_f05\_person\_s205, axisB\_f05\_person\_s206, axisB\_f05\_person\_s207, axisB\_f05\_person\_s4, axisB\_f05\_person\_s5, axisB\_f05\_person\_s6, axisB\_f05\_person\_s7, axisB\_f05\_vehicle\_s204, axisB\_f05\_vehicle\_s205, axisB\_f05\_vehicle\_s206, axisB\_f05\_vehicle\_s207, axisB\_f05\_vehicle\_s4, axisB\_f05\_vehicle\_s5, axisB\_f05\_vehicle\_s6, axisB\_f05\_vehicle\_s7, axisB\_f10\_abstract\_s204, axisB\_f10\_abstract\_s205, axisB\_f10\_abstract\_s206, axisB\_f10\_abstract\_s207, axisB\_f10\_abstract\_s4, axisB\_f10\_abstract\_s5, axisB\_f10\_abstract\_s6, axisB\_f10\_abstract\_s7, axisB\_f10\_household\_s204, axisB\_f10\_household\_s205, axisB\_f10\_household\_s206, axisB\_f10\_household\_s207, axisB\_f10\_household\_s4, axisB\_f10\_household\_s5, axisB\_f10\_household\_s6, axisB\_f10\_household\_s7, axisB\_f10\_person\_s204, axisB\_f10\_person\_s205, axisB\_f10\_person\_s206, axisB\_f10\_person\_s207, axisB\_f10\_person\_s4, axisB\_f10\_person\_s5, axisB\_f10\_person\_s6, axisB\_f10\_person\_s7, axisB\_f10\_vehicle\_s204, axisB\_f10\_vehicle\_s205, axisB\_f10\_vehicle\_s206, axisB\_f10\_vehicle\_s207, axisB\_f10\_vehicle\_s4, axisB\_f10\_vehicle\_s5, axisB\_f10\_vehicle\_s6, axisB\_f10\_vehicle\_s7, axisB\_f20\_abstract\_s204, axisB\_f20\_abstract\_s205, axisB\_f20\_abstract\_s206, axisB\_f20\_abstract\_s207, axisB\_f20\_abstract\_s4, axisB\_f20\_abstract\_s5, axisB\_f20\_abstract\_s6, axisB\_f20\_abstract\_s7, axisB\_f20\_household\_s204, axisB\_f20\_household\_s205, axisB\_f20\_household\_s206, axisB\_f20\_household\_s207, axisB\_f20\_household\_s4, axisB\_f20\_household\_s5, axisB\_f20\_household\_s6, axisB\_f20\_household\_s7, axisB\_f20\_person\_s204, axisB\_f20\_person\_s205, axisB\_f20\_person\_s206, axisB\_f20\_person\_s207, axisB\_f20\_person\_s4, axisB\_f20\_person\_s5, axisB\_f20\_person\_s6, axisB\_f20\_person\_s7, axisB\_f20\_vehicle\_s204, axisB\_f20\_vehicle\_s205, axisB\_f20\_vehicle\_s206, axisB\_f20\_vehicle\_s207, axisB\_f20\_vehicle\_s4, axisB\_f20\_vehicle\_s5, axisB\_f20\_vehicle\_s6, axisB\_f20\_vehicle\_s7, axisB\_f30\_abstract\_s204, axisB\_f30\_abstract\_s205, axisB\_f30\_abstract\_s206, axisB\_f30\_abstract\_s207, axisB\_f30\_abstract\_s4, axisB\_f30\_abstract\_s5, axisB\_f30\_abstract\_s6, axisB\_f30\_abstract\_s7, axisB\_f30\_household\_s204, axisB\_f30\_household\_s205, axisB\_f30\_household\_s206, axisB\_f30\_household\_s207, axisB\_f30\_household\_s4, axisB\_f30\_household\_s5, axisB\_f30\_household\_s6, axisB\_f30\_household\_s7, axisB\_f30\_person\_s204, axisB\_f30\_person\_s205, axisB\_f30\_person\_s206, axisB\_f30\_person\_s207, axisB\_f30\_person\_s4, axisB\_f30\_person\_s5, axisB\_f30\_person\_s6, axisB\_f30\_person\_s7, axisB\_f30\_vehicle\_s204, axisB\_f30\_vehicle\_s205, axisB\_f30\_vehicle\_s206, axisB\_f30\_vehicle\_s207, axisB\_f30\_vehicle\_s4, axisB\_f30\_vehicle\_s5, axisB\_f30\_vehicle\_s6, axisB\_f30\_vehicle\_s7, axisB\_f40\_abstract\_s204, axisB\_f40\_abstract\_s205, axisB\_f40\_abstract\_s206, axisB\_f40\_abstract\_s207, axisB\_f40\_abstract\_s4, axisB\_f40\_abstract\_s5, axisB\_f40\_abstract\_s6, axisB\_f40\_abstract\_s7, axisB\_f40\_household\_s204, axisB\_f40\_household\_s205, axisB\_f40\_household\_s206, axisB\_f40\_household\_s207, axisB\_f40\_household\_s4, axisB\_f40\_household\_s5, axisB\_f40\_household\_s6, axisB\_f40\_household\_s7, axisB\_f40\_person\_s204, axisB\_f40\_person\_s205, axisB\_f40\_person\_s206, axisB\_f40\_person\_s207, axisB\_f40\_person\_s4, axisB\_f40\_person\_s5, axisB\_f40\_person\_s6, axisB\_f40\_person\_s7, axisB\_f40\_vehicle\_s204, axisB\_f40\_vehicle\_s205, axisB\_f40\_vehicle\_s206, axisB\_f40\_vehicle\_s207, axisB\_f40\_vehicle\_s4, axisB\_f40\_vehicle\_s5, axisB\_f40\_vehicle\_s6, axisB\_f40\_vehicle\_s7, axisB\_f50\_abstract\_s204, axisB\_f50\_abstract\_s205, axisB\_f50\_abstract\_s206, axisB\_f50\_abstract\_s207, axisB\_f50\_abstract\_s4, axisB\_f50\_abstract\_s5, axisB\_f50\_abstract\_s6, axisB\_f50\_abstract\_s7, axisB\_f50\_household\_s204, axisB\_f50\_household\_s205, axisB\_f50\_household\_s206, axisB\_f50\_household\_s207, axisB\_f50\_household\_s4, axisB\_f50\_household\_s5, axisB\_f50\_household\_s6, axisB\_f50\_household\_s7, axisB\_f50\_person\_s204, axisB\_f50\_person\_s205, axisB\_f50\_person\_s206, axisB\_f50\_person\_s207, axisB\_f50\_person\_s4, axisB\_f50\_person\_s5, axisB\_f50\_person\_s6, axisB\_f50\_person\_s7, axisB\_f50\_vehicle\_s204, axisB\_f50\_vehicle\_s205, axisB\_f50\_vehicle\_s206, axisB\_f50\_vehicle\_s207, axisB\_f50\_vehicle\_s4, axisB\_f50\_vehicle\_s5, axisB\_f50\_vehicle\_s6, axisB\_f50\_vehicle\_s7, axisB\_f60\_abstract\_s204, axisB\_f60\_abstract\_s205, axisB\_f60\_abstract\_s206, axisB\_f60\_abstract\_s207, axisB\_f60\_abstract\_s4, axisB\_f60\_abstract\_s5, axisB\_f60\_abstract\_s6, axisB\_f60\_abstract\_s7, axisB\_f60\_household\_s204, axisB\_f60\_household\_s205, axisB\_f60\_household\_s206, axisB\_f60\_household\_s207, axisB\_f60\_household\_s4, axisB\_f60\_household\_s5, axisB\_f60\_household\_s6, axisB\_f60\_household\_s7, axisB\_f60\_person\_s204, axisB\_f60\_person\_s205, axisB\_f60\_person\_s206, axisB\_f60\_person\_s207, axisB\_f60\_person\_s4, axisB\_f60\_person\_s5, axisB\_f60\_person\_s6, axisB\_f60\_person\_s7, axisB\_f60\_vehicle\_s204, axisB\_f60\_vehicle\_s205, axisB\_f60\_vehicle\_s206, axisB\_f60\_vehicle\_s207, axisB\_f60\_vehicle\_s4, axisB\_f60\_vehicle\_s5, axisB\_f60\_vehicle\_s6, axisB\_f60\_vehicle\_s7, axisB\_f70\_abstract\_s204, axisB\_f70\_abstract\_s205, axisB\_f70\_abstract\_s206, axisB\_f70\_abstract\_s207, axisB\_f70\_abstract\_s4, axisB\_f70\_abstract\_s5, axisB\_f70\_abstract\_s6, axisB\_f70\_abstract\_s7, axisB\_f70\_household\_s204, axisB\_f70\_household\_s205, axisB\_f70\_household\_s206, axisB\_f70\_household\_s207, axisB\_f70\_household\_s4, axisB\_f70\_household\_s5, axisB\_f70\_household\_s6, axisB\_f70\_household\_s7, axisB\_f70\_person\_s204, axisB\_f70\_person\_s205, axisB\_f70\_person\_s206, axisB\_f70\_person\_s207, axisB\_f70\_person\_s4, axisB\_f70\_person\_s5, axisB\_f70\_person\_s6, axisB\_f70\_person\_s7, axisB\_f70\_vehicle\_s204, axisB\_f70\_vehicle\_s205, axisB\_f70\_vehicle\_s206, axisB\_f70\_vehicle\_s207, axisB\_f70\_vehicle\_s4, axisB\_f70\_vehicle\_s5, axisB\_f70\_vehicle\_s6, axisB\_f70\_vehicle\_s7, axisB\_f80\_abstract\_s204, axisB\_f80\_abstract\_s205, axisB\_f80\_abstract\_s206, axisB\_f80\_abstract\_s207, axisB\_f80\_abstract\_s4, axisB\_f80\_abstract\_s5, axisB\_f80\_abstract\_s6, axisB\_f80\_abstract\_s7, axisB\_f80\_household\_s204, axisB\_f80\_household\_s205, axisB\_f80\_household\_s206, axisB\_f80\_household\_s207, axisB\_f80\_household\_s4, axisB\_f80\_household\_s5, axisB\_f80\_household\_s6, axisB\_f80\_household\_s7, axisB\_f80\_person\_s204, axisB\_f80\_person\_s205, axisB\_f80\_person\_s206, axisB\_f80\_person\_s207, axisB\_f80\_person\_s4, axisB\_f80\_person\_s5, axisB\_f80\_person\_s6, axisB\_f80\_person\_s7, axisB\_f80\_vehicle\_s204, axisB\_f80\_vehicle\_s205, axisB\_f80\_vehicle\_s206, axisB\_f80\_vehicle\_s207, axisB\_f80\_vehicle\_s4, axisB\_f80\_vehicle\_s5, axisB\_f80\_vehicle\_s6, axisB\_f80\_vehicle\_s7, axisB\_f90\_abstract\_s204, axisB\_f90\_abstract\_s205, axisB\_f90\_abstract\_s206, axisB\_f90\_abstract\_s207, axisB\_f90\_abstract\_s4, axisB\_f90\_abstract\_s5, axisB\_f90\_abstract\_s6, axisB\_f90\_abstract\_s7, axisB\_f90\_household\_s204, axisB\_f90\_household\_s205, axisB\_f90\_household\_s206, axisB\_f90\_household\_s207, axisB\_f90\_household\_s4, axisB\_f90\_household\_s5, axisB\_f90\_household\_s6, axisB\_f90\_household\_s7, axisB\_f90\_person\_s204, axisB\_f90\_person\_s205, axisB\_f90\_person\_s206, axisB\_f90\_person\_s207, axisB\_f90\_person\_s4, axisB\_f90\_person\_s5, axisB\_f90\_person\_s6, axisB\_f90\_person\_s7, axisB\_f90\_vehicle\_s204, axisB\_f90\_vehicle\_s205, axisB\_f90\_vehicle\_s206, axisB\_f90\_vehicle\_s207, axisB\_f90\_vehicle\_s4, axisB\_f90\_vehicle\_s5, axisB\_f90\_vehicle\_s6, axisB\_f90\_vehicle\_s7, axisB\_f95\_abstract\_s204, axisB\_f95\_abstract\_s205, axisB\_f95\_abstract\_s206, axisB\_f95\_abstract\_s207, axisB\_f95\_abstract\_s4, axisB\_f95\_abstract\_s5, axisB\_f95\_abstract\_s6, axisB\_f95\_abstract\_s7, axisB\_f95\_household\_s204, axisB\_f95\_household\_s205, axisB\_f95\_household\_s206, axisB\_f95\_household\_s207, axisB\_f95\_household\_s4, axisB\_f95\_household\_s5, axisB\_f95\_household\_s6, axisB\_f95\_household\_s7, axisB\_f95\_person\_s204, axisB\_f95\_person\_s205, axisB\_f95\_person\_s206, axisB\_f95\_person\_s207, axisB\_f95\_person\_s4, axisB\_f95\_person\_s5, axisB\_f95\_person\_s6, axisB\_f95\_person\_s7, axisB\_f95\_vehicle\_s204, axisB\_f95\_vehicle\_s205, axisB\_f95\_vehicle\_s206, axisB\_f95\_vehicle\_s207, axisB\_f95\_vehicle\_s4, axisB\_f95\_vehicle\_s5, axisB\_f95\_vehicle\_s6, axisB\_f95\_vehicle\_s7\par}

\paragraph{Axis C (80).}
{\tiny\raggedright axisC\_g10\_boat\_s4, axisC\_g10\_boat\_s5, axisC\_g10\_boat\_s6, axisC\_g10\_boat\_s7, axisC\_g10\_elevator\_s4, axisC\_g10\_elevator\_s5, axisC\_g10\_elevator\_s6, axisC\_g10\_elevator\_s7, axisC\_g10\_escalator\_s4, axisC\_g10\_escalator\_s5, axisC\_g10\_escalator\_s6, axisC\_g10\_escalator\_s7, axisC\_g10\_train\_s4, axisC\_g10\_train\_s5, axisC\_g10\_train\_s6, axisC\_g10\_train\_s7, axisC\_g10\_treadmill\_s4, axisC\_g10\_treadmill\_s5, axisC\_g10\_treadmill\_s6, axisC\_g10\_treadmill\_s7, axisC\_g30\_boat\_s4, axisC\_g30\_boat\_s5, axisC\_g30\_boat\_s6, axisC\_g30\_boat\_s7, axisC\_g30\_elevator\_s4, axisC\_g30\_elevator\_s5, axisC\_g30\_elevator\_s6, axisC\_g30\_elevator\_s7, axisC\_g30\_escalator\_s4, axisC\_g30\_escalator\_s5, axisC\_g30\_escalator\_s6, axisC\_g30\_escalator\_s7, axisC\_g30\_train\_s4, axisC\_g30\_train\_s5, axisC\_g30\_train\_s6, axisC\_g30\_train\_s7, axisC\_g30\_treadmill\_s4, axisC\_g30\_treadmill\_s5, axisC\_g30\_treadmill\_s6, axisC\_g30\_treadmill\_s7, axisC\_g50\_boat\_s4, axisC\_g50\_boat\_s5, axisC\_g50\_boat\_s6, axisC\_g50\_boat\_s7, axisC\_g50\_elevator\_s4, axisC\_g50\_elevator\_s5, axisC\_g50\_elevator\_s6, axisC\_g50\_elevator\_s7, axisC\_g50\_escalator\_s4, axisC\_g50\_escalator\_s5, axisC\_g50\_escalator\_s6, axisC\_g50\_escalator\_s7, axisC\_g50\_train\_s4, axisC\_g50\_train\_s5, axisC\_g50\_train\_s6, axisC\_g50\_train\_s7, axisC\_g50\_treadmill\_s4, axisC\_g50\_treadmill\_s5, axisC\_g50\_treadmill\_s6, axisC\_g50\_treadmill\_s7, axisC\_g70\_boat\_s4, axisC\_g70\_boat\_s5, axisC\_g70\_boat\_s6, axisC\_g70\_boat\_s7, axisC\_g70\_elevator\_s4, axisC\_g70\_elevator\_s5, axisC\_g70\_elevator\_s6, axisC\_g70\_elevator\_s7, axisC\_g70\_escalator\_s4, axisC\_g70\_escalator\_s5, axisC\_g70\_escalator\_s6, axisC\_g70\_escalator\_s7, axisC\_g70\_train\_s4, axisC\_g70\_train\_s5, axisC\_g70\_train\_s6, axisC\_g70\_train\_s7, axisC\_g70\_treadmill\_s4, axisC\_g70\_treadmill\_s5, axisC\_g70\_treadmill\_s6, axisC\_g70\_treadmill\_s7\par}

\paragraph{Axis D (240).}
{\tiny\raggedright axisD\_t00\_sc2\_mover, axisD\_t00\_sc2\_static, axisD\_t00\_sc3\_mover, axisD\_t00\_sc3\_static, axisD\_t01\_sc2\_mover, axisD\_t01\_sc2\_static, axisD\_t01\_sc3\_mover, axisD\_t01\_sc3\_static, axisD\_t02\_sc2\_mover, axisD\_t02\_sc2\_static, axisD\_t02\_sc3\_mover, axisD\_t02\_sc3\_static, axisD\_t03\_sc2\_mover, axisD\_t03\_sc2\_static, axisD\_t03\_sc3\_mover, axisD\_t03\_sc3\_static, axisD\_t04\_sc2\_mover, axisD\_t04\_sc2\_static, axisD\_t04\_sc3\_mover, axisD\_t04\_sc3\_static, axisD\_t05\_sc2\_mover, axisD\_t05\_sc2\_static, axisD\_t05\_sc3\_mover, axisD\_t05\_sc3\_static, axisD\_t06\_sc2\_mover, axisD\_t06\_sc2\_static, axisD\_t06\_sc3\_mover, axisD\_t06\_sc3\_static, axisD\_t07\_sc2\_mover, axisD\_t07\_sc2\_static, axisD\_t07\_sc3\_mover, axisD\_t07\_sc3\_static, axisD\_t08\_sc2\_mover, axisD\_t08\_sc2\_static, axisD\_t08\_sc3\_mover, axisD\_t08\_sc3\_static, axisD\_t09\_sc2\_mover, axisD\_t09\_sc2\_static, axisD\_t09\_sc3\_mover, axisD\_t09\_sc3\_static, axisD\_t10\_sc2\_mover, axisD\_t10\_sc2\_static, axisD\_t10\_sc3\_mover, axisD\_t10\_sc3\_static, axisD\_t11\_sc2\_mover, axisD\_t11\_sc2\_static, axisD\_t11\_sc3\_mover, axisD\_t11\_sc3\_static, axisD\_t12\_sc2\_mover, axisD\_t12\_sc2\_static, axisD\_t12\_sc3\_mover, axisD\_t12\_sc3\_static, axisD\_t13\_sc2\_mover, axisD\_t13\_sc2\_static, axisD\_t13\_sc3\_mover, axisD\_t13\_sc3\_static, axisD\_t14\_sc2\_mover, axisD\_t14\_sc2\_static, axisD\_t14\_sc3\_mover, axisD\_t14\_sc3\_static, axisD\_t15\_sc2\_mover, axisD\_t15\_sc2\_static, axisD\_t15\_sc3\_mover, axisD\_t15\_sc3\_static, axisD\_t16\_sc2\_mover, axisD\_t16\_sc2\_static, axisD\_t16\_sc3\_mover, axisD\_t16\_sc3\_static, axisD\_t17\_sc2\_mover, axisD\_t17\_sc2\_static, axisD\_t17\_sc3\_mover, axisD\_t17\_sc3\_static, axisD\_t18\_sc2\_mover, axisD\_t18\_sc2\_static, axisD\_t18\_sc3\_mover, axisD\_t18\_sc3\_static, axisD\_t19\_sc2\_mover, axisD\_t19\_sc2\_static, axisD\_t19\_sc3\_mover, axisD\_t19\_sc3\_static, axisD\_t20\_sc2\_mover, axisD\_t20\_sc2\_static, axisD\_t20\_sc3\_mover, axisD\_t20\_sc3\_static, axisD\_t21\_sc2\_mover, axisD\_t21\_sc2\_static, axisD\_t21\_sc3\_mover, axisD\_t21\_sc3\_static, axisD\_t22\_sc2\_mover, axisD\_t22\_sc2\_static, axisD\_t22\_sc3\_mover, axisD\_t22\_sc3\_static, axisD\_t23\_sc2\_mover, axisD\_t23\_sc2\_static, axisD\_t23\_sc3\_mover, axisD\_t23\_sc3\_static, axisD\_t24\_sc2\_mover, axisD\_t24\_sc2\_static, axisD\_t24\_sc3\_mover, axisD\_t24\_sc3\_static, axisD\_t25\_sc2\_mover, axisD\_t25\_sc2\_static, axisD\_t25\_sc3\_mover, axisD\_t25\_sc3\_static, axisD\_t26\_sc2\_mover, axisD\_t26\_sc2\_static, axisD\_t26\_sc3\_mover, axisD\_t26\_sc3\_static, axisD\_t27\_sc2\_mover, axisD\_t27\_sc2\_static, axisD\_t27\_sc3\_mover, axisD\_t27\_sc3\_static, axisD\_t28\_sc2\_mover, axisD\_t28\_sc2\_static, axisD\_t28\_sc3\_mover, axisD\_t28\_sc3\_static, axisD\_t29\_sc2\_mover, axisD\_t29\_sc2\_static, axisD\_t29\_sc3\_mover, axisD\_t29\_sc3\_static, axisD\_t30\_sc2\_mover, axisD\_t30\_sc2\_static, axisD\_t30\_sc3\_mover, axisD\_t30\_sc3\_static, axisD\_t31\_sc2\_mover, axisD\_t31\_sc2\_static, axisD\_t31\_sc3\_mover, axisD\_t31\_sc3\_static, axisD\_t32\_sc2\_mover, axisD\_t32\_sc2\_static, axisD\_t32\_sc3\_mover, axisD\_t32\_sc3\_static, axisD\_t33\_sc2\_mover, axisD\_t33\_sc2\_static, axisD\_t33\_sc3\_mover, axisD\_t33\_sc3\_static, axisD\_t34\_sc2\_mover, axisD\_t34\_sc2\_static, axisD\_t34\_sc3\_mover, axisD\_t34\_sc3\_static, axisD\_t35\_sc2\_mover, axisD\_t35\_sc2\_static, axisD\_t35\_sc3\_mover, axisD\_t35\_sc3\_static, axisD\_t36\_sc2\_mover, axisD\_t36\_sc2\_static, axisD\_t36\_sc3\_mover, axisD\_t36\_sc3\_static, axisD\_t37\_sc2\_mover, axisD\_t37\_sc2\_static, axisD\_t37\_sc3\_mover, axisD\_t37\_sc3\_static, axisD\_t38\_sc2\_mover, axisD\_t38\_sc2\_static, axisD\_t38\_sc3\_mover, axisD\_t38\_sc3\_static, axisD\_t39\_sc2\_mover, axisD\_t39\_sc2\_static, axisD\_t39\_sc3\_mover, axisD\_t39\_sc3\_static, axisD\_t40\_sc2\_mover, axisD\_t40\_sc2\_static, axisD\_t40\_sc3\_mover, axisD\_t40\_sc3\_static, axisD\_t41\_sc2\_mover, axisD\_t41\_sc2\_static, axisD\_t41\_sc3\_mover, axisD\_t41\_sc3\_static, axisD\_t42\_sc2\_mover, axisD\_t42\_sc2\_static, axisD\_t42\_sc3\_mover, axisD\_t42\_sc3\_static, axisD\_t43\_sc2\_mover, axisD\_t43\_sc2\_static, axisD\_t43\_sc3\_mover, axisD\_t43\_sc3\_static, axisD\_t44\_sc2\_mover, axisD\_t44\_sc2\_static, axisD\_t44\_sc3\_mover, axisD\_t44\_sc3\_static, axisD\_t45\_sc2\_mover, axisD\_t45\_sc2\_static, axisD\_t45\_sc3\_mover, axisD\_t45\_sc3\_static, axisD\_t46\_sc2\_mover, axisD\_t46\_sc2\_static, axisD\_t46\_sc3\_mover, axisD\_t46\_sc3\_static, axisD\_t47\_sc2\_mover, axisD\_t47\_sc2\_static, axisD\_t47\_sc3\_mover, axisD\_t47\_sc3\_static, axisD\_t48\_sc2\_mover, axisD\_t48\_sc2\_static, axisD\_t48\_sc3\_mover, axisD\_t48\_sc3\_static, axisD\_t49\_sc2\_mover, axisD\_t49\_sc2\_static, axisD\_t49\_sc3\_mover, axisD\_t49\_sc3\_static, axisD\_t50\_sc2\_mover, axisD\_t50\_sc2\_static, axisD\_t50\_sc3\_mover, axisD\_t50\_sc3\_static, axisD\_t51\_sc2\_mover, axisD\_t51\_sc2\_static, axisD\_t51\_sc3\_mover, axisD\_t51\_sc3\_static, axisD\_t52\_sc2\_mover, axisD\_t52\_sc2\_static, axisD\_t52\_sc3\_mover, axisD\_t52\_sc3\_static, axisD\_t53\_sc2\_mover, axisD\_t53\_sc2\_static, axisD\_t53\_sc3\_mover, axisD\_t53\_sc3\_static, axisD\_t54\_sc2\_mover, axisD\_t54\_sc2\_static, axisD\_t54\_sc3\_mover, axisD\_t54\_sc3\_static, axisD\_t55\_sc2\_mover, axisD\_t55\_sc2\_static, axisD\_t55\_sc3\_mover, axisD\_t55\_sc3\_static, axisD\_t56\_sc2\_mover, axisD\_t56\_sc2\_static, axisD\_t56\_sc3\_mover, axisD\_t56\_sc3\_static, axisD\_t57\_sc2\_mover, axisD\_t57\_sc2\_static, axisD\_t57\_sc3\_mover, axisD\_t57\_sc3\_static, axisD\_t58\_sc2\_mover, axisD\_t58\_sc2\_static, axisD\_t58\_sc3\_mover, axisD\_t58\_sc3\_static, axisD\_t59\_sc2\_mover, axisD\_t59\_sc2\_static, axisD\_t59\_sc3\_mover, axisD\_t59\_sc3\_static\par}

\section{Qualitative Results and Video}
\label{app:qual}
\begin{figure*}[t]
  \centering
  \includegraphics[width=0.84\textwidth]{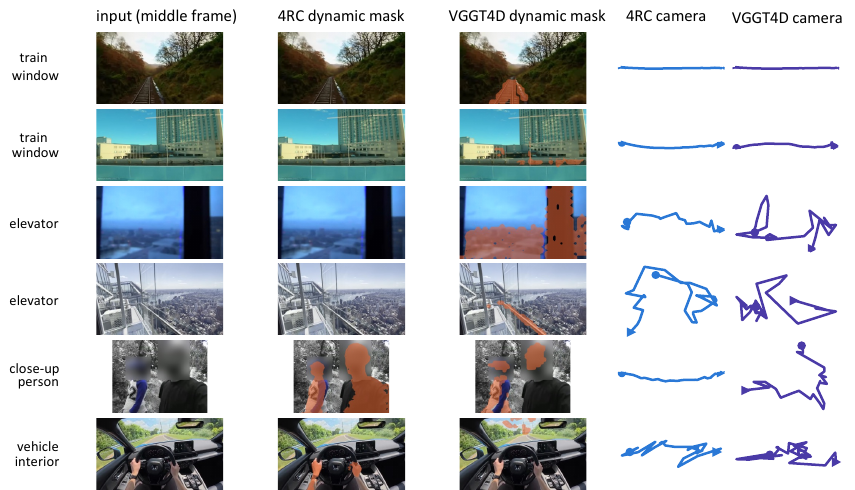}
  \caption{Real clips, two models. Per row: the middle frame, the dynamic masks of 4RC (which keeps the physical frame on Axis~C 78\% of the time) and of VGGT4D (16\%), and their camera trajectories as top-down traces. On the train and elevator clips VGGT4D marks the passing world as dynamic and reconstructs a wandering camera; on the close-up person (faces blurred) both masks find the mover and both cameras differ.}
  \label{fig:qual}
\end{figure*}
\Cref{fig:qual} uses six web clips (two train windows, two glass elevators, one person close to the camera, one car interior), 30 frames each, referenced by public video identifier and timestamp in the released index; 4RC and VGGT4D were run as released and their dynamic masks and camera trajectories are drawn without post-processing. The supplementary video (95\,s, 1080p, captions and no narration) shows one Axis~B scene across all eleven mover fractions, two Axis~C sequences (train window and glass elevator at $g=0.5$) and three real clips, each with the dynamic masks and camera traces of MonST3R, 4RC and VGGT4D side by side. The labelled real-world panel of Amendment~6 was reduced to four situations by Amendment~9.4 and, with four of the forty labels in place at submission, is dropped as that amendment prescribes; the in-the-wild evidence of this paper is the real driving and egocentric sets of \cref{app:realvideo} and the existing-benchmark transfer.

\section{Practical Guidance}
\label{app:guidance}
\Cref{tab:guidance} reads the leaderboard of the main text as advice, one row per ambiguous situation with the models to prefer and to avoid, the number that decides it, and whether the data recipe of the main text repairs it, with the source of each statement.

\begin{table*}[t]
\centering
\scriptsize
\caption{Practical guidance per ambiguous situation, read three ways (Amendment~22, A22.10): by mask quality, by camera accuracy and by the joint reconstruction, with the models to prefer and to avoid, the number behind each, and whether the data recipe of Section 4.4 of the main text repairs it (with the source). A model that leads on the mask's frame is not thereby the model for an accurate reconstruction.}
\label{tab:guidance}
\begin{tabular}{p{1.6cm}p{1.4cm}p{3.9cm}p{2.3cm}p{2.4cm}p{3.0cm}}
\toprule
Situation & Criterion & Prefer & Avoid & Number & Repairable by the recipe \\
\midrule
Moving platform, physical frame wanted & mask quality & PAGE-4D (0.75 synthetic, 0.94 real), V-DPM (0.74 synthetic, 0.95 real); 4RC leads the synthetic axis (0.78) and reverses on real video (0.23) & MoRe (0.21, 0.31), VGGT4D (0.16, 0.33) & physical-frame rate from the dynamic mask: Axis C sequences (synthetic) and ADT hands clips (real) & yes with a motion head: MoRe 0.31 to 0.94 (A18), VGGT4D head 0.33 to 0.88 with camera error +4.8\% (A19.3); no for fine-tunes that move the geometry (MonST3R, CUT3R) \\
 & camera accuracy & Depth Anything 3 (0.10), MoRe (0.11), $\pi^3$ (0.11) & Easi3R (0.27), MegaSaM (0.24) & median camera ATE/path, Axis C sequences (synthetic) & no: the camera's attribution does not track the mask's (Amendment~21) \\
 & joint reconstruction & $\pi^3$ (0.30, 0.27), Point4D (0.36, 0.36), V-DPM (0.39, 0.27) & CUT3R (1.24, 0.47), MonST3R (0.86, 0.54) & median error of the camera-attached foreground's world trajectory (A17.5): Axis C (synthetic) and ADT hands clips (real) & partly: MoRe 0.44 to 0.30 on the Axis C test split and 0.43 to 0.42 on the hands clips after 400 sequences (A17.5) \\
\midrule
Dynamic mask with a small mover & mask quality & TraceAnything (0.93), MonST3R (0.88), V-DPM (0.80) & CUT3R (0.25), Depth Anything 3 (0.32) (derived masks) & dynamic-mask IoU at $f=0.10$, Axis B (synthetic) & not targeted (the recipe addresses frame choice) \\
 & camera accuracy & MegaSaM (0.017), VGGT4D (0.035), Depth Anything 3 (0.037) & TraceAnything (0.234), CUT3R (0.098) & camera ATE/path at $f=0.10$, Axis B (synthetic) & not targeted \\
 & joint reconstruction & none leads on both & CUT3R & models in the top three of both rows above & not targeted \\
\midrule
Static scene under fast rotation & mask quality & UniQuery4R (0.001), MoRe (0.009), MegaSaM (0.022) & Point4D (0.46), Depth Anything 3 (0.40) & dynamic-mask false-positive rate, worst rotation bin, Axis D (synthetic) & not targeted (the recipe addresses frame choice) \\
 & camera accuracy & UniQuery4R (0.149), PAGE-4D (0.152), 4RC (0.163) & Easi3R (0.243), MonST3R (0.226) & camera ATE/path in the highest rotation bin, static scenes, Axis D (synthetic) & not targeted \\
 & joint reconstruction & UniQuery4R & none leads on both & models in the top three of both rows above & not targeted \\
\midrule
Camera under a large mover & mask quality & 4RC (none up to 0.95), TraceAnything (0.9), V-DPM (0.9) & VGGT4D (0.3), MonST3R (0.5) & mask degradation point $f_\mathrm{mask}$, Axis B (synthetic), models with IoU at least 0.6 at $f=0.10$; later is better & the recipe moves the mask's frame, not its degradation point \\
 & camera accuracy & Depth Anything 3 ($f_{>0.10}$ = 0.9), MegaSaM ($f_{>0.10}$ = 0.9), 4RC ($f_{>0.10}$ = 0.6) & CUT3R (0.2), Any4D (0.3) & camera ATE/path: first $f$ above 0.10, Axis B (synthetic) & no: training leaves the degradation threshold (Gates 5b, 6b) \\
 & joint reconstruction & 4RC & none leads on both & models in the top three of both rows above & no \\
\bottomrule
\end{tabular}
\end{table*}

\section{Attribution Across Outputs and Controls}
\label{app:a20}
Amendment~20 (in response to an external review) fixed seven analyses before they ran, and Amendment~21 added the decomposition of the hands camera-depth attribution (A21.1); every reading is in \texttt{results/phase9/} and in the Gate~9 section of the results log, and no pre-registered verdict is re-decided. This section carries what the main text summarises.

\begin{figure*}[t]
  \centering
  \includegraphics[width=\textwidth]{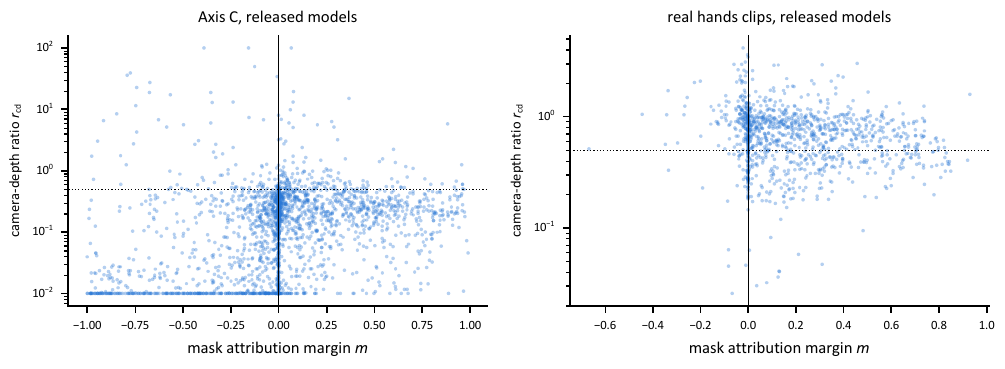}
  \caption{Attribution across outputs (A20.1). Every released-model run on the Axis~C sequences (left) and the real hands clips (right): the mask attribution margin $m$ against the camera-depth ratio $r_\mathrm{cd}$ (log scale); the lines mark $m=0$ and $r_\mathrm{cd}=0.5$, the two decision rules.}
  \label{fig:agreement}
\end{figure*}

\paragraph{Attribution Across Outputs (A20.1).} Table~3 of the main text (read with the camera label of Amendment~23), \cref{tab:agreement_full} (every checkpoint, both labels) and \cref{fig:agreement} read every cached output twice: the mask attribution margin $m$, the mean dynamic-mask value on the camera-attached foreground minus that on the background (physical when $m>0$; ties, $|m|<0.02$, are reported as ties), and the camera-depth ratio $r_\mathrm{cd}$ (the geometric attribution ratio $r_\mathrm{geo}$ of Amendment~20, renamed and read as a diagnostic under A22.3), the predicted camera path length over the model's median predicted depth of the foreground pixels, against the same quantity from the ground truth (physical when $r_\mathrm{cd}>0.5$). Over the 14 released models with poses and depth, on the 160 Axis~C sequences the mask-physical rate runs from 0.16 to 0.78, the camera-depth-physical rate from 0.01 to 0.37 and the agreement from 0.28 to 0.82 (8 ties for MonST3R, 515 of 2232 in all); on the 81 hands clips the three run from 0.23 to 0.95, 0.44 to 0.88 and 0.32 to 0.79. The camera-depth ratio reads egocentric on most Axis~C sequences for every model: the predicted path is short against the predicted foreground depth, whatever the mask says, so the two attributions disagree wherever the mask reads physical. The recipe on MoRe moves the mask-physical rate on the Axis~C test split from 0.19 to 1.00 and the camera-depth-physical rate from 0.00 to 0.54 (both move on the rendered distribution), and on the hands clips the mask from 0.31 to 0.91 with the camera-depth ratio from 0.56 to 0.60 (the mask moves and the geometric cell stays on this checkpoint; the main text reports every checkpoint, \cref{tab:agreement_full}, and Amendment~21 reads the movement); the VGGT4D head moves the mask from 0.14 to 1.00 on Axis~C and from 0.33 to 0.88 on the hands clips with the camera-depth ratio at 0.06 and 0.81 on both, unchanged. The downstream foreground-trajectory error (A17.5) by agreement cell, pooled over the released models: on Axis~C 0.46 where both read physical ($n=156$), 0.45 for a physical mask with an egocentric ratio ($n=1001$), 0.48 for the reverse ($n=93$) and 0.58 where both read egocentric ($n=982$); on the hands clips 0.52, 0.29, 0.70 and 0.35 in the same order ($n=591$, 257, 190, 76). The coherent cell does not have the lowest error on either set: the mask decides the trajectory's frame and the camera-depth ratio adds no separation beyond it.

\paragraph{Factors of the Camera-Depth Ratio (A21.1).} The camera-depth ratio on the 81 hands clips is the quotient of two scale-free factors, per clip $r_\mathrm{cd} = (L/B)/(Z/B)$ against the same from the ground truth, with $L$ the predicted camera path length, $B$ the median predicted depth of the static background and $Z$ the median predicted depth of the camera-attached foreground: the camera factor $L/B$ over the foreground factor $Z/B$. The identity is applied per clip before any median is taken (the script computes both factors per clip and takes the median over clips of each factor's log ratio against the released checkpoint on the same clip; verified in a later audit, no recomputation needed), each factor read here as the median over clips of its log ratio against the released checkpoint on the same clip, with a 2000-resample bootstrap interval (seed 0), from the cached outputs of the released MoRe and its five fine-tunes (\texttt{results/phase9/attribution/rgeo\_factors.json}; no model run). 5 rendered: camera factor $-0.109$ [$-0.150$, $-0.046$] ($-10.3\%$), foreground factor $-0.660$ [$-0.768$, $-0.620$] ($-48.3\%$); 25 rendered: camera factor $-0.033$ [$-0.059$, $-0.017$] ($-3.3\%$), foreground factor $+0.007$ [$-0.134$, $+0.110$] ($+0.7\%$); 400 rendered: camera factor $+0.005$ [$-0.011$, $+0.026$] ($+0.6\%$), foreground factor $-0.149$ [$-0.218$, $-0.054$] ($-13.8\%$); 400 control: camera factor $-0.093$ [$-0.130$, $-0.009$] ($-8.9\%$), foreground factor $-0.683$ [$-0.716$, $-0.664$] ($-49.5\%$); real labels: camera factor $+0.173$ [$+0.137$, $+0.197$] ($+18.9\%$), foreground factor $-0.258$ [$-0.290$, $-0.232$] ($-22.7\%$) (81 clips each). The foreground factor's absolute change exceeds the camera factor's on 4 of the 5 fine-tunes, so by the rule fixed in advance the paper says that the ratio moves with the predicted depth of the hands against the room and not with the camera's travel; the largest absolute change of the camera factor is 18.9\% (the real-label fine-tune), rounded up to 19\% in the main text.

% generated by scripts/phase10_camlabel.py (Amendment 23, A23.1); traced in paper/NUMBERS.md (F16); do not edit
\begin{table*}[t]
  \centering
  \scriptsize
  \setlength{\tabcolsep}{3pt}
  \caption{Attribution across outputs, every checkpoint (Amendments~20 and 23). Per released model and recipe checkpoint on the Axis~C sequences (all 160 for the released models, the 80 test sequences for the checkpoints) and the real hands clips: the share whose dynamic mask reads the physical frame (mask), whose camera label reads it (camera, Amendment~23: predicted travel over predicted background depth, above half the ground truth's), whose foreground-normalised depth diagnostic reads it ($r_\mathrm{cd} > 0.5$, Amendment~20), the share on which the mask and the camera label agree, and the median downstream foreground-trajectory error (A17.5, error over path). The camera label is undefined where no background pixel has a finite depth (four Axis~C treadmill sequences at $g = 0.70$); its rate and the agreement are over the labelled sequences, the mask and $r_\mathrm{cd}$ over the A20.1 denominator.}
  \label{tab:agreement_full}
  \resizebox{\textwidth}{!}{\begin{tabular}{@{}lrrrrr|rrrrr@{}}
    \toprule
    & \multicolumn{5}{c|}{Axis C (synthetic)} & \multicolumn{5}{c}{real hands clips} \\
    model & mask & camera & $r_\mathrm{cd}$ & agree & downstream & mask & camera & $r_\mathrm{cd}$ & agree & downstream \\
    \midrule
    MonST3R & 0.62 & 0.48 & 0.14 & 0.64 & 0.86 & 0.93 & 1.00 & 0.57 & 0.93 & 0.54 \\
    CUT3R & 0.42 & 0.42 & 0.12 & 0.63 & 1.24 & 0.85 & 1.00 & 0.63 & 0.85 & 0.47 \\
    Easi3R & 0.39 & 0.66 & 0.35 & 0.38 & 0.54 & 0.77 & 1.00 & 0.44 & 0.77 & 0.55 \\
    VGGT4D & 0.16 & 0.70 & 0.06 & 0.37 & 0.47 & 0.33 & 1.00 & 0.81 & 0.33 & 0.52 \\
    $\pi^3$ & 0.65 & 0.85 & 0.35 & 0.67 & 0.30 & 0.85 & 1.00 & 0.88 & 0.85 & 0.27 \\
    Any4D & 0.70 & 0.60 & 0.12 & 0.64 & 0.57 & 0.86 & 1.00 & 0.81 & 0.86 & 0.49 \\
    MegaSaM & 0.37 & 0.37 & 0.03 & 0.88 & 0.67 & 0.89 & 0.98 & 0.64 & 0.89 & 0.41 \\
    Depth Anything 3 & 0.57 & 0.65 & 0.05 & 0.72 & 0.46 & 0.90 & 1.00 & 0.85 & 0.90 & 0.29 \\
    4RC & 0.78 & 0.55 & 0.09 & 0.65 & 0.50 & 0.23 & 1.00 & 0.60 & 0.23 & 0.53 \\
    V-DPM & 0.74 & 0.71 & 0.02 & 0.83 & 0.39 & 0.95 & 1.00 & 0.81 & 0.95 & 0.27 \\
    MoRe & 0.21 & 0.78 & 0.01 & 0.43 & 0.41 & 0.31 & 1.00 & 0.56 & 0.31 & 0.43 \\
    UniQuery4R & 0.41 & 0.43 & 0.00 & 0.67 & 0.49 & 0.94 & 1.00 & 0.81 & 0.94 & 0.20 \\
    Point4D & 0.47 & 0.60 & 0.10 & 0.77 & 0.36 & 0.78 & 1.00 & 0.79 & 0.78 & 0.36 \\
    PAGE-4D & 0.75 & 0.82 & 0.06 & 0.67 & 0.46 & 0.94 & 1.00 & 0.67 & 0.94 & 0.51 \\
    \midrule
    MoRe, released & 0.19 & 0.75 & 0.00 & 0.43 & 0.44 & 0.31 & 1.00 & 0.56 & 0.31 & 0.43 \\
    MoRe, 5 rendered & 1.00 & 0.70 & 0.54 & 0.70 & 0.29 & 0.94 & 1.00 & 0.86 & 0.94 & 0.36 \\
    MoRe, 25 rendered & 1.00 & 0.71 & 0.54 & 0.71 & 0.32 & 0.91 & 1.00 & 0.54 & 0.91 & 0.57 \\
    MoRe, 400 rendered & 1.00 & 0.75 & 0.53 & 0.75 & 0.30 & 0.91 & 1.00 & 0.60 & 0.91 & 0.42 \\
    MoRe, 400 control & 0.85 & 0.74 & 0.17 & 0.87 & 0.46 & 0.94 & 1.00 & 0.90 & 0.94 & 0.38 \\
    MoRe, real labels & 0.57 & 0.78 & 0.11 & 0.72 & 0.46 & 0.96 & 1.00 & 0.80 & 0.96 & 0.29 \\
    VGGT4D, released & 0.14 & 0.66 & 0.05 & 0.36 & 0.48 & 0.33 & 1.00 & 0.81 & 0.33 & 0.52 \\
    VGGT4D head, 25 rendered & 1.00 & 0.61 & 0.04 & 0.61 & 0.49 & 0.88 & 1.00 & 0.81 & 0.88 & 0.51 \\
    VGGT4D head, 25 control & 0.96 & 0.64 & 0.04 & 0.68 & 0.49 & 0.85 & 1.00 & 0.83 & 0.85 & 0.53 \\
    \bottomrule
  \end{tabular}}
\end{table*}

\begin{figure*}[t]
  \centering
  \includegraphics[width=\textwidth]{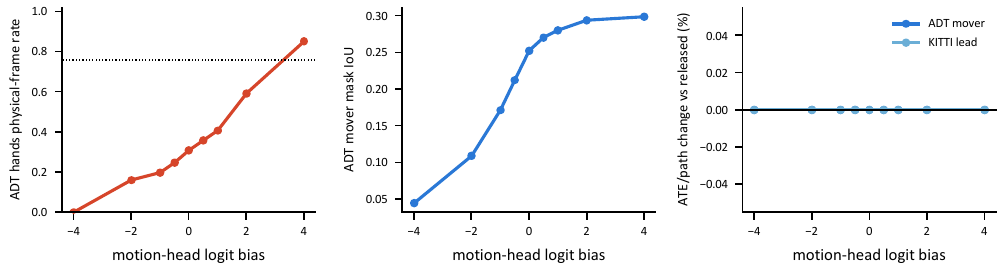}
  \caption{Calibration control (A20.2). The released MoRe's hands physical-frame rate, mover mask IoU and the two camera errors against the motion-head threshold (logit bias), paired bootstrap intervals against the released threshold; the dotted line is the 45-point bar.}
  \label{fig:calibration}
\end{figure*}

%% A20.2 begin
\paragraph{Calibration Control for the Recipe (A20.2).} The released MoRe was run once on the 227 real clips (81 hands, 105 mover, 41 leading-vehicle) with its motion-head probability map kept, and the mask threshold swept offline over nine values, the released one (0.5) and logit biases of $\pm0.5$, $\pm1$, $\pm2$ and $\pm4$ (threshold $1/(1+e^{b})$), the same evaluation code as the recipe runs and a paired bootstrap over clips against the released threshold (\cref{fig:calibration}). The hands physical-frame rate runs from 0.00 to 0.85, the best at bias $+4.0$: $+54.3$ points [$+43.2$, $+65.4$] against the released threshold, mover mask IoU $+4.6$ points, mover ATE/path $+0.0\%$ [$+0.0$, $+0.0$] and leading-vehicle ATE/path $+0.0\%$ [$+0.0$, $+0.0$] (the camera output of MoRe does not depend on the mask threshold, so the camera terms of the regression bar are unchanged by construction and the mover mask IoU is the operative one). At that threshold the hands masks mark 0.136 of the camera-attached foreground and 0.026 of the background, 0.051 of the frame (released: 0.011, 0.007, 0.019), and the mover IoU is 0.298 against 0.252 released: the released motion head already ranks the hands above the background and its threshold leaves both near zero; the frame-choice rule (the foreground's mask rate above the background's) is met by a sparse mask at a low threshold. The curve: bias $-4.0$ (threshold 0.98): rate 0.00, mover IoU 0.045; bias $-2.0$ (threshold 0.88): rate 0.16, mover IoU 0.109; bias $-1.0$ (threshold 0.73): rate 0.20, mover IoU 0.172; bias $-0.5$ (threshold 0.62): rate 0.25, mover IoU 0.212; bias $0$ (threshold 0.50): rate 0.31, mover IoU 0.252; bias $+0.5$ (threshold 0.38): rate 0.36, mover IoU 0.270; bias $+1.0$ (threshold 0.27): rate 0.41, mover IoU 0.280; bias $+2.0$ (threshold 0.12): rate 0.59, mover IoU 0.294; bias $+4.0$ (threshold 0.02): rate 0.85, mover IoU 0.298. Reading: a threshold raises the hands rate by at least 45 points with no detected regression, so under the pre-registered reading the recipe's gain is attributable to calibration. The recipe's fine-tunes raise the foreground rate to a dense mask (\cref{tab:recipe}); what the pre-registered reading tests is the frame-choice rate, and it is reported as fixed. The attribution margin $m$ of A20.1 separates the two on the same 81 clips (\texttt{results/phase9/calibration/margins.json}; paired bootstrap over clips): the released MoRe marks 0.011 of the camera-attached foreground and 0.007 of the background, $m = 0.004$ [$-0.003$, $+0.012$], rate 0.31, mover IoU 0.252; at bias $+4$ it marks 0.136 and 0.026, $m = 0.110$ [0.082, 0.140], rate 0.85, mover IoU 0.298; after five rendered sequences (seed 0) it marks 0.368 and 0.032, $m = 0.336$ [0.290, 0.382], rate 0.94, mover IoU 0.431; after 400 sequences 0.217 and 0.018, $m = 0.199$ [0.163, 0.234], rate 0.91, mover IoU 0.350; the VGGT4D head on 25 sequences marks 0.620 and 0.282, $m = 0.337$ [0.279, 0.390], rate 0.88, mover IoU 0.221 (released VGGT4D 0.15, Amendment~19). The main text reports the recipe by the margin and the mover IoU, with the rate alongside, for this reason.

\begin{figure}[t]
  \centering
  \includegraphics[width=\linewidth]{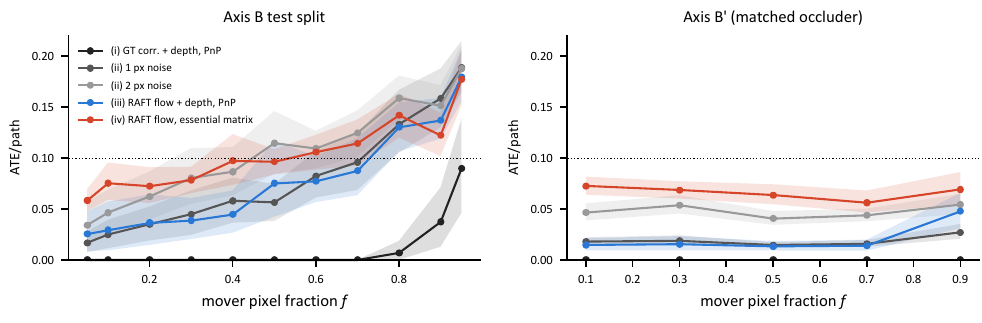}
  \caption{Classical pose estimation on the remaining static evidence (A20.4), Axis~B test split. Mean ATE/path against the dynamic fraction $f$ for the four conditions (exact correspondences, 1 and 2 px noise, RAFT flow with depth, RAFT flow with the five-point essential matrix), 95\% bootstrap intervals; the dotted line is the absolute reading's 0.10.}
  \label{fig:classical}
\end{figure}

%% A20.2 end
%% A20.3 begin
\paragraph{Replication of the Five-Sequence Result (A20.3).} Three further MoRe fine-tunes with the Gate 6b recipe on five Axis~C sequences each, the first five of the seed-1, seed-2 and seed-3 permutations of the 400 rendered sequences (seed 1: axisC\_g70\_train\_s3, axisC\_g10\_escalator\_s103, axisC\_g10\_elevator\_s3, axisC\_g30\_treadmill\_s2, axisC\_g10\_escalator\_s109; seed 2: axisC\_g30\_escalator\_s115, axisC\_g30\_train\_s100, axisC\_g70\_escalator\_s113, axisC\_g50\_boat\_s109, axisC\_g10\_boat\_s109; seed 3: axisC\_g10\_escalator\_s100, axisC\_g30\_elevator\_s104, axisC\_g70\_train\_s103, axisC\_g50\_train\_s109, axisC\_g50\_treadmill\_s114; listed in \texttt{results/phase9/five\_replication/sequences.json} before training), and one on the first five of the seed-0 permutation of the dominant-mover control set, each evaluated once as the A17.2 runs (\cref{tab:recipe}, Figure~4 of the main text). Reading, the A17.2 bar per draw (a hands gain of at least 45 points with an interval excluding zero and no detected regression): seed 0 (Amendment~18): hands rate 0.31 to 0.94 ($+63.0$ points [$+51.9$, $+74.1$]), Axis~C test rate 1.00, no detected regression, meets the bar: yes; seed 1: hands rate 0.31 to 0.91 ($+60.5$ points [$+49.4$, $+71.6$]), Axis~C test rate 0.97, one detected regression, the egocentric mover error $+5.1\%$ [$+3.1$, $+7.7$], meets the bar: no; seed 2: hands rate 0.31 to 0.90 ($+59.3$ points [$+49.4$, $+70.4$]), Axis~C test rate 0.86, no detected regression, meets the bar: yes; seed 3: hands rate 0.31 to 0.91 ($+60.5$ points [$+49.4$, $+71.6$]), Axis~C test rate 0.99, no detected regression, meets the bar: yes. Three of the four five-sequence draws meet the bar: the five-sequence claim holds, and the paper says five sequences in three of four draws (the A20.3 wording note in the protocol). The five-sequence dominant-mover control (seed 0 of the control set) changes the hands rate by $+35.8$ points [$+25.9$, $+46.9$] to 0.67 and the Axis~C test rate to 0.68; A15.3(ii): the control rises by at least 15 points with a CI excluding zero: the gain is attributed to physical-frame motion supervision in general rather than to moving-platform data.
%% A20.3 end
\paragraph{Classical Estimator on Static Evidence (A20.4).} \Cref{fig:classical} gives, on the Axis~B test split (176 sequences, every $f$), the ATE/path of frame-to-frame camera estimation restricted to the ground-truth static region, chained over the 30 frames and Sim(3)-aligned as every ATE number is: (i) ground-truth correspondences with ground-truth depth and PnP (OpenCV SQPnP inside RANSAC, refined on the inliers), (ii) the same with Gaussian pixel noise of 1 and 2 px, (iii) RAFT flow (torchvision release, default weights) restricted to the static region with ground-truth depth and PnP, and (iv) the same flow with no depth, the five-point essential matrix (OpenCV, RANSAC) with the scale of every step from the ground-truth path; every condition is oracle-assisted (the ground-truth static region, and ground-truth depth or step length), and A22.9 (\cref{app:a22}) tests a scale-free chain. The degradation thresholds under the absolute reading (ATE/path above 0.10) are: exact correspondences none (the mean stays below 0.10 up to $f=0.95$, 0.090); 1 px noise 0.8; 2 px noise 0.5; flow with depth 0.8; flow without depth 0.6, the condition the reading names, whose curve runs from 0.059 at $f=0.05$ through 0.097 at $f=0.5$ to 0.177 at $f=0.95$. The networks' absolute thresholds run from 0.2 to 0.9 (\cref{tab:leaderboard}), ten of the fourteen at or below $f=0.5$, so the statement the curves support is the mixed one: the evidence is there up to $f=0.5$ and insufficient above it (superseded by A22.9 and the main text's bounded reading: the controls bound conditioning, not what any estimator could recover). Two implementation details are recorded in the attempts table of the results log: OpenCV's EPnP mis-solves the near-planar static region (a first run gave thresholds of 0.05 to 0.2 for exact correspondences), and the cheirality test of \texttt{recoverPose} discards every triangulated point farther than fifty baselines, which under centimetre steps is every point; the final run uses SQPnP and an unbounded cheirality distance (\texttt{results/phase9/classical/}).

\begin{figure*}[t]
  \centering
  \includegraphics[width=\textwidth]{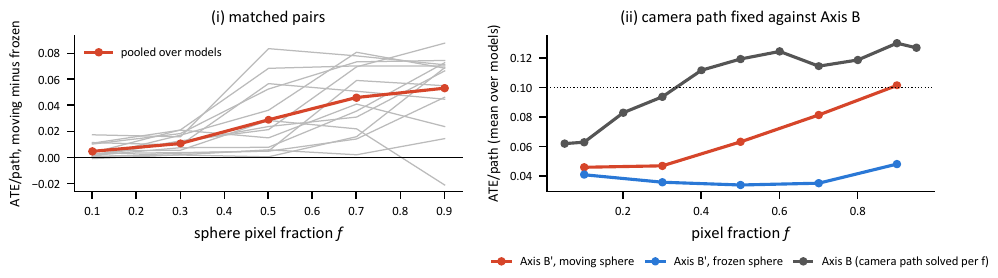}
  \caption{Matched occluder control, Axis~B$'$ (A20.5). Left: the paired difference moving minus frozen per model and $f$ (cluster bootstrap over the five scene seeds; filled where the interval excludes zero). Right: ATE/path against the covered fraction $f$ for the moving and the frozen member of each pair, pooled over models, with the Axis~B curve at the same fractions.}
  \label{fig:matched}
\end{figure*}

\paragraph{Matched Occluder Control, Axis~B$'$ (A20.5).} For five Axis~B scene seeds the camera path and the background are held at their $f=0.10$ configuration (the solved camera path of the abstract-class sequence of each seed; the camera never moves closer) and a textured sphere four metres ahead of the mid-path camera covers $f \in \{0.1, 0.3, 0.5, 0.7, 0.9\}$ of the pixels by its radius alone, solved per sequence (\texttt{whomoved/render/axis\_b\_matched.py}, solve mode \texttt{scale\_fixed}). Each configuration is rendered twice, with the sphere spinning about its centre at $30^\circ$ per second (moving) and frozen, and with a sparse and a dense texture: 100 sequences of 30 frames with the Axis~B ground truth (index \texttt{data/bench/axis\_b\_matched\_index.parquet}). Within a pair the silhouette and the visible static background are identical by construction; the check on the rendered instance masks records the minimum per-frame IoU of the sphere mask between the two members (\texttt{results/phase9/matched/silhouette.json}). The set enters the extended release as Axis~B$'$ (\cref{app:release}).
%% A20.5 readings begin
\paragraph{Axis~B$'$ Readings (A20.5).} The silhouette check records a minimum per-frame IoU of the sphere mask between the two members of 0.9991 over the 50 pairs (50 above 0.99 on every frame). 1399 evaluation runs over the fourteen pose models (TraceAnything, which carries no camera, is evaluated and excluded from the camera readings; \texttt{results/runs/diag\_bm\_*}). (i) The paired difference moving minus frozen in ATE/path, pooled over models with a cluster bootstrap over the five scene seeds, is $+0.0289$ [$+0.0237$, $+0.0343$] over all $f$ (699 pairs); per $f$: $f=0.1$: $+0.0050$ [$+0.0016$, $+0.0083$]; $f=0.3$: $+0.0110$ [$+0.0046$, $+0.0175$]; $f=0.5$: $+0.0291$ [$+0.0213$, $+0.0412$]; $f=0.7$: $+0.0461$ [$+0.0316$, $+0.0573$]; $f=0.9$: $+0.0533$ [$+0.0427$, $+0.0657$]; the reading is the moving half worse at $f \in \{0.1, 0.3, 0.5, 0.7, 0.9\}$, the attributional reading at those fractions (per-model curves in \cref{fig:matched}). (ii) The ATE/path curve against $f$ with the camera path fixed, against the Axis~B curve at the same fractions: $f=0.1$: moving 0.046, frozen 0.041, Axis~B 0.063; $f=0.3$: moving 0.047, frozen 0.036, Axis~B 0.094; $f=0.5$: moving 0.063, frozen 0.034, Axis~B 0.119; $f=0.7$: moving 0.082, frozen 0.035, Axis~B 0.115; $f=0.9$: moving 0.102, frozen 0.048, Axis~B 0.130; the paired difference as a share of the frozen error, aggregated as the mean over pairs of the within-pair ratio, is 35\% at $f=0.1$, 113\% at 0.3, 299\% at 0.5, 833\% at 0.7 and 699\% at 0.9; the change of the pooled means, mean moving minus mean frozen over mean frozen, is 12\%, 30\%, 85\%, 130\% and 110\% at the same fractions; the main text reports the paired absolute effects (A22.10). (iii) With the true static region supplied on the moving half, the four native-mask models change ATE/path by $-15.0\%$ [$-22.4$, $-7.3$] over 199 pairs, read as in Gate 3b-2 (per $f$: $+4.5\%$ at 0.1, $+16.1\%$ at 0.3, $-30.5\%$ [$-48.0$, $-11.8$] at 0.5, $-25.7\%$ at 0.7, $-13.1\%$ at 0.9; MegaSaM $-43.5\%$, MonST3R $-23.1\%$, VGGT4D $-6.7\%$, Easi3R $-0.4\%$).

%% A20.5 readings end
\paragraph{Resampling Unit (A20.6).} \Cref{tab:cluster} lists every pooled interval of the main text as recorded, with the sequence as the resampling unit, next to its recomputation with the scene seed (synthetic), the recording (egocentric) or the drive (KITTI) as the unit, model predictions kept together within a sampled unit. The main text carries the cluster intervals. The intervals widen where the units are few (four scene seeds on Axis~C; eight recordings for the hands clips, whose pooled true-static-region change becomes $-6.7\%$ [$-19.7$, $+0.6$]) and are unchanged in sign everywhere; the 1945-pair validation ceiling, read over seventeen units, becomes $+1.9\%$ [$-6.9$, $+12.1$].

% generated by scripts/phase9_cluster.py (Amendment 20, A20.6); results/phase9/cluster/readings.json; do not edit
\begin{table*}[t]
  \centering
  \scriptsize
  \caption{Resampling unit (Amendment~20, A20.6). Every pooled interval of the main text as recorded (the sequence as the resampling unit) and recomputed with the scene seed, the recording or the drive as the unit, model predictions kept together within a sampled unit (cluster bootstrap, 2000 resamples). The main text carries the cluster intervals; no verdict is re-decided.}
  \label{tab:cluster}
  \resizebox{\textwidth}{!}{\begin{tabular}{lllc}
    \toprule
    quantity & sequence-level interval & cluster interval & unit (count) \\
    \midrule
    Axis B above threshold, true static region, native-mask models & $+7.5\%$ [$+2.4$, $+13.0$] & $+7.5\%$ [$+3.9$, $+12.8$] & scene seed (4) \\
    Axis B above threshold, true static region, all eleven & $+17.1\%$ [$+13.9$, $+20.3$] & $+17.1\%$ [$+14.4$, $+19.4$] & scene seed (4) \\
    Axis C, true static region, native-mask models & $-12.8\%$ [$-17.2$, $-8.5$] & $-12.8\%$ [$-14.9$, $-10.8$] & scene seed (4) \\
    Axis C, true static region, all eleven & $-10.3\%$ [$-12.7$, $-7.7$] & $-10.3\%$ [$-11.7$, $-8.6$] & scene seed (4) \\
    Axis B below threshold, true static region, all eleven & $+5.0\%$ [$+2.0$, $+8.1$] & $+5.0\%$ [$+1.6$, $+8.6$] & scene seed (4) \\
    Axis B Spearman of the change with $1-f$ & $-0.23$ [$-0.29$, $-0.16$] & $-0.23$ [$-0.31$, $-0.15$] & scene seed (4) \\
    per-sequence oracle over the decompositions & $-7.7\%$ [$-8.5$, $-7.0$] & $-7.8\%$ [$-8.3$, $-7.3$] & scene seed (8) \\
    true static region over the 1945 validation pairs & $+1.9\%$ [$-0.3$, $+4.1$] & $+1.9\%$ [$-6.9$, $+12.1$] & scene seed (MOVi-F rows: the sequence) (17) \\
    hands clips, true static region & $-6.7\%$ [$-10.7$, $-2.5$] & $-6.7\%$ [$-19.7$, $+0.6$] & recording (8) \\
    mover clips, true static region & $-6.0\%$ [$-9.3$, $-2.9$] & $-6.0\%$ [$-14.8$, $+1.8$] & recording (16) \\
    KITTI leading-vehicle clips, static region & $+2.1\%$ [$-2.2$, $+6.6$] & $+2.1\%$ [$-3.8$, $+8.0$] & drive (10) \\
    hands clips, top selection bin, true static region & $-14.6\%$ [$-24.7$, $-4.4$] & $-14.6\%$ [$-26.8$, $-0.8$] & recording (6) \\
    egocentric ambiguous over standard ratio & 1.59 [1.30, 1.97] & 1.59 [1.29, 2.24] & recording (17) \\
    KITTI ambiguous over standard ratio, all 83 clips & 3.11 [1.52, 6.29] & 3.11 [0.88, 6.59] & drive (18) \\
    KITTI ratio, 67 moving-ego clips & 1.33 [0.92, 1.90] & 1.33 [0.91, 1.65] & drive (15) \\
    ratio pooled over both real sets & 1.95 [1.52, 2.52] & 1.95 [1.44, 2.87] & recording or drive (35) \\
    MoRe, 5 rendered: hands rate (points) & n/a & $+63.0$ [$+50.8$, $+85.4$] & recording (8) \\
    MoRe, 25 rendered: hands rate (points) & n/a & $+60.5$ [$+49.1$, $+80.5$] & recording (8) \\
    MoRe, 400 rendered: hands rate (points) & n/a & $+60.5$ [$+49.1$, $+80.5$] & recording (8) \\
    VGGT4D head, 25 rendered: hands rate (points) & n/a & $+54.3$ [$+45.5$, $+68.4$] & recording (8) \\
    \bottomrule
  \end{tabular}}
\end{table*}

\paragraph{Score Check (A20.7).} The partial-correlation check of the ambiguity score is reported with the existing benchmarks in \cref{app:existing}.

\section{Measurement Audits}
\label{app:a22}
Amendment~22 (8 November 2026) fixed nine readings and the paper's wording for each outcome before the analyses ran, from cached outputs and ground truth (no 4D model run); every reading is in \texttt{results/phase10/<item>/readings.json} and registered in \texttt{NUMBERS.md} (F16). Nothing in it changes a hypothesis threshold or re-decides a pre-registered verdict.

\begin{figure*}[t]
  \centering
  \includegraphics[width=\textwidth]{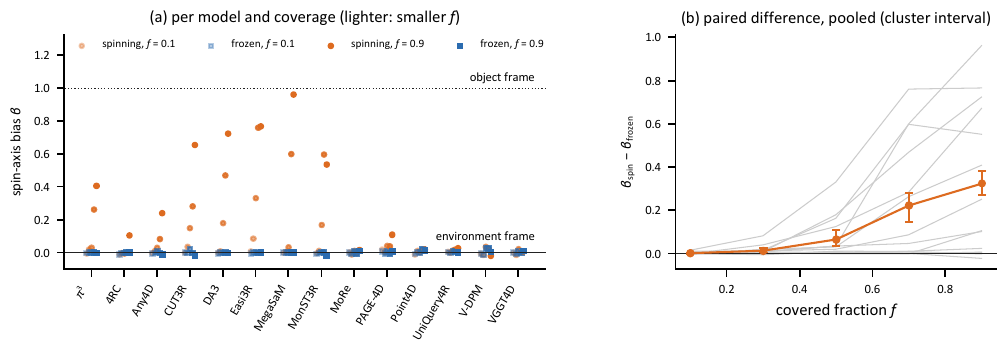}
  \caption{Directional analysis of the matched control (A22.1). (a) The spin-axis bias $\beta$ per model and covered fraction, spinning (circles) against frozen (squares); 0 is the environment-frame camera, 1 the camera that follows the sphere. (b) The paired difference spinning minus frozen per $f$, pooled over models with its cluster interval over scene seeds; grey lines are the models.}
  \label{fig:directional}
\end{figure*}

\paragraph{Directional Analysis of the Matched Control (A22.1).} For every Axis~B$'$ pair, every pose model and every consecutive frame pair, the predicted relative camera rotation is compared with the ground-truth increment in the environment frame and with the increment of the same camera expressed in the sphere's frame (the camera pose premultiplied by the inverse of the sphere's recorded pose, the generator's seeded spin axis at $30^\circ$ per second, one degree per frame at 30 fps; the recorded axis matches the generator's rule on 50 of 50 clips). The spin-axis bias $b_t$ is the signed component, along the spin axis in the camera frame at $t$, of the rotation that takes the ground-truth increment to the predicted one, normalised by the spin angle: $\beta$ is its mean over increments, 0 for the environment-frame camera and $+1$ for the camera that follows the sphere (the sign is fixed so that the object-frame candidate gives $+1$ on every clip; the frozen member is normalised by its spinning partner's angle). Reading, fixed: a signed bias toward the object frame exists if $\beta_\mathrm{spin} - \beta_\mathrm{frozen}$, pooled over models with a cluster bootstrap over the five scene seeds, has an interval excluding zero on the object-frame side. Result (14 models, 699 pairs): $\beta_\mathrm{spin} = +0.128$ [$+0.098$, $+0.153$], $\beta_\mathrm{frozen} = +0.002$ [$-0.001$, $+0.004$], difference $+0.126$ [$+0.096$, $+0.153$]: the bias exists, and the paper keeps ``the camera absorbs the motion'' and states $\beta$. Per $f$: $f=0.1$: $+0.002$ [$+0.001$, $+0.004$]; $f=0.3$: $+0.014$ [$+0.002$, $+0.026$]; $f=0.5$: $+0.066$ [$+0.034$, $+0.109$]; $f=0.7$: $+0.223$ [$+0.145$, $+0.282$]; $f=0.9$: $+0.325$ [$+0.274$, $+0.380$]; 11 of 14 models have a positive per-model interval (\cref{fig:directional}). The orthogonal residual is 0.132 degrees per increment on the spinning member against 0.082 on the frozen one, the total angular error 0.269 against 0.107. A22.2 (the spin reversed, reported below) ran afterwards and its reading decides the wording in place of this one.

\paragraph{Camera-Depth Ratio, Constructed Cases (A22.3).} For 40 Axis~C validation sequences (every second of the 80 sorted by name) and the 81 hands clips, predictions were built from the ground truth and labelled by the ratio (\cref{tab:rcd_constructed}): the true camera with the foreground depth scaled by 3 reads egocentric on every sequence (median ratio 0.33) and by 1/3 physical (3.00); the true path with its direction reversed, rotated by 90 degrees about the vertical, or time-warped at the same length, reads physical (1.00) as does the environment-frame reconstruction; the object-frame reconstruction (camera static, the scene carrying the motion) reads egocentric (0.00). The ratio reacts to the foreground depth and to the camera's travel and to nothing else, which is why the paper renames it the camera-depth ratio $r_\mathrm{cd}$, calls it a diagnostic and states the direct observation wherever it read the geometry as choosing a world. The background-normalised camera label of A21.1 (predicted path over predicted background depth against the ground truth, physical above 0.5) reacts to the travel alone (\cref{tab:rcd_constructed}); on the released models it reads physical on 0.37 to 0.85 of the Axis~C sequences and 0.98 to 1.00 of the hands clips, agreeing with the mask on 0.37 to 0.88 and 0.23 to 0.95 (\cref{tab:camlabel_released}); reported here under A22.3 and adopted as the main text's camera label (Amendment~23, next paragraph).

\paragraph{The Camera Label (Amendment~23, A23.1).} Under Amendment~23 the main text's camera label is the background-normalised label validated above (predicted camera travel over predicted median background depth, against the same from the ground truth, physical above 0.5), and $r_\mathrm{cd}$ stays in this supplement as the depth diagnostic (\cref{tab:agreement_full} carries both). On the released models the camera label reads the physical frame on 0.37 to 0.85 of the Axis~C sequences and 0.98 to 1.00 of the hands clips (the mask: 0.16 to 0.78 and 0.23 to 0.95), and agrees with the mask on 0.37 to 0.88 and 0.23 to 0.95 (Table~3 of the main text; the label is undefined on 4 Axis~C treadmill sequences at $g = 0.70$, where no background pixel has a finite depth, and those are excluded from the camera and agreement columns while the mask column keeps the A20.1 denominator). Per recipe checkpoint on the hands clips the camera label reads physical on: MoRe, released 1.00 (mask 0.31, agreement 0.31); MoRe, 5 rendered 1.00 (mask 0.94, agreement 0.94); MoRe, 25 rendered 1.00 (mask 0.91, agreement 0.91); MoRe, 400 rendered 1.00 (mask 0.91, agreement 0.91); MoRe, 400 control 1.00 (mask 0.94, agreement 0.94); MoRe, real labels 1.00 (mask 0.96, agreement 0.96); VGGT4D, released 1.00 (mask 0.33, agreement 0.33); VGGT4D head, 25 rendered 1.00 (mask 0.88, agreement 0.88); VGGT4D head, 25 control 1.00 (mask 0.85, agreement 0.85); the five MoRe fine-tunes span 1.00 to 1.00 against 1.00 released (Figure~4c of the main text). The downstream foreground-trajectory error of A17.5 (median error over path, released models pooled) by agreement cell of the mask and the camera label: on Axis~C both egocentric 0.85 ($n = 564$), mask egocentric, camera physical 0.47 ($n = 506$), mask physical, camera egocentric 0.63 ($n = 275$), both physical 0.36 ($n = 831$); on the hands clips both egocentric 0.42 ($n = 1$), mask egocentric, camera physical 0.52 ($n = 265$), mask physical, camera egocentric 0.19 ($n = 1$), both physical 0.41 ($n = 847$). On the hands clips the camera label reads egocentric on one run in each of its two cells, so a cell is eligible to be the lowest when it holds at least 10 runs (Amendment~23.2, written and signed after the one-run cell was observed; both readings are recorded in \texttt{results/phase10/camlabel/readings.json}). Under the literal rule the lowest hands cell is the one-run cell (mask physical, camera egocentric, 0.19, $n = 1$); under the eligibility rule it is the cell in which both read the environment frame (0.41, $n = 847$). Reading, fixed in advance and applied over the eligible cells: the cell in which both outputs read the environment frame has the lowest median on both sets, so the main text's sentence is the corresponding one of the three fixed forms. No pre-registered verdict changes.

% generated by scripts/phase10_rcd_validation.py (Amendment 22, A22.3); traced in paper/NUMBERS.md (F16); do not edit
\begin{table*}[t]
  \centering
  \scriptsize
  \setlength{\tabcolsep}{3pt}
  \caption{Constructed predictions under the camera-depth ratio $r_\mathrm{cd}$ and the background-normalised camera label (Amendment~22, A22.3). For 40 Axis~C validation sequences and the 81 hands clips, predictions built from the ground truth; the median ratio and the share labelled physical (ratio above 0.5) under each diagnostic. The ratio reacts to the foreground depth (1a, 1b) and to the camera's travel (5) and to nothing else; the camera label reacts to the travel alone.}
  \label{tab:rcd_constructed}
  \resizebox{0.7\textwidth}{!}{\begin{tabular}{@{}lrrrr|rrrr@{}}
    \toprule
    & \multicolumn{4}{c|}{Axis C validation (40)} & \multicolumn{4}{c}{hands clips (81)} \\
    constructed prediction & $r_\mathrm{cd}$ & phys. & cam. & phys. & $r_\mathrm{cd}$ & phys. & cam. & phys. \\
    \midrule
    (1a) true camera, fg depth $\times 3$ & 0.33 & 0.00 & 1.00 & 1.00 & 0.33 & 0.00 & 1.00 & 1.00 \\
    (1b) true camera, fg depth $\times 1/3$ & 3.00 & 1.00 & 1.00 & 1.00 & 3.00 & 1.00 & 1.00 & 1.00 \\
    (2a) path reversed & 1.00 & 1.00 & 1.00 & 1.00 & 1.00 & 1.00 & 1.00 & 1.00 \\
    (2b) path rotated $90^\circ$ & 1.00 & 1.00 & 1.00 & 1.00 & 1.00 & 1.00 & 1.00 & 1.00 \\
    (3) time warped, same length & 1.00 & 1.00 & 1.00 & 1.00 & 1.00 & 1.00 & 1.00 & 1.00 \\
    (4) environment-frame reconstruction & 1.00 & 1.00 & 1.00 & 1.00 & 1.00 & 1.00 & 1.00 & 1.00 \\
    (5) object-frame reconstruction & 0.00 & 0.00 & 0.00 & 0.00 & 0.00 & 0.00 & 0.00 & 0.00 \\
    \bottomrule
  \end{tabular}}
\end{table*}

% generated by scripts/phase10_rcd_validation.py (Amendment 22, A22.3); traced in paper/NUMBERS.md (F16); do not edit
\begin{table}[t]
  \centering
  \scriptsize
  \setlength{\tabcolsep}{3pt}
  \caption{The background-normalised camera label of A21.1 on every released model (Amendment~22, A22.3): the share of sequences on which the predicted path over the predicted background depth, against the ground truth, exceeds 0.5 (cam.), next to the mask's physical-frame share (mask) and the camera-depth ratio's (r$_\mathrm{cd}$), and the agreement of the camera label with the mask. Reported in the supplement; the main text uses the mask and the background-normalised camera label (Amendment~23); $r_\mathrm{cd}$ stays here as the depth diagnostic.}
  \label{tab:camlabel_released}
  \resizebox{\linewidth}{!}{\begin{tabular}{@{}lrrrr|rrrr@{}}
    \toprule
    & \multicolumn{4}{c|}{Axis C (160)} & \multicolumn{4}{c}{hands clips (81)} \\
    model & mask & $r_\mathrm{cd}$ & cam. & agree & mask & $r_\mathrm{cd}$ & cam. & agree \\
    \midrule
    MonST3R & 0.61 & 0.14 & 0.48 & 0.64 & 0.93 & 0.57 & 1.00 & 0.93 \\
    CUT3R & 0.41 & 0.12 & 0.42 & 0.63 & 0.85 & 0.63 & 1.00 & 0.85 \\
    Easi3R & 0.37 & 0.35 & 0.66 & 0.38 & 0.77 & 0.44 & 1.00 & 0.77 \\
    VGGT4D & 0.14 & 0.06 & 0.70 & 0.37 & 0.33 & 0.81 & 1.00 & 0.33 \\
    $\pi^3$ & 0.65 & 0.35 & 0.85 & 0.67 & 0.85 & 0.88 & 1.00 & 0.85 \\
    Any4D & 0.69 & 0.12 & 0.60 & 0.64 & 0.86 & 0.81 & 1.00 & 0.86 \\
    MegaSaM & 0.35 & 0.03 & 0.37 & 0.88 & 0.89 & 0.64 & 0.98 & 0.89 \\
    Depth Anything 3 & 0.56 & 0.05 & 0.65 & 0.72 & 0.90 & 0.85 & 1.00 & 0.90 \\
    4RC & 0.77 & 0.09 & 0.55 & 0.65 & 0.23 & 0.60 & 1.00 & 0.23 \\
    V-DPM & 0.74 & 0.02 & 0.71 & 0.83 & 0.95 & 0.81 & 1.00 & 0.95 \\
    MoRe & 0.21 & 0.01 & 0.78 & 0.43 & 0.31 & 0.56 & 1.00 & 0.31 \\
    UniQuery4R & 0.39 & 0.00 & 0.43 & 0.67 & 0.94 & 0.81 & 1.00 & 0.94 \\
    Point4D & 0.47 & 0.10 & 0.60 & 0.77 & 0.78 & 0.79 & 1.00 & 0.78 \\
    PAGE-4D & 0.74 & 0.06 & 0.82 & 0.67 & 0.94 & 0.67 & 1.00 & 0.94 \\
    \bottomrule
  \end{tabular}}
\end{table}

\begin{figure*}[t]
  \centering
  \includegraphics[width=\textwidth]{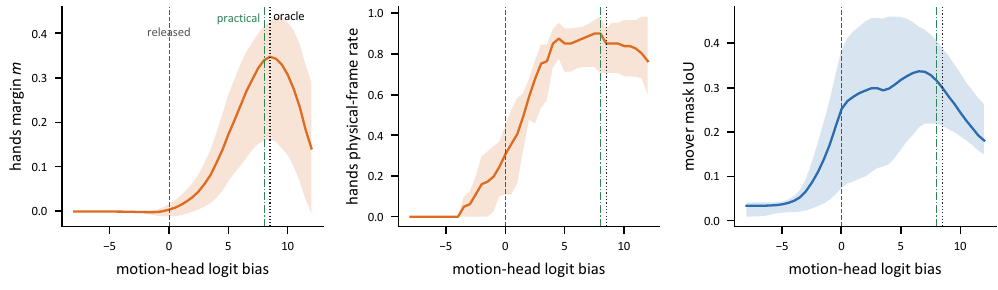}
  \caption{The full calibration envelope of the released MoRe (A22.4): the hands attribution margin, the hands physical-frame rate and the mover mask IoU against the motion-head logit bias, with recording-level intervals; the released threshold (dashed), the oracle threshold (dotted) and the practical threshold selected on the training recordings (dash-dotted).}
  \label{fig:envelope}
\end{figure*}

\paragraph{The Full Calibration Envelope (A22.4).} The released MoRe's motion-head threshold was swept over logit biases from $-8.0$ to $+12.0$ in steps of 0.5 on the cached probability maps of A20.2; the margin peaks at bias $+8.5$, the hands physical-frame rate at $+7.5$ and the mover IoU at $+6.5$, all inside the grid (\cref{fig:envelope}). The oracle threshold (the best margin on the hands clips themselves) is bias $+8.5$, probability threshold 0.0002: margin 0.348 [0.161, 0.426] (recording-level interval over the eight recordings), rate 0.85, mover IoU 0.299, the mask covering 0.547 of the foreground and 0.199 of the background. The practical threshold, selected on the 51 hands clips of the ten recordings that trained the real-label variant (bias $+8.0$, margin 0.332 there), gives 0.354 [0.143, 0.426] on the 30 held-out hands clips (rate 0.87; the oracle bias gives 0.384 on the same clips). The threshold-free ranking statistic, the per-clip AUROC of the mask value for foreground against background pixels, is 0.718 [0.630, 0.773] for the released probability map (a monotone threshold leaves it unchanged); the fine-tunes' runs keep the binary mask alone, whose AUROC is $(1 + \text{foreground rate} - \text{background rate})/2$: MoRe, 5 rendered 0.668 [0.596, 0.704]; MoRe, 25 rendered 0.600 [0.550, 0.632]; MoRe, 400 rendered 0.599 [0.552, 0.630]; MoRe, 400 control 0.641 [0.565, 0.670]; MoRe, real labels 0.713 [0.648, 0.827]; VGGT4D head, 25 rendered 0.669 [0.620, 0.735]; MoRe, released 0.502 [0.494, 0.508]; MoRe, recalibrated at bias $+8.5$ 0.674 [0.580, 0.713]. On binary masks the AUROC restates the margin and says nothing about the ordering of the continuous scores, so the main text does not use it. Reading, fixed: $k = 0.336 / 0.348 = 0.97$, and the paired difference margin(five sequences, seed 0) minus margin(oracle threshold) is $-0.012$ [$-0.055$, $+0.043$] over 81 clips and 8 recordings: $k < 1.33$, so the recipe's margin sentence is dropped from the abstract, Section~4.4 of the main text reports both margins with the interval, and the recipe's contribution is stated by what exceeds recalibration with an interval excluding zero: the mover mask IoU, 0.431 against 0.299 at the oracle threshold ($+0.131$ [$+0.040$, $+0.210$], 105 mover clips), does; the ranking statistic does not (binary against binary $-0.006$ [$-0.027$, $+0.022$]; against the released probabilities $-0.050$ [$-0.095$, $+0.019$]). The margin is a transfer that recalibration matches; the A20.2 verdict on the rate stands in its record.

\paragraph{Per-Draw Margins (A22.5).} \Cref{tab:per_draw} lists, for every scaling checkpoint, the four five-sequence draws, the three single sequences and the dominant-mover controls, the margin with its foreground and background coverage, the physical-frame rate, the mover IoU, the binary-mask ranking statistic, the paired difference against the oracle-threshold margin with a recording-level interval, and the regression record of the A17.2 rule. Reading, fixed: the number of five-sequence draws whose margin exceeds the oracle-threshold margin with an interval excluding zero is 0 of four (seed 0: 0.336, against the oracle $-0.012$ [$-0.055$, $+0.043$]; seed 1: 0.280, against the oracle $-0.068$ [$-0.146$, $-0.004$]; seed 2: 0.171, against the oracle $-0.177$ [$-0.268$, $-0.093$]; seed 3: 0.265, against the oracle $-0.083$ [$-0.193$, $-0.013$]), against three of four by the pre-registered rate rule of A20.3; the abstract and Section~4.4 report the two counts in separate sentences.

% generated by scripts/phase10_per_draw.py (Amendment 22, A22.5); traced in paper/NUMBERS.md (F16); do not edit
\begin{table*}[t]
  \centering
  \scriptsize
  \setlength{\tabcolsep}{3pt}
  \caption{Per-draw margins of every recipe checkpoint (Amendment~22, A22.5). On the 81 hands clips: the mask's coverage of the camera-attached foreground (fg) and of the background (bg), the attribution margin $m$ with a recording-level interval, the physical-frame rate, the mover mask IoU (105 clips), the ranking statistic of the binary mask (AUROC$_\mathrm{bin}$, A22.4), the paired difference of $m$ against the released model recalibrated at its oracle threshold (A22.4, bias +8.5, $m_\mathrm{cal}$ = 0.348) with a recording-level interval, and the checkpoint's regression record under the A17.2 rule (mover and leading-vehicle ATE/path change in percent with intervals; a check mark when no regression was detected). Resampling unit per column: the hands columns (fg, bg, $m$, rate, AUROC$_\mathrm{bin}$, $m - m_\mathrm{cal}$) resample the eight ADT recordings that hold the 81 hands clips; the mover IoU is the mean over the 105 mover clips; the regression columns are the A17.2 record, which resamples clips (the 105 mover clips of 16 recordings, the 41 leading-vehicle clips of 10 drives).}
  \label{tab:per_draw}
  \resizebox{\textwidth}{!}{\begin{tabular}{@{}llrrrrrrrrrc@{}}
    \toprule
    group & checkpoint & fg & bg & $m$ & rate & IoU & AUROC$_\mathrm{bin}$ & $m - m_\mathrm{cal}$ & mover ATE & lead ATE & no regr. \\
    \midrule
    data scaling & 1 rendered (seed 0) & 0.36 & 0.03 & 0.325 [0.195, 0.384] & 0.93 & 0.393 & 0.663 & -0.023 [-0.083, +0.046] & +1.2 [-0.5, +3.1] & +0.5 [-3.5, +5.4] & $\checkmark$ \\
     & 5 rendered (seed 0) & 0.37 & 0.03 & 0.336 [0.192, 0.407] & 0.94 & 0.431 & 0.668 & -0.012 [-0.055, +0.043] & -3.1 [-5.1, -0.9] & +2.5 [-0.1, +6.2] & $\checkmark$ \\
     & 25 rendered & 0.22 & 0.02 & 0.201 [0.100, 0.265] & 0.91 & 0.349 & 0.600 & -0.147 [-0.258, -0.058] & -6.2 [-8.7, -3.8] & -2.0 [-6.1, +0.8] & $\checkmark$ \\
     & 50 rendered & 0.23 & 0.02 & 0.204 [0.105, 0.273] & 0.90 & 0.352 & 0.602 & -0.144 [-0.266, -0.047] & -7.1 [-9.8, -4.6] & -1.5 [-5.3, +1.4] & $\checkmark$ \\
     & 100 rendered & 0.22 & 0.02 & 0.197 [0.099, 0.266] & 0.90 & 0.347 & 0.599 & -0.151 [-0.273, -0.054] & -7.2 [-9.8, -4.8] & -1.9 [-6.0, +0.8] & $\checkmark$ \\
     & 200 rendered & 0.21 & 0.02 & 0.188 [0.093, 0.247] & 0.90 & 0.343 & 0.594 & -0.160 [-0.269, -0.062] & -7.2 [-9.8, -4.9] & -1.1 [-4.9, +1.9] & $\checkmark$ \\
     & 400 rendered & 0.22 & 0.02 & 0.199 [0.104, 0.259] & 0.91 & 0.350 & 0.599 & -0.149 [-0.257, -0.053] & -6.7 [-9.3, -4.4] & -1.3 [-6.2, +2.0] & $\checkmark$ \\
    \midrule
    five-sequence draws & 5 rendered, seed 1 & 0.31 & 0.03 & 0.280 [0.151, 0.346] & 0.91 & 0.395 & 0.640 & -0.068 [-0.146, -0.004] & +5.1 [+3.1, +7.7] & -2.1 [-5.0, +0.2] & no \\
     & 5 rendered, seed 2 & 0.19 & 0.02 & 0.171 [0.067, 0.226] & 0.90 & 0.380 & 0.585 & -0.177 [-0.268, -0.093] & +0.8 [-1.1, +2.8] & +0.3 [-1.1, +2.1] & $\checkmark$ \\
     & 5 rendered, seed 3 & 0.30 & 0.03 & 0.265 [0.135, 0.343] & 0.91 & 0.391 & 0.632 & -0.083 [-0.193, -0.013] & -9.7 [-12.5, -7.1] & +0.4 [-3.9, +3.4] & $\checkmark$ \\
    \midrule
    single sequences & 1 rendered, seed 1 & 0.24 & 0.03 & 0.206 [0.080, 0.273] & 0.93 & 0.373 & 0.603 & -0.142 [-0.242, -0.074] & +2.7 [+0.5, +5.3] & +0.5 [-3.6, +3.8] & no \\
     & 1 rendered, seed 2 & 0.04 & 0.01 & 0.037 [-0.002, 0.070] & 0.56 & 0.250 & 0.519 & -0.311 [-0.402, -0.158] & -4.2 [-7.4, -0.9] & +0.6 [-3.3, +4.3] & $\checkmark$ \\
    \midrule
    dominant-mover controls & 400 dominant-mover control & 0.32 & 0.04 & 0.281 [0.130, 0.339] & 0.94 & 0.388 & 0.641 & -0.067 [-0.119, -0.016] & -2.2 [-4.1, -0.0] & +3.6 [-0.1, +8.7] & $\checkmark$ \\
     & 5 dominant-mover control & 0.07 & 0.01 & 0.053 [0.001, 0.094] & 0.67 & 0.297 & 0.527 & -0.295 [-0.392, -0.154] & -0.4 [-2.9, +2.4] & +5.1 [+1.5, +11.9] & no \\
     & 1 dominant-mover control & 0.01 & 0.01 & 0.006 [-0.012, 0.018] & 0.37 & 0.243 & 0.503 & -0.342 [-0.420, -0.170] & +6.4 [+3.0, +10.2] & +2.8 [-0.3, +5.6] & no \\
    \bottomrule
  \end{tabular}}
\end{table*}

\paragraph{One Uncertainty Policy (A22.6).} \Cref{tab:uncertainty} recomputes every margin, calibration and factor interval with the recording as the resampling unit (eight recordings hold the 81 hands clips), next to the clip-level interval of the record. The five-sequence margin (seed 0) is 0.336 [0.192, 0.407] at the recording level against [0.290, 0.382] at the clip level; per recording it runs from 0.015 to 0.558 (decoration seq132 M1292 0.015 on 2, decoration seq135 M1292 0.430 on 19, decoration seq137 M1292 0.115 on 4, meal seq149 M1292 0.459 on 5, multiskeleton party seq115 71292 0.184 on 15, multiskeleton party seq125 71292 0.373 on 34, multiuser cook seq150 M1292 0.558 on 1, work skeleton seq138 M1292 0.260 on 1 clips), and leaving one recording out moves it between 0.307 and 0.371. The pooled Axis~B$'$ difference is $+0.0289$ per seed $+0.0343$, $+0.0261$, $+0.0377$, $+0.0242$, $+0.0221$, and leaving one seed out keeps it between $+0.0267$ and $+0.0306$. The A20.2 recalibration at bias $+4$ keeps its sign at the recording level (hands rate $+54.3$ [$+44.8$, $+68.6$] points, mover IoU $+4.6$ [$+2.1$, $+13.3$]). Pre-registered verdicts stay in their record and are not re-decided.

% generated by scripts/phase10_uncertainty.py (Amendment 22, A22.6); traced in paper/NUMBERS.md (F16); do not edit
\begin{table}[t]
  \centering
  \scriptsize
  \setlength{\tabcolsep}{3pt}
  \caption{One uncertainty policy (Amendment~22, A22.6). Intervals of the record (clip-level bootstrap) next to the same quantity with the recording as the resampling unit (eight ADT recordings hold the 81 hands clips; 2000 resamples). Margins on the hands clips, the A20.2 recalibration at bias $+4$ against the released threshold, the A21.1 factors (median log ratio against the released checkpoint). No verdict is re-decided.}
  \label{tab:uncertainty}
  \resizebox{\linewidth}{!}{\begin{tabular}{@{}lrrr@{}}
    \toprule
    quantity & estimate & clip-level & recording-level \\
    \midrule
    margin, MoRe, released & 0.004 & [-0.003, 0.012] & [-0.012, 0.015] \\
    margin, MoRe, released, logit bias +4 (A20.2) & 0.110 & [0.082, 0.140] & [0.030, 0.176] \\
    margin, MoRe, 5 rendered (seed 0) & 0.336 & [0.290, 0.382] & [0.192, 0.407] \\
    margin, MoRe, 400 rendered & 0.199 & [0.163, 0.234] & [0.104, 0.259] \\
    margin, VGGT4D head, 25 rendered & 0.337 & [0.279, 0.390] & [0.241, 0.470] \\
    margin, MoRe, 25 rendered & 0.201 & [0.165, 0.237] & [0.100, 0.265] \\
    margin, MoRe, 400 control & 0.281 & [0.239, 0.323] & [0.130, 0.339] \\
    margin, MoRe, real labels & 0.426 & [0.366, 0.484] & [0.296, 0.654] \\
    \midrule
    recalibration $+4$: hands rate (points) & +54.3 & [+43.2, +65.4] & [+44.8, +68.6] \\
    recalibration $+4$: mover IoU (points) & +4.6 & [+1.7, +7.5] & [+2.1, +13.3] \\
    recalibration $+4$: mover ATE/path (\%) & +0.0 & [+0.0, +0.0] & [+0.0, +0.0] \\
    \midrule
    camera factor, MoRe, 5 rendered & -0.109 & [-0.150, -0.046] & [-0.155, -0.008] \\
    foreground factor, MoRe, 5 rendered & -0.660 & [-0.768, -0.620] & [-0.821, -0.612] \\
    camera factor, MoRe, 25 rendered & -0.033 & [-0.059, -0.017] & [-0.078, +0.039] \\
    foreground factor, MoRe, 25 rendered & +0.007 & [-0.134, +0.110] & [-0.146, +0.327] \\
    camera factor, MoRe, 400 rendered & +0.005 & [-0.011, +0.026] & [-0.030, +0.065] \\
    foreground factor, MoRe, 400 rendered & -0.149 & [-0.218, -0.054] & [-0.256, +0.067] \\
    camera factor, MoRe, 400 control & -0.093 & [-0.130, -0.009] & [-0.141, +0.031] \\
    foreground factor, MoRe, 400 control & -0.683 & [-0.716, -0.664] & [-0.720, -0.614] \\
    camera factor, MoRe, real labels & +0.173 & [+0.137, +0.197] & [+0.125, +0.227] \\
    foreground factor, MoRe, real labels & -0.258 & [-0.290, -0.232] & [-0.312, -0.194] \\
    \bottomrule
  \end{tabular}}
\end{table}

\paragraph{Ties (A22.7).} \Cref{tab:ties} recomputes the agreement of A20.1 with ties ($|m| < 0.02$) excluded and the denominator stated: over the released models, Axis~C has 515 ties of 2232 runs and the agreement runs from 0.28 to 0.82 with ties retained and 0.25 to 0.84 with them excluded; the hands clips have 255 ties of 1134, 0.32 to 0.79 against 0.40 to 0.79.

% generated by scripts/phase10_ties.py (Amendment 22, A22.7); traced in paper/NUMBERS.md (F16); do not edit
\begin{table}[t]
  \centering
  \scriptsize
  \setlength{\tabcolsep}{2.5pt}
  \caption{Agreement of the mask and the camera-depth ratio with ties excluded (Amendment~22, A22.7). Per released model and set: the number of runs with a ratio ($n$), the ties ($|m| < 0.02$), the agreement with ties retained (the record, A20.1) and with ties excluded (denominator $n$ minus ties).}
  \label{tab:ties}
  \resizebox{\linewidth}{!}{\begin{tabular}{@{}lrrrr|rrrr@{}}
    \toprule
    & \multicolumn{4}{c|}{Axis C} & \multicolumn{4}{c}{hands clips} \\
    model & $n$ & ties & retained & excluded & $n$ & ties & retained & excluded \\
    \midrule
    MonST3R & 157 & 8 & 0.45 & 0.44 & 81 & 9 & 0.49 & 0.53 \\
    CUT3R & 160 & 15 & 0.57 & 0.57 & 81 & 3 & 0.48 & 0.47 \\
    Easi3R & 155 & 10 & 0.44 & 0.43 & 81 & 10 & 0.43 & 0.44 \\
    VGGT4D & 160 & 22 & 0.82 & 0.84 & 81 & 38 & 0.32 & 0.40 \\
    $\pi^3$ & 160 & 25 & 0.53 & 0.53 & 81 & 13 & 0.75 & 0.79 \\
    Any4D & 160 & 64 & 0.38 & 0.36 & 81 & 4 & 0.70 & 0.69 \\
    MegaSaM & 160 & 22 & 0.65 & 0.69 & 81 & 8 & 0.56 & 0.58 \\
    Depth Anything 3 & 160 & 18 & 0.44 & 0.44 & 81 & 2 & 0.75 & 0.75 \\
    4RC & 160 & 50 & 0.29 & 0.25 & 81 & 65 & 0.56 & 0.75 \\
    V-DPM & 160 & 22 & 0.28 & 0.25 & 81 & 7 & 0.79 & 0.78 \\
    MoRe & 160 & 79 & 0.78 & 0.73 & 81 & 66 & 0.51 & 0.73 \\
    UniQuery4R & 160 & 118 & 0.60 & 0.62 & 81 & 20 & 0.75 & 0.79 \\
    Point4D & 160 & 13 & 0.61 & 0.62 & 81 & 4 & 0.57 & 0.58 \\
    PAGE-4D & 160 & 49 & 0.29 & 0.30 & 81 & 6 & 0.63 & 0.64 \\
    \midrule
    MoRe, released & 80 & 41 & 0.81 & 0.82 & 81 & 66 & 0.51 & 0.73 \\
    MoRe, 5 rendered & 80 & 1 & 0.56 & 0.57 & 81 & 2 & 0.80 & 0.81 \\
    MoRe, 25 rendered & 80 & 0 & 0.55 & 0.55 & 81 & 11 & 0.46 & 0.47 \\
    MoRe, 400 rendered & 80 & 0 & 0.54 & 0.54 & 81 & 10 & 0.52 & 0.55 \\
    MoRe, 400 control & 80 & 0 & 0.31 & 0.31 & 81 & 3 & 0.84 & 0.85 \\
    MoRe, real labels & 80 & 21 & 0.49 & 0.56 & 81 & 2 & 0.77 & 0.78 \\
    VGGT4D, released & 80 & 7 & 0.85 & 0.86 & 81 & 38 & 0.32 & 0.40 \\
    VGGT4D head, 25 rendered & 80 & 0 & 0.06 & 0.06 & 81 & 2 & 0.69 & 0.70 \\
    VGGT4D head, 25 control & 80 & 1 & 0.10 & 0.10 & 81 & 2 & 0.68 & 0.70 \\
    \bottomrule
  \end{tabular}}
\end{table}

\paragraph{The Oracle Residual (A22.8).} For the exact-correspondence PnP control at $f=0.95$ (16 Axis~B test sequences, a static fraction of 0.021 of the frame): the exact correspondences reproject onto themselves exactly (a residual of 0 px), and the rendered ground truth is self-consistent at the landing pixel (median relative depth error 0.0029, at most 0.0293 over sequences; surface reprojection 0.41 px, the pixel-centre quantisation). The single-step PnP errors have a median of 0.0000 degrees and 0.0000 of the step, with maxima of 149 degrees and 4392 steps on degenerate frame pairs, and 10.6 of 29 steps fail per sequence on average (too few static pixels for the solver); the chained estimate has an absolute ATE of 0.0028 m (median; mean 0.0148) over a path of 0.071 m, ATE/path 0.090 as in the A20.4 record. Chaining the ground-truth relative transforms recovers the ground-truth trajectory to $9 \times 10^{-15}$ m.

\begin{figure}[t]
  \centering
  \includegraphics[width=\linewidth]{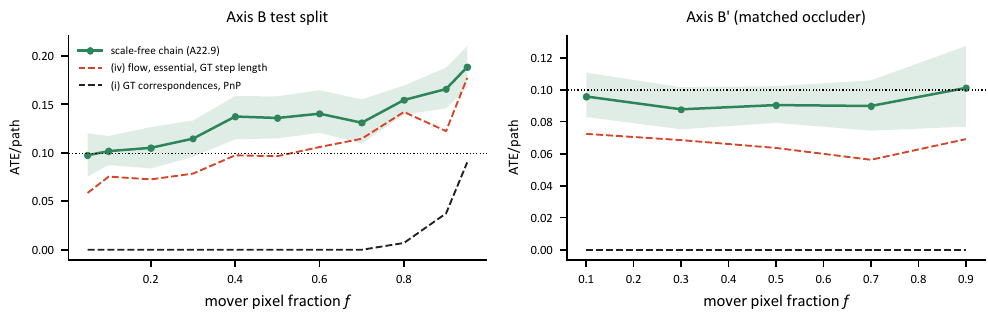}
  \caption{The scale-free classical chain (A22.9) against the two oracle-assisted conditions of A20.4 on the Axis~B test split and on Axis~B$'$: mean ATE/path against the covered fraction with 95\% bootstrap intervals; the dotted line is the operational threshold of 0.10.}
  \label{fig:scalefree}
\end{figure}

\paragraph{A Scale-Free Classical Chain (A22.9).} RAFT flow on the ground-truth static region, the five-point essential matrix, the relative scale of every step carried by triangulated static points tracked from the previous pair (the median ratio of their depths in the two triangulations), one Sim(3) at the end and no ground-truth depth or per-step scale (\texttt{scripts/phase10\_scalefree.py}; 176 Axis~B test and 100 Axis~B$'$ sequences, time box four hours, about eight minutes used). ATE/path on Axis~B: $f=0.05$: 0.097 [0.075, 0.120]; $f=0.1$: 0.102 [0.087, 0.117]; $f=0.2$: 0.105 [0.084, 0.127]; $f=0.3$: 0.115 [0.096, 0.133]; $f=0.4$: 0.138 [0.114, 0.159]; $f=0.5$: 0.136 [0.115, 0.158]; $f=0.6$: 0.140 [0.121, 0.165]; $f=0.7$: 0.131 [0.110, 0.155]; $f=0.8$: 0.155 [0.140, 0.170]; $f=0.9$: 0.166 [0.146, 0.188]; $f=0.95$: 0.189 [0.168, 0.210]; on Axis~B$'$: $f=0.1$: 0.096 [0.083, 0.111]; $f=0.3$: 0.088 [0.075, 0.102]; $f=0.5$: 0.090 [0.079, 0.102]; $f=0.7$: 0.090 [0.074, 0.106]; $f=0.9$: 0.101 [0.077, 0.127] (frozen members 0.052 to 0.096, spinning 0.096 to 0.151; \cref{fig:scalefree}). Reading, fixed: the chain does not stay below 0.10 up to $f=0.5$ on Axis~B nor at every $f$ on Axis~B$'$, so every classical control of this paper is described as oracle-assisted and the positive-control sentence of Section~4.2 says so; the scale drift of a monocular chain, not the correspondences, is what the oracle step length removes.

\begin{figure}[t]
  \centering
  \includegraphics[width=\linewidth]{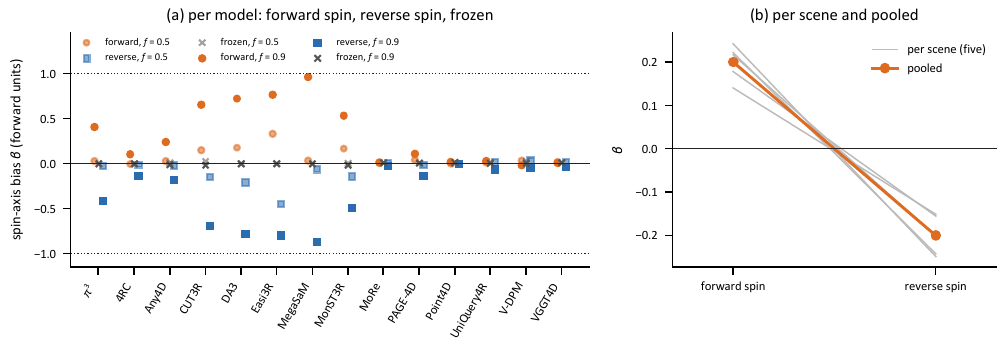}
  \caption{Opposite-spin control (A22.2). (a) The spin-axis bias $\beta$ per model on the forward-spin (circles), the reverse-spin (squares) and the frozen (crosses) member of the twenty cells, in forward units ($+1$ and $-1$ are the object-frame cameras). (b) Forward against reverse per scene (grey) and pooled over models and scenes (orange).}
  \label{fig:opposite_spin}
\end{figure}

\paragraph{Opposite-Spin Control (A22.2).} Twenty Axis~B$'$ cells (the five scenes at $f \in \{0.5, 0.9\}$, both textures) were rendered again with the sphere's spin reversed: the forward member's specification with every recorded rotation transposed, so the axis, rate, starting orientation, texture, scale, camera trajectory and background are those of the existing pair (render check: silhouette IoU against the forward member at least 0.9992 on every frame, 20 of 20 clips above 0.999; the camera trajectory identical to the frozen member on 20 of 20; the recorded spin axis the forward axis reversed on every clip; the background outside the dilated sphere masks differs from the frozen render by path-tracing noise alone, a mean of 0.25 of 255 over 20\% of the pixels, against 0.25 between the existing forward and frozen members; \texttt{results/phase10/opposite\_spin/}\allowbreak\texttt{render\_check.json}). The fourteen pose models ran as released through the A20.5 harness (279 complete model-cell triples of 280); the forward-spin and frozen outputs are the cached A20.5 runs. $\beta$ is computed as in A22.1 with the forward member's axis and angle for every member, so that the object-frame camera gives $+1$ on the forward and $-1$ on the reverse member (checked per clip). Result: $\beta_\mathrm{forward} = +0.200$ [$+0.165$, $+0.229$], $\beta_\mathrm{reverse} = -0.200$ [$-0.237$, $-0.163$], $\beta_\mathrm{frozen} = +0.004$; forward minus reverse $+0.400$ [$+0.334$, $+0.466$] (cluster bootstrap over the five scenes), per $f$: $f=0.5$: $+0.141$ [$+0.081$, $+0.226$]; $f=0.9$: $+0.658$ [$+0.555$, $+0.771$]; the two signs agree on 14 of 14 models; per scene the difference is $+0.418$, $+0.334$, $+0.492$, $+0.465$, $+0.292$ (\cref{fig:opposite_spin}). Reading, fixed before any output was opened: a directional signature exists (an interval excluding zero on the object-frame side and 14 of 14 models agreeing, against nine required), so the paper keeps ``the camera absorbs the motion'' and states both betas with their interval. ATE/path of the reverse member against its frozen pair: $+0.0414$ [$+0.0321$, $+0.0513$] pooled over the twenty cells, against $+0.0412$ [$+0.0341$, $+0.0482$] for the forward member on the same cells ($f=0.5$: reverse $+0.0243$ [$+0.0175$, $+0.0309$], forward $+0.0291$ [$+0.0213$, $+0.0412$]; $f=0.9$: reverse $+0.0585$ [$+0.0412$, $+0.0747$], forward $+0.0533$ [$+0.0427$, $+0.0657$]; the A20.5 record over all five seeds and both textures: $f=0.5$: $+0.0291$ [$+0.0213$, $+0.0412$]; $f=0.9$: $+0.0533$ [$+0.0427$, $+0.0657$]).

\paragraph{Wording and Figure Corrections (A22.10).} The exact replacements are listed in the results log and in \texttt{results/phase10/wording\_replacements.json}; \cref{tab:zoo} of this supplement now lists TraceAnything without cameras (its wrapper returns placeholder poses, \texttt{poses\_source} = none, as Section~3.4 of the main text states; the table was wrong), \cref{tab:bench} records the generating decomposition on Axis~A, \cref{tab:guidance} reads every situation by mask quality, camera accuracy and the joint reconstruction, and the \cref{fig:matched} caption's panels are named in their drawn order.

\end{document}